\documentclass{article}

\usepackage{arxiv}

\usepackage[utf8]{inputenc} 
\usepackage[T1]{fontenc}    
\usepackage{hyperref}       
\usepackage{url}            
\usepackage{booktabs}       
\usepackage{amsfonts,amsmath,amssymb}       
\usepackage{nicefrac}       
\usepackage{microtype}      
\usepackage{graphicx}

\usepackage{color, colortbl}

\usepackage{mathptmx,amsthm}
\usepackage{xcolor}
\usepackage{multirow}
\usepackage{listings}%
\usepackage{mathrsfs}%
\usepackage{array}%
\usepackage{placeins}%

\newcommand{\ftick}{\textcolor{blue!70!black}{\ensuremath{\checkmark}}}
\newcommand{\ltick}{\textcolor{red!75!black}{\ensuremath{\checkmark}}}

\newcommand{\ie}{\textit{i.e.,}}
\newcommand{\eg}{\textit{e.g.,}}
\newcommand{\commenttxt}[1]{}

\newcommand{\mybar}{\kern1pt\rule[-\dp\strutbox]{.8pt}{\baselineskip}\kern1pt}

\renewcommand{\mytagline}{} 
\renewcommand{\mydate}{August 8, 2026} 
\newcommand{\citep}[1]{\cite{#1}}
\newcommand{\vifhigh}[1]{\underline{#1}}

\theoremstyle{definition}

\title{Spatial Heterogeneity-Aware Multi-Hazard Susceptibility and Risk Mapping at Regional Scale}
\author{
  Aswathi Mundayatt,
  Siddharth Anil\thanks{Contributed equally to this work.},
  Hitanshu Seth\textsuperscript{*}, and 
  Jaya Sreevalsan-Nair\thanks{Corresponding author, \texttt{jnair@iiitb.ac.in}}
  \\
  Graphics-Visualization-Computing Lab,\\
  International Institute of Information Technology Bangalore, Karnataka 560100, India \\
  \texttt{http://www.iiitb.ac.in/gvcl} 
}

\begin{document}
\maketitle

\begin{abstract}
Floods and landslides often occur within the same regional landscapes, but their relationships with environmental controls vary across space. This study develops a spatial heterogeneity-aware framework for flood--landslide susceptibility and relative-risk mapping in the state of Kerala in India and Nepal. The framework combines 15~km x 15~km grid cells with region-specific contextual zones (from a zonation map) and compares two training strategies. These two strategies for selection of training data within this framework are proximity-gated cross-zone training (S1) and ecology-gated zone-constrained training (S2). S1 is designed to allow geographically nearby models to be assigned across contextual boundaries, whereas S2 is designed to restrict model development and assignment to the same contextual zone. Random Forest models are developed separately for flood and landslide using strategy-specific predictor sets and evaluated on spatially held-out test samples. The resulting susceptibility surfaces are integrated with CRITIC-weighted exposure and vulnerability indices to produce hazard-specific and nine-class bivariate relative-risk maps. S1 achieved higher mean accuracy, precision, recall, F1-score, AUC--ROC, and PR--AUC for both hazards and study regions. The largest difference occurred for Nepal flood susceptibility, where AUC--ROC increased from 0.728 under S2 to 0.886 under S1 and PR--AUC increased from 0.512 to 0.823. S2 nevertheless produced lower mean Brier scores for both Nepal hazards and retained zone-specific differences in predictor selection, SHAP rankings, and response patterns, particularly in Kerala. Both strategies reproduced the broad separation between flood-prone lowlands and landslide-prone uplands, but differed in the spatial patterns of susceptibility and risk classes. Overall likelihood of agreement between the bivariate risk maps was 0.521 in Kerala and 0.711 in Nepal, with allocation disagreement exceeding quantity disagreement in every S1--S2 risk-map comparison. Susceptibility-to-risk correspondence for the bivariate maps remained below 0.350, showing that exposure and vulnerability substantially changed the locations identified as priorities. The findings show that cross-zone learning supports stronger regional discrimination, while zone-constrained learning preserves differences among environmental settings. Hence, a regional multi-hazard system is recommended to combine both forms of model learning rather than rely on either strategy alone.
\end{abstract}


\section{Introduction} \label{sec:introduction}
Natural hazards continue to cause substantial human and economic losses and impede sustainable development worldwide. 7,348 major disaster events were recorded globally during 2000-2019, involving $\sim$1.23 million casualties, affecting about 4.2 billion people, and causing USD\$2.97 trillion in reported economic losses; where the affected-population estimate includes individuals affected more than once~\citep{cred2021human}. Disaster-risk research has consequently shifted from hazard-centered assessment towards broader risk-based perspectives. Nevertheless, single-hazard studies remain dominant despite growing international interest in multi-hazard and multi-risk approaches~\citep{ward2022invited}. Assessing hazards separately can overlook their combined implications for populations, infrastructure, and other exposed and vulnerable elements~\citep{stalhandske2024global}. The limitations of separate hazard assessment are especially evident in monsoon-dominated mountain regions, where intense rainfall, steep terrain, and fragile geological conditions connect hydrological and gravitational processes. Floods and landslides may therefore occur in the same locations or form cascading hazard sequences~\citep{shrestha2026compound}. Regional multi-hazard susceptibility mapping identifies areas susceptible to individual and overlapping hazards and can support disaster-risk reduction and the evaluation of mitigation strategies~\citep{pourghasemi2020assessing}.

Several environmental and anthropogenic factors affect floods and landslides at varying degrees. In a national-scale assessment of riverine flood susceptibility in India, proximity to rivers, drainage density, and annual rainfall received the highest importance weights among the factors considered~\citep{kumar2023assessment}. Landslide susceptibility in Shahpur Valley was assessed using geological, topographic, land-use, and proximity-related variables, while prolonged rainfall and human modification of fragile slopes were also identified as important influences on landslide occurrence~\citep{rahman2022assessment}. Multi-hazard frameworks therefore often assess individual hazards separately before examining their spatial combination~\citep{ullah2022multi}. Susceptibility represents the relative spatial propensity for hazard occurrence, whereas risk additionally considers the exposure and vulnerability of the elements located in susceptible areas~\citep{fell2008guidelines}. This distinction is especially relevant in mountain regions, where areas of overlapping hazard susceptibility can also contain substantial exposed populations. In the Hindu Kush Himalaya, 49\% of the population within the assessed region occupied areas classified as highly susceptible to multiple hazards~\citep{rusk2022multi}. Multi-hazard assessment must therefore consider both the spatial overlap of flood and landslide susceptibility and the exposure and vulnerability of affected elements. The reliability of these assessments depends in part on how hazard--factor relationships are modelled across regional landscapes.

Flood and landslide susceptibility studies commonly integrate remote sensing, geographic information systems (GIS), hazard inventories, and spatial conditioning-factor datasets within data-driven modelling frameworks. These studies generally prepare hazard (positive) and non-hazard (negative) samples, extract environmental variables, train and validate predictive models, and apply the fitted relationships across the study area. Statistical, machine-learning (ML), deep-learning (DL), ensemble, and hybrid approaches have been used for this purpose, with performance depending on data preparation, predictor selection, model configuration, and validation strategy~\citep{pourzangbar2025analysis,huang2024modelling}. Regional applications frequently compare several candidate algorithms before selecting an optimal model for susceptibility mapping. For example, five boosting algorithms based on sixteen hydrometeorological and geo-environmental factors have been evaluated for flood susceptibility mapping in Idukki District~\citep{saravanan2023flood}, while four tree-based ensemble models using thirteen conditioning factors have been compared for landslide susceptibility mapping in Murgul District~\citep{usta2024comparison}. The selected models are then applied to conditioning-factor data across the study area to generate continuous susceptibility surfaces. The reliability of these maps depends not only on predictive accuracy but also on whether the learned relationships remain valid across spatially distinct parts of the study area. Random data partitioning can place nearby and environmentally similar observations in both model-development and evaluation subsets, producing optimistic performance estimates for geographically independent locations. Regional landslide modelling has shown that performance estimates obtained through random cross-validation can exceed those obtained through spatial cross-validation by 6--18\%~\citep{kumar2025random}. Thus, transfer learning across micro-regions requires explicit consideration when susceptibility models are developed for environmentally diverse regions.

Regional susceptibility models must also account for spatial heterogeneity. Topographic, geological, hydrological, land-cover, and anthropogenic conditions vary across space, and the relationships between these factors and hazard occurrence can also vary geographically. This variation is commonly described as spatial non-stationarity. A study-wide model represents hazard--factor relationships using one general formulation for the entire region, whereas local models allow these relationships to vary between locations. Local analyses have shown that study-wide models can obscure spatial differences in predictor effects. Elevation, for example, appeared uninformative in a global logistic-regression model but showed a weak positive relationship in the northern part of the study area when analyzed locally~\citep{chalkias2020exploring}. Landslide-conditioning factors have also been shown to operate at different spatial scales, indicating that a single bandwidth may not adequately represent all relationships~\citep{li2022spatial}. Local feature-importance patterns derived through geographical random forest modelling also varied across Liangshan, demonstrating that the contributions of conditioning factors were not spatially uniform~\citep{dai2023examining}. Similar patterns have been reported for floods. Incorporating spatial variation improved flood-susceptibility model performance~\citep{lin2021spatial}, while analyses across geomorphic units revealed different dominant controls, including impervious surfaces, topographic wetness, and elevation~\citep{zhang2026interpretable}. High overall predictive performance or a study-wide feature-importance ranking therefore does not necessarily indicate that hazard--factor relationships are adequately represented within every environmental context.

Susceptibility studies commonly address spatial heterogeneity through regional partitioning, geographic neighborhoods, and information transfer. Regional-partitioning approaches divide the study area according to environmental context and develop separate models for the resulting spatial units. An ecoregion-based landslide assessment in Oregon partitioned the study area into seven Level III ecoregions and developed region-specific models, which revealed differences in dominant conditioning factors and delineated high-susceptibility areas more effectively than a state-wide model~\citep{xu2025ecoregion}. Geographic-neighborhood approaches instead define local relationships without relying on predefined environmental zones. A geographical random forest applied to 4,452 slope-based mapping units represented spatial variation in model relationships and factor importance across the study area~\citep{teke2024spatially}. The type and scale of the spatial units can also influence the resulting susceptibility representation. Urban flood models developed using catchment, street, and grid units showed that analytical scale affected predictive accuracy, mapped susceptibility, and estimated predictor responses~\citep{wu2026multi}. Information-transfer methods use observations from other regions to support modelling in the target area. These methods either use entire source areas that resemble the target region or select only source observations with environmental conditions similar to those of the target~\citep{wang2022transfer}. Environmental similarity has also been used to select source-region landslide records that match the target conditions, thereby reducing the transfer of information that is not representative of those conditions~\citep{zhao2024heterogeneous}. These methods differ in how they partition regional landscapes and in how they select training information and assign models to prediction locations.

Two principles commonly guide the selection of training information and the assignment of models across heterogeneous landscapes. Proximity-based approaches emphasize geographically local information on the assumption that local spatial structure can reveal relationships obscured by a single study-wide model. This principle has been implemented through geographically weighted random forest modelling to represent spatial variation in landslide-conditioning effects~\citep{lu2025geographically}. Environmental-similarity approaches instead select training data with predictor conditions similar to those of the target area, regardless of geographic distance. The distinction between these forms of correspondence is also evident in model evaluation, where geographic blocking is based on spatial separation and environmental blocking is based on similarity in predictor characteristics~\citep{koldasbayeva2025foundation}. Although developed for cross-validation, this distinction shows that geographic proximity and environmental similarity represent different forms of correspondence. Transfer-learning experiments have further shown that the source region with the greatest measured similarity to a target can produce the strongest transferred model, whereas less similar source regions produce weaker results~\citep{wang2025region}. The usefulness of information from other regions nevertheless varies between target areas. For Random Forest models, source-only transfer outperformed target-only training in one target region but performed substantially worse in another, while combining source information with local target data produced the highest AUC in both regions~\citep{singh2024ensembled}. Models that fit local conditions closely may also fail to retain the same predictive performance outside the training area. In urban flood modelling, Random Forest achieved stronger performance within the training domains, whereas a convolutional neural network (CNN) retained greater predictive capability outside those domains and benefited more from transfer learning~\citep{seleem2022transferability}. Geographic proximity and environmental similarity therefore provide different, but potentially complementary, bases for regional model construction.

Multi-hazard assessment often models individual hazards separately before combining their spatial outputs. Separate flood and landslide susceptibility maps have been integrated into compound multi-hazard representations to identify areas affected by individual and overlapping hazards~\citep{rehman2022multi}. Preserving the individual hazard components allows spatial overlap to be examined without assuming that floods and landslides have identical controls. Similar overlay-based approaches have superimposed flood and landslide risk maps to delineate areas affected by their combined effects~\citep{sajid2025integrated}. Multi-hazard risk assessment extends this integration by considering the exposure and vulnerability of populations and assets located in hazard-prone areas. A regional multi-hazard assessment combined seismic, landslide, and flood susceptibility with population density, vulnerable population, literacy, employment, building density, and road density to derive a multi-hazard risk index~\citep{chauhan2025machine}. Vulnerability can therefore be represented through multiple demographic, socioeconomic, physical, and infrastructural dimensions. A systematic review of multi-hazard social vulnerability identified demographic characteristics, socioeconomic status, health, living conditions, exposure, and the capacity to withstand and recover from hazard impacts as recurrent determinants~\citep{drakes2022social}. Composite exposure and vulnerability indices also require an explicit procedure for assigning relative importance to their indicators. Flood-risk frameworks have addressed this requirement through combined subjective and objective procedures, including CRITIC--entropy weighting for exposure and vulnerability indicators~\citep{chen2024urban}.

Heterogeneity-aware susceptibility modelling and multi-hazard risk integration have largely been studied using different regions, hazards, models, spatial units, and performance measures. Consequently, there is limited evidence on how alternative procedures for constructing training regions and assigning local models across heterogeneous landscapes affect results when the remaining modelling conditions are kept comparable. This matters in multi-hazard assessment because separately modelled susceptibility surfaces are subsequently combined and can also be integrated with exposure and vulnerability information~\citep{rehman2022multi,chauhan2025machine}. Differences in susceptibility-model construction can therefore affect predictive performance, factor interpretation, local reliability, mapped susceptibility, flood--landslide overlap, and the final prioritization of risk. A common regional framework is needed to examine these effects from model development through to the final risk maps. 

This study develops a spatial heterogeneity-aware framework for regional flood--landslide susceptibility and risk mapping in Nepal and Kerala, two monsoon-affected regions with contrasting physiographic and environmental settings. Zoning or zonation maps (\eg~ecological, environmental, seismic zone maps, etc.) are usable in the framework to reflect spatial heterogeneity in the region. Hence, these zones are generically referred to as \textit{contextual zones}, hereafter. This study examines how different procedures for constructing training regions and assigning models across contextual zones affect susceptibility and risk mapping in the framework. Two training strategies of interest in this study are where the models include information of neighboring ecological zones, \ie~proximity-based training, and where the models for different ecological zones are independently trained, \ie~ecology zone-based training.

The core contributions of the study are as follows:
\begin{itemize}
    \item A common regional framework that combines regular spatial units with contextual zones, allowing training information either to be shared across zone boundaries or restricted to the corresponding contextual zone.

     \item An explicit representation of two complementary strategies for using regional training information: (i) sharing geographically proximate information across contextual boundaries, referred to as \textit{proximity-gated cross-zone training} (S1), and (ii) restricting training and model assignment to the corresponding contextual zone, referred to as \textit{ecology-gated zone-constrained training} (S2).
     
    \item A direct connection between susceptibility-model design and multi-hazard risk mapping, allowing the effects of training-region construction and model assignment to be traced through factor interpretation, flood--landslide overlap, and the prioritization of exposed and vulnerable areas.

    \item Evidence from regions selected as study areas, Kerala and Nepal, on the relative value of cross-zone generalization and zone-specific modelling under contrasting physiographic and environmental conditions.
\end{itemize}

Implementing both strategies S1 and S2 within the proposed framework using the same datasets and Random Forest models enables extensive evaluation of their differences in predictor-screening scope, training-region construction, and model assignment under comparable conditions.

The codebase is available at \url{https://github.com/GVCL/Susceptibility-Mapping-FL-Hetero}

\section{Study Area} \label{sec:studyarea}
\begin{figure*}[t]
    \centering
    \includegraphics[
        width=\textwidth,
        height=0.95\textheight,
        keepaspectratio
    ]{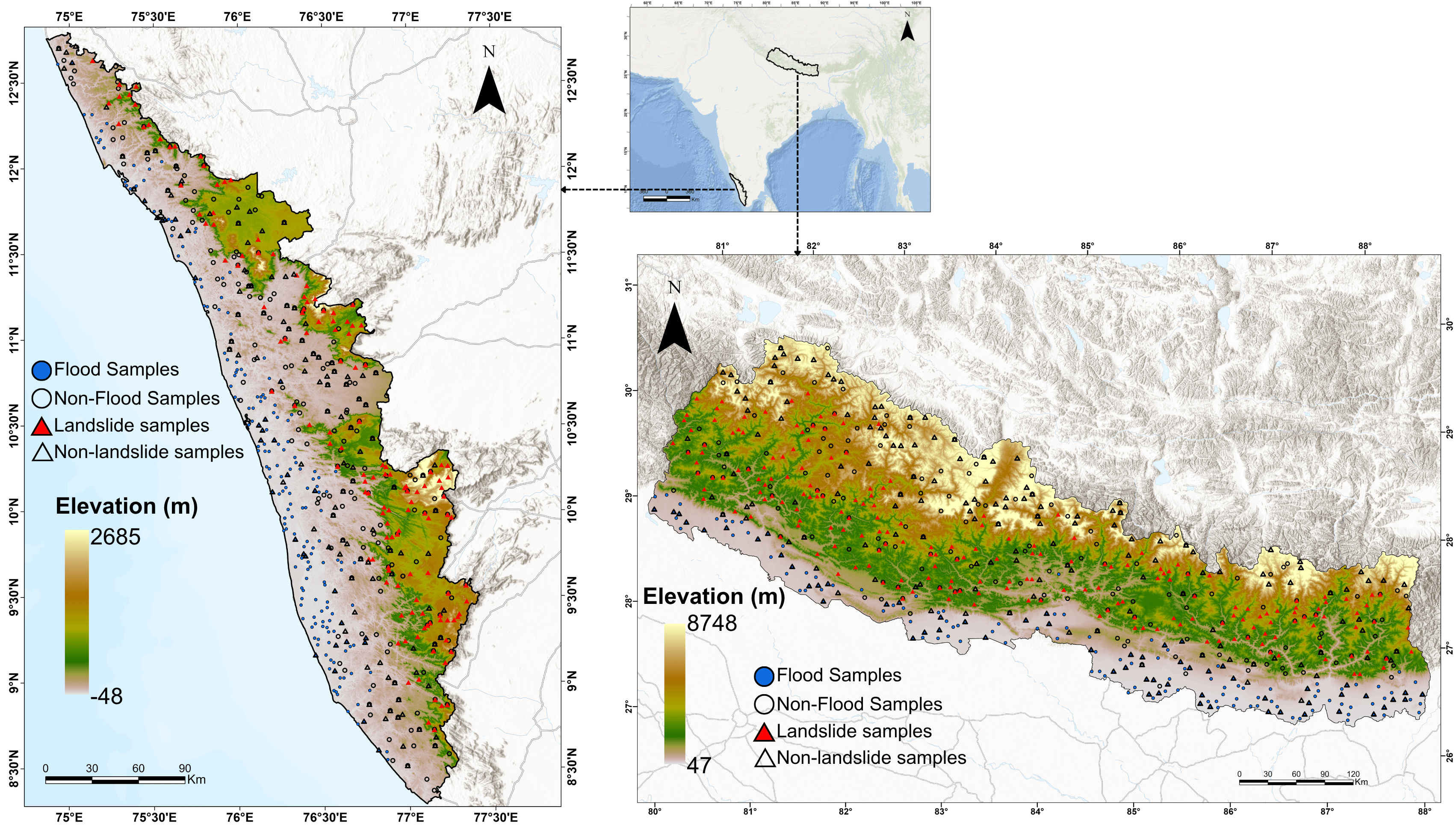}
    \caption{Location map of the study area with multi-hazard sample points for (Left) Kerala, India, and (Right) Nepal.}
    \label{fig:study_area}
\end{figure*}

This study is conducted in two regions, namely, the state of Kerala in India and the country of Nepal.  They both are South Asian monsoon regions affected by recurrent floods and landslides under contrasting physiographic, climatic, drainage, land-use, and settlement conditions. Kerala represents a narrow humid tropical coastal--mountain system in which the Western Ghats, midlands, and coastal lowlands are closely connected over a short east--west distance. Nepal represents a large Himalayan mountain--plain system characterized by pronounced north--south gradients. These contrasting settings provide a suitable basis for evaluating spatial susceptibility and risk modelling across different monsoon-driven multi-hazard environments. The locations of the two study regions are shown in Figure~\ref{fig:study_area}.

Kerala is located along the southwestern coast of India between the Arabian Sea and the Western Ghats. The state covers approximately $38{,}863~\mathrm{km}^{2}$ and extends from about $8^{\circ}17^{\prime}30^{\prime\prime}\mathrm{N}$--$12^{\circ}47^{\prime}40^{\prime\prime}\mathrm{N}$ and $74^{\circ}27^{\prime}47^{\prime\prime}\mathrm{E}$--$77^{\circ}37^{\prime}12^{\prime\prime}\mathrm{E}$. Its physiography forms a west--east transition from coastal lowlands through undulating midlands to the steep highlands of the Western Ghats. Elevation rises from near sea level along the coast to approximately $2695~\mathrm{m}$ at Anamudi, resulting in strong spatial variation in slope, drainage, rainfall, vegetation, land use, and settlement conditions~\citep{velmurugan2026climate}.

Kerala has a humid tropical monsoon climate. The southwest monsoon from June to September contributes the major share of annual rainfall, while the northeast monsoon provides additional rainfall during October and November. Rainfall is generally higher over the Western Ghats because of orographic uplift and varies considerably across the state. Kerala is drained predominantly by short, steep west-flowing rivers that rapidly convey runoff from the highlands toward densely settled lowlands, floodplains, wetlands, agricultural areas, and coastal zones~\citep{hao2020constructing}.

The close spatial association of steep highlands, intense monsoon rainfall, short drainage basins, dense settlement, plantation agriculture, road networks, and modified slopes makes Kerala susceptible to both floods and landslides. Landslides, including shallow failures and debris flows, occur mainly along the Western Ghats and adjacent highlands, where intense or prolonged rainfall increases pore-water pressure and destabilizes weathered slopes. Their occurrence is further influenced by vegetation removal, slope modification, terracing, road construction, and obstruction of natural drainage, which are largely attributed to anthropogenic activities~\citep{gadgil2014western}. Floods are concentrated primarily in low-lying river valleys, coastal plains, wetlands, urban areas, and agricultural floodplains, where rapid runoff and limited drainage capacity favor water accumulation~\citep{kuriakose2009history}.

Nepal lies along the southern slope of the central Himalaya between approximately $26^{\circ}22^{\prime}\mathrm{N}$--$30^{\circ}27^{\prime}\mathrm{N}$ and $80^{\circ}04^{\prime}\mathrm{E}$--$88^{\circ}12^{\prime}\mathrm{E}$, covering an area of about $147{,}181~\mathrm{km}^{2}$. Elevation ranges from approximately $60~\mathrm{m}$ in the southern Terai plains to $8848~\mathrm{m}$ in the High Himalaya. The country is commonly divided into the Terai, Siwalik, Middle Mountains, High Mountains, and High Himalaya physiographic belts. These belts differ markedly in relief, slope, drainage, geology, vegetation, settlement density, and land-use intensity, creating a highly heterogeneous hydro-geomorphic environment~\citep{uddin2015development}.

The climate of Nepal is strongly influenced by the South Asian summer monsoon, which contributes nearly 80\% of the annual precipitation. Rainfall varies substantially across the country because of elevation, topographic orientation, and orographic effects. Dry trans-Himalayan districts receive less than $200~\mathrm{mm}$ of annual rainfall, whereas highly exposed southern mountain slopes may receive more than $5000~\mathrm{mm}$~\citep{karki2017rising}. This climatic and physiographic variability strongly controls the spatial distribution of flood and landslide hazards.

Flooding is concentrated mainly in the Terai, lowland river valleys, and floodplains, where monsoon runoff from the Himalayan catchments, high river discharge, sedimentation, and restricted drainage promote inundation. Landslides occur predominantly in the hill and mountain belts, where steep slopes, fragile geological conditions, river incision, road construction, and other forms of slope disturbance increase terrain instability. Although monsoon rainfall is a major trigger of both hazards, their spatial expressions differ, with flood impacts concentrated more strongly in the southern plains and landslide impacts in the middle hills and mountain terrain~\citep{gaire2015disaster}.

The study areas of Kerala and Nepal therefore represent two distinct but complementary flood--landslide environments. Nepal is characterized by large-scale Himalayan elevational, climatic, and river-basin gradients, whereas Kerala is characterized by the close coupling of steep humid uplands, short river systems, and densely occupied coastal lowlands. Their comparison enables the assessment of susceptibility and risk patterns under markedly different monsoon-controlled physiographic settings. The selection of these study areas demonstrates the generalizability of the proposed framework.

\section{Multi-Source Datasets of Interest} \label{sec:datasets}

\begin{table}[!tbp]
\caption{Candidate conditioning factors used for flood and landslide susceptibility modelling in Kerala and Nepal.}
\label{tab:conditioning_factors}
\centering
\small
\renewcommand{\arraystretch}{1.10}

\begin{tabular*}{\linewidth}{@{\extracolsep{\fill}}lcccc@{}}
\toprule
\multirow{2}{*}{\textbf{Conditioning factor}} &
\multicolumn{2}{c}{\textbf{Kerala}} &
\multicolumn{2}{c}{\textbf{Nepal}} \\
\cmidrule(lr){2-3}
\cmidrule(lr){4-5}
&
\textbf{F} &
\textbf{L} &
\textbf{F} &
\textbf{L} \\
\midrule

\multicolumn{5}{@{}l}{\textit{Topographic}} \\
Elevation
& \ftick & \ltick & \ftick & \ltick \\
Profile curvature
& \ftick & \ltick & \ftick & \ltick \\
Plan curvature
& \ftick & \ltick & \ftick & \ltick \\
Slope
& \ftick & \ltick & \ftick & \ltick \\
Aspect
& \ftick & \ltick & \ftick & \ltick \\
TPI
& -- & \ltick & \ftick & \ltick \\
TRI
& -- & \ltick & \ftick & \ltick \\

\addlinespace[3pt]
\multicolumn{5}{@{}l}{\textit{Hydrological}} \\
Distance to rivers
& \ftick & \ltick & \ftick & \ltick \\
TWI
& \ftick & \ltick & -- & -- \\
Flow accumulation
& \ftick & \ltick & -- & -- \\
Flow direction
& \ftick & \ltick & -- & -- \\
Drainage density
& \ftick & \ltick & -- & -- \\
SPI
& \ftick & \ltick & -- & -- \\
STI
& -- & \ltick & -- & -- \\

\addlinespace[3pt]
\multicolumn{5}{@{}l}{\textit{Geological and soil}} \\
Distance to faults
& -- & \ltick & \ftick & \ltick \\
Soil type
& \ftick & \ltick & -- & -- \\
Lithology
& \ftick & \ltick & -- & -- \\

\addlinespace[3pt]
\multicolumn{5}{@{}l}{\textit{Climatic}} \\
Precipitation
& \ftick & \ltick & \ftick & \ltick \\

\addlinespace[3pt]
\multicolumn{5}{@{}l}{\textit{Land-surface characteristics}} \\
LULC
& \ftick & \ltick & \ftick & \ltick \\
NDVI
& \ftick & \ltick & \ftick & \ltick \\
NDBI
& -- & \ltick & \ftick & \ltick \\

\addlinespace[3pt]
\multicolumn{5}{@{}l}{\textit{Anthropogenic}} \\
Population density
& \ftick & \ltick & \ftick & \ltick \\
Distance to roads
& \ftick & \ltick & \ftick & \ltick \\

\bottomrule
\end{tabular*}

\vspace{3pt}

\begin{minipage}{\linewidth}
\scriptsize
\textit{Note:} F and L denote flood and landslide, respectively.
Blue and red checkmarks indicate inclusion in the corresponding initial
conditioning-factor set; -- indicates that the factor was not considered.
\end{minipage}
\end{table}

\begin{table}[!tbp]
\caption{Data sources of the candidate conditioning factors used in Kerala and Nepal.}
\label{tab:conditioning_factor_sources}
\centering
\small
\renewcommand{\arraystretch}{1.10}

\begin{tabular*}{\linewidth}
{@{\extracolsep{\fill}}p{0.30\linewidth}p{0.65\linewidth}@{}}
\toprule
\textbf{Conditioning factor} &
\textbf{Data source/product} \\
\midrule

\multicolumn{2}{@{}l}{\textit{Topographic}} \\
Elevation
& \multirow{7}{=}{NASA Shuttle Radar Topography Mission Digital Elevation Model (SRTM DEM)} \\
Profile curvature & \\
Plan curvature & \\
Slope & \\
Aspect & \\
TPI & \\
TRI & \\

\midrule

\multicolumn{2}{@{}l}{\textit{Hydrological}} \\
Distance to rivers
& Global River Widths from Landsat database and HydroSHEDS (Kerala);
WWF HydroSHEDS Free-Flowing Rivers Network v1 (Nepal) \\
Drainage density (Kerala)
& Global River Widths from Landsat database and HydroSHEDS \\
TWI (Kerala)
& \multirow{5}{=}{NASA Shuttle Radar Topography Mission Digital Elevation Model (SRTM DEM)} \\
Flow accumulation (Kerala) & \\
Flow direction (Kerala) & \\
SPI (Kerala) & \\
STI (Kerala) & \\

\midrule

\multicolumn{2}{@{}l}{\textit{Geological and soil}} \\
Distance to faults
& GEM Global Active Faults Database (Nepal);
Geological Survey of India Bhukosh Portal (Kerala) \\
Soil type (Kerala)
& SoilGrids \\
Lithology (Kerala)
& Geological Survey of India Bhukosh Portal \\

\midrule

\multicolumn{2}{@{}l}{\textit{Climatic}} \\
Precipitation
& Climate Hazards Group InfraRed Precipitation with Station data (CHIRPS) \\

\midrule

\multicolumn{2}{@{}l}{\textit{Land-surface characteristics}} \\
LULC
& MODIS Land Cover Type product (MCD12Q1) \\
NDVI
& Landsat~8 near-infrared and red bands \\
NDBI
& Landsat~8 short-wave infrared and near-infrared bands \\

\midrule

\multicolumn{2}{@{}l}{\textit{Anthropogenic}} \\
Population density
& Gridded Population of the World, Version~4.11 (GPWv4.11) \\
Distance to roads
& Global Roads Inventory Project (GRIP), 2018 \\

\bottomrule
\end{tabular*}
\end{table}

For susceptibility mapping studies, conditioning factors are required to provide variables for the ML models. The list of factors used in this study and their sources are given in Tables~\ref{tab:conditioning_factors} and~\ref{tab:conditioning_factor_sources}, respectively. The hazard inventory is necessary for providing positive and negative samples (Figure~\ref{fig:study_area}) for training ML and DL models. For regional studies, computational grids are used to scale using data parallelism, and a zonation map provides contextual zones, where one model is trained for each zone. The selection of conditioning factors for flood--landslide is explained in detail. In this study, susceptibility maps are further used to compute risk maps, which require exposure and vulnerability indicators.

\subsection{Hazard Inventory}
Flood and landslide inventories are compiled separately for Kerala and Nepal using hazard- and region-specific data sources.

Flood occurrences in both study regions are mapped from Sentinel-1 Synthetic Aperture Radar imagery using the UN-SPIDER flood-mapping workflow~\citep{unspider2024gee} implemented in Google Earth Engine~\citep{gorelick2017google}. Flooded areas are identified from changes in radar backscatter between pre- and post-event image composites. Permanent water bodies and steep terrain are excluded to minimize false detections. The resulting inundation maps are converted into flood occurrence locations for susceptibility modelling.

The Kerala landslide inventory is obtained from the dataset of the 2018 monsoon event~\citep{hao2020constructing}. The dataset was prepared using satellite imagery, field observations, and visual interpretation of pre- and post-event images. The mapped locations are used as landslide occurrences in the present study.

For Nepal, landslide records are obtained from the Cooperative Open Online Landslide Repository (COOLR)~\citep{juang2019using}. The repository combines records from the NASA Global Landslide Catalog with observations submitted through the Landslide Reporter platform. Records located within the Nepal study boundary are extracted for susceptibility modelling.

\begin{figure*}[!ht]
    \centering
    \includegraphics[
        width=\textwidth,
        height=0.95\textheight,
        keepaspectratio
    ]{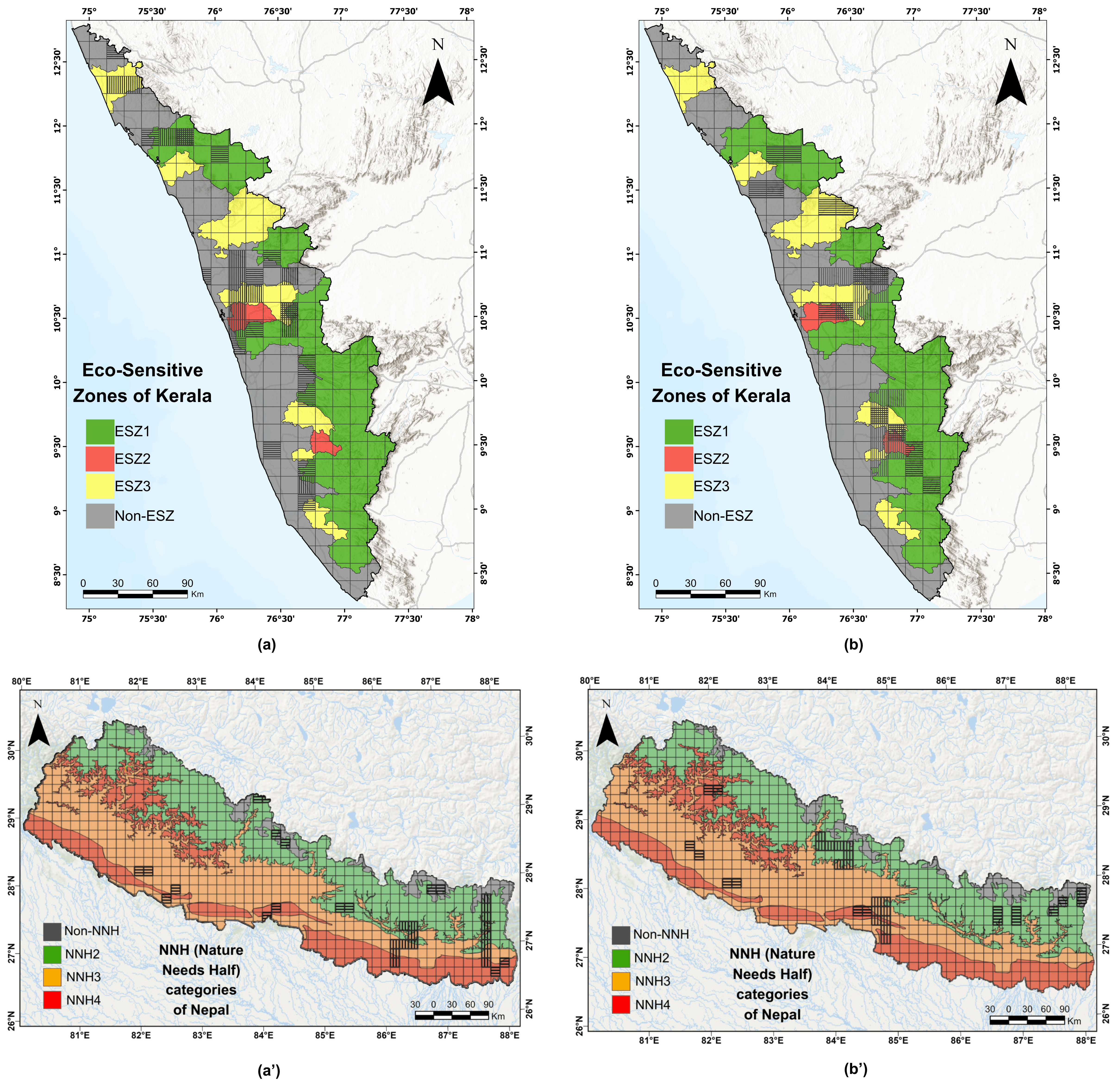}
    \caption{Contextual zones and computational units used in the spatial training design for Kerala [(a), (b)] and Nepal [(a$'$), (b$'$)]. Highlighted units denote the training regions selected for flood susceptibility in (a) and (a$'$) and for landslide susceptibility in (b) and (b$'$). Vertical and horizontal hatching indicate regions selected under training strategies, S1 and S2, respectively.}
    \label{fig:contextual_zones}
\end{figure*}

\subsection{Computational Units and Contextual Zones} \label{sec:units-zones}
To scale for larger regions, 15~km $\times$ 15~km grid cells are used to partition data for parallel computation. Contextual zones are used to distinguish broad environmental settings in which the controls on flood and landslide occurrence may vary. Both spatial partitionings are combined by identifying contextual zones to portions of computational units, as shown in Figure~\ref{fig:contextual_zones}. Thus, data from a computational unit is used for multiple zones sharing it, by using only the portions of the unit specific to the zone.

For Kerala, the zones are derived from the Eco-Sensitive Zone (ESZ) classification proposed in the 2011 Madhav Gadgil Report~\citep{wgeep2011report,gadgil2014western}. The classification comprises ESZ1, ESZ2, and ESZ3, representing progressively lower levels of ecological sensitivity. ESZ1 is concentrated mainly along the eastern Western Ghats and encompasses much of the steep, high-relief terrain and headwater environments of the state. This zone represents conditions associated with slope instability, rapid runoff, and flow concentration within narrow valleys. ESZ2 occurs as smaller and discontinuous patches within the midland--upland transition, where moderate relief and mixed land-cover conditions are relevant to both floods and landslides. ESZ3 is more widely distributed across the midlands and lower upland margins and represents a transition towards gentler and more modified landscapes. The Non-ESZ class occupies much of the coastal plain and adjoining midlands, where low relief, river proximity, extensive settlement, and modified drainage conditions are particularly relevant to flood occurrence.

For Nepal, the zones are derived from the Nature Needs Half (NNH) classification associated with the RESOLVE terrestrial ecoregion framework~\citep{dinerstein2017ecoregion}. The classes present within Nepal are Category~2 (Nature Could Reach Half), Category~3 (Nature Could Recover), and Category~4 (Nature Imperiled), whereas Category~1 is absent. Within the study area, their distribution broadly follows the south--north environmental gradient. Category~2 is concentrated mainly across the northern hill and mountain regions and represents steep terrain, narrow valleys, and mountain drainage environments in which landslides and confined river flooding are important. Category~3 forms a broad central belt across the hill and lower-mountain regions, where dissected terrain and monsoon-driven drainage processes are relevant to both hazards. Category~4 occurs predominantly across the southern lowlands and represents low-gradient river and floodplain environments in which flooding is the principal hazard. Areas outside Categories~2--4 are retained as a separate Non-NNH class.

The ESZ and NNH classifications are retained as region-specific partitions for the zone-based modelling strategies described in Section~\ref{sec:spatial_partitioning_scoring}. The spatial distributions of the ESZ and NNH classes are shown in Figure~\ref{fig:contextual_zones}.

\subsection{Hazard-conditioning Factors}
23 candidate conditioning factors are considered for Kerala and 15 for Nepal, representing topographic, hydrological, geological, climatic, land-surface, and anthropogenic conditions associated with flood and landslide occurrence. The factors are selected based on their established use in susceptibility studies, their relevance to the environmental characteristics of the study regions, and data availability. The factor combinations considered for flood and landslide modelling are summarized in Table~\ref{tab:conditioning_factors}.

The data sources used to prepare these factors are summarized in Table~\ref{tab:conditioning_factor_sources}. For both study regions, elevation is obtained from the Shuttle Radar Topography Mission Digital Elevation Model (SRTM DEM). Slope and aspect are calculated from local elevation gradients, while profile curvature and plan curvature are derived from the rate of slope change along the downslope and contour directions, respectively. The Topographic Position Index (TPI) is calculated as the difference between the elevation of each cell and the mean elevation of its surrounding neighborhood, whereas the Terrain Ruggedness Index (TRI) is calculated from elevation differences between each cell and its neighboring cells to quantify local terrain ruggedness. Precipitation is obtained from the Climate Hazards Group InfraRed Precipitation with Station data (CHIRPS), while population density is obtained from the Gridded Population of the World, Version~4.11 (GPWv4.11). Land use/land cover (LULC) is obtained using the classes of the MODIS Land Cover Type product~\citep{FriedlSullaMenashe2022}. The Normalized Difference Vegetation Index (NDVI) is calculated from the near-infrared and red bands of Landsat~8, whereas the Normalized Difference Built-up Index (NDBI) is calculated from the short-wave infrared and near-infrared bands. Distance to roads is calculated as the shortest distance from each raster cell to the Global Roads Inventory Project road network.

\begin{figure}[!htbp]
    \centering
    \includegraphics[
        width=\linewidth,
        height=0.90\textheight,
        keepaspectratio
    ]{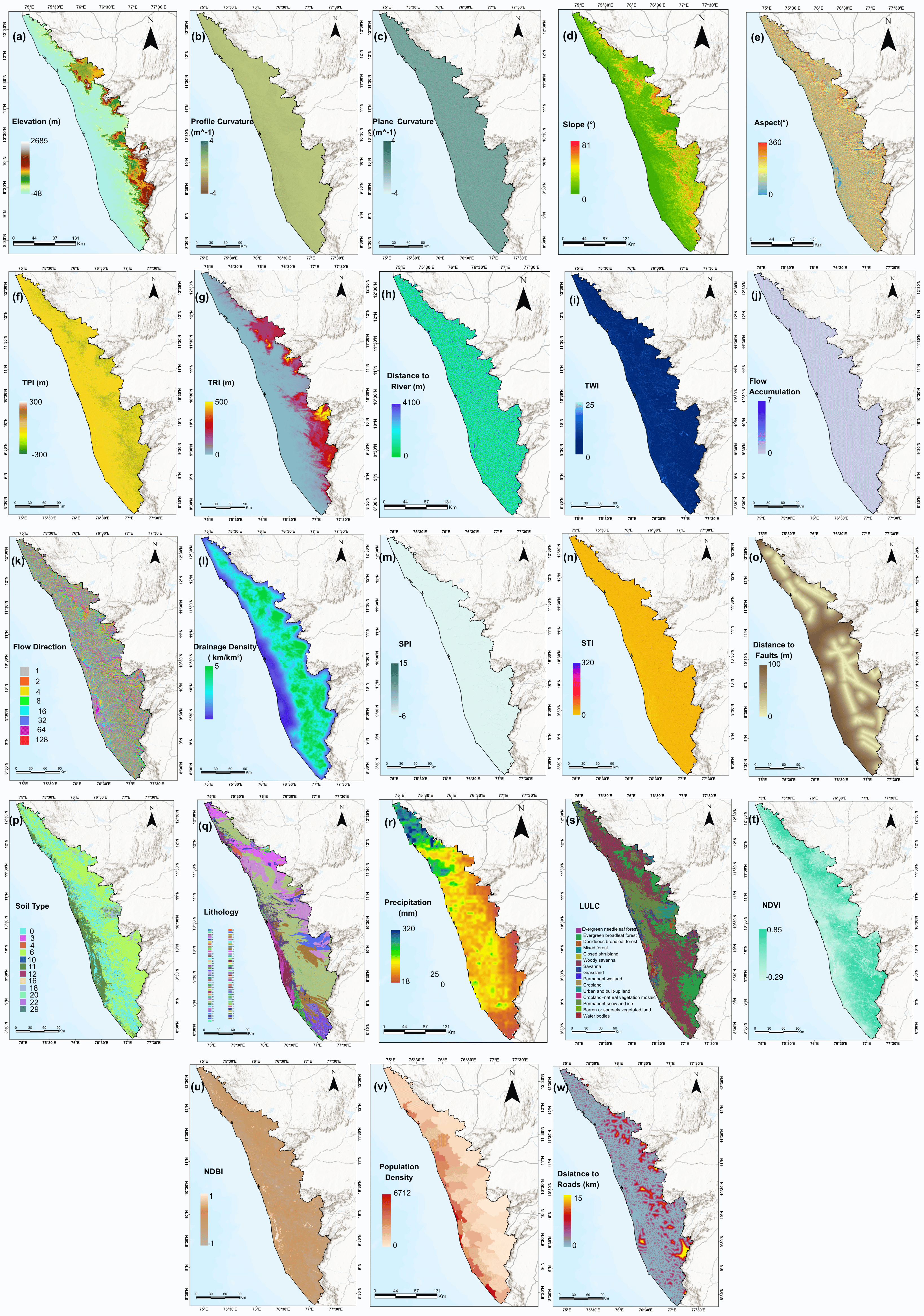}
    \caption{Candidate conditioning factors used for susceptibility modelling in Kerala: (a) elevation, (b) profile curvature, (c) plan curvature, (d) slope, (e) aspect, (f) TPI, (g) TRI, (h) distance to rivers, (i) TWI, (j) Flow Accumulation, (k) Flow Direction, (l) Drainage Density, (m) SPI, (n) STI, and (o) distance to Faults. (p) Soil Type. (q) Lithology. (r) Precipitation, (s) LULC, (t) NDVI, (u) NDBI, (v) Population Density, (w) Distance to Roads.}
    \label{fig:kerala_conditioning_factors}
\end{figure}

\begin{figure}[!htbp]
    \centering
    \includegraphics[
        width=\linewidth,
        height=0.90\textheight,
        keepaspectratio
    ]{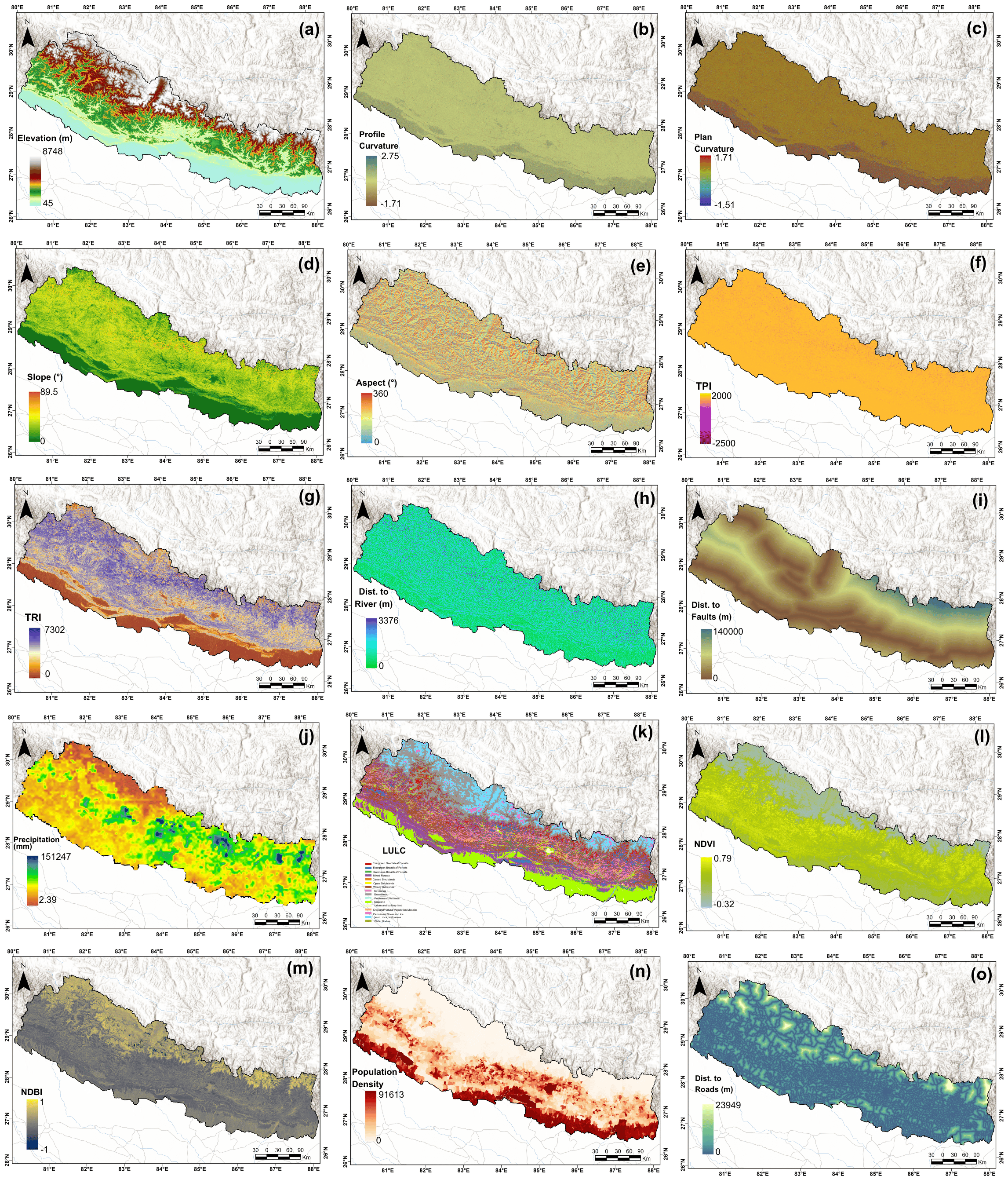}
    \caption{Candidate conditioning factors used for susceptibility modelling in Nepal: (a) elevation, (b) profile curvature, (c) plan curvature, (d) slope, (e) aspect, (f) TPI, (g) TRI, (h) distance to rivers, (i) distance to faults, (j) precipitation, (k) LULC, (l) NDVI, (m) NDBI, (n) population density, and (o) distance to roads.}
    \label{fig:nepal_conditioning_factors}
\end{figure}

For Kerala, the SRTM DEM is further used to derive the Topographic Wetness Index (TWI), flow accumulation, flow direction, Stream Power Index (SPI), and Sediment Transport Index (STI). Flow direction represents the steepest downslope flow path, while flow accumulation represents the number of upstream cells contributing flow to each location. TWI describes the combined influence of contributing area and local slope on terrain wetness, whereas SPI and STI represent runoff erosive power and sediment-transport potential, respectively. Distance to rivers is calculated as the shortest distance from each cell to the mapped river network, while drainage density is derived from the total length of river segments per unit area. Distance to faults and Lithology are obtained from the active-fault network and geological units obtained from the Geological Survey of India Bhukosh Portal~\citep{gsi2026bhukosh}. Soil type is obtained using the SoilGrids dataset~\citep{PoggioDeSousa2020} to extract World Reference Base (2006) Soil Groups~\citep{IUSSWorkingGroupWRB2006}.

For Nepal, distance to rivers is calculated as the shortest distance from each raster cell to the WWF HydroSHEDS Free-Flowing Rivers Network v1. Fault information is obtained from the GEM Global Active Faults Database~\citep{StyronPagani2020} and distance to faults is calculated as the shortest distance from each cell to the mapped active-fault network. These proximity factors represent the influence of drainage systems and regional geological structures on flood and landslide susceptibility. The candidate conditioning factors used for Kerala and Nepal are shown in Figures~\ref{fig:kerala_conditioning_factors} and~\ref{fig:nepal_conditioning_factors}, respectively.

These variables constitute the initial candidate factor sets. Predictor screening and final factor selection are performed separately for each hazard, study region, and modelling scope, as described in Section~\ref{sec:predictor_screening}.

\subsection{Exposure Indicators}
Exposure is characterized using population, built-up surface, and road density, representing the spatial distribution of residents, built assets, and transport infrastructure potentially affected by floods and landslides. The same indicators are adopted for Kerala and Nepal to maintain consistency between the two regional assessments.

Population data are obtained from the 2020 Global Human Settlement Layer population grid (GHS-POP), which reports the residential population within each grid cell~\citep{schiavina2026ghspop}. Built-up surface is obtained from the corresponding 2020 Global Human Settlement Layer product (GHS-BUILT-S), which provides the area occupied by built surfaces within each grid cell~\citep{pesaresi2023ghsbuilt}. These datasets are used to represent the distribution of exposed population and built assets, respectively.

Road-network data is obtained from OpenStreetMap through Geofabrik~\citep{geofabrik_osm}. Motorable road classes are retained, whereas footpaths, cycleways, steps, and other non-motorized features are excluded. The resulting network is used to derive road density as an indicator of the concentration of transport infrastructure potentially affected or disrupted by hazard events.

\subsection{Vulnerability Indicators}
Vulnerability is characterized using socioeconomic deprivation, demographic dependency, and healthcare accessibility. These indicators represent spatial differences in social sensitivity and access to resources that may influence the ability of communities to respond to and recover from flood and landslide impacts.

Socioeconomic deprivation is represented by Version~1 of the Global Gridded Relative Deprivation Index (GRDI)~\citep{ciesin2022grdi}. GRDI measures multidimensional relative deprivation on a scale from 0 to 100, with higher values indicating greater deprivation.

Demographic dependency is derived from the 2020 WorldPop unconstrained age--sex population grids~\citep{worldpop_agesex2020}. The dependent population share is given as:
\begin{equation}
D=\frac{P_{0-14}+P_{65+}}{P_{\mathrm{total}}},
\label{eq:dependent_population_share}
\end{equation}
\noindent where $P_{0-14}$ and $P_{65+}$ denote the populations aged 0--14 years and 65 years and above, respectively, and $P_{\mathrm{total}}$ denotes the total population derived from the same WorldPop dataset. Higher values indicate a greater proportion of children and older adults within the resident population.

Healthcare accessibility is obtained from the 2019 Malaria Atlas Project accessibility-to-healthcare dataset~\citep{weiss2020global}. The dataset reports land-based travel time, in minutes, to the nearest hospital or clinic. Longer travel times indicate lower access to healthcare services and are therefore associated with greater vulnerability.

\subsection{Spatial Data Preprocessing}
The hazard-conditioning factors are harmonized to common reference grids at 30~m for Kerala and 100~m for Nepal. Continuous layers are resampled using bilinear interpolation, whereas nearest-neighbor interpolation is applied to categorical layers.

Exposure and vulnerability indicators are processed at 100~m in both study regions. Road density is calculated from the motorable road network using a 1~km moving window and expressed in km per km\textsuperscript{2}. Cells representing the absence of population, built-up surface, or road length are assigned a value of zero. The dependent population share is also set to zero where the total population is zero. Missing values in the vulnerability indicators are replaced using the nearest valid cell.

The exposure and vulnerability indices are retained at 100~m for Nepal. For Kerala, they are resampled and aligned to the 30~m susceptibility grids before flood and landslide risk estimation.

\section{Methodology} \label{sec:method}
\begin{figure}[!htbp]
    \centering
    \includegraphics[
        width=\linewidth,
        height=0.90\textheight,
        keepaspectratio
    ]{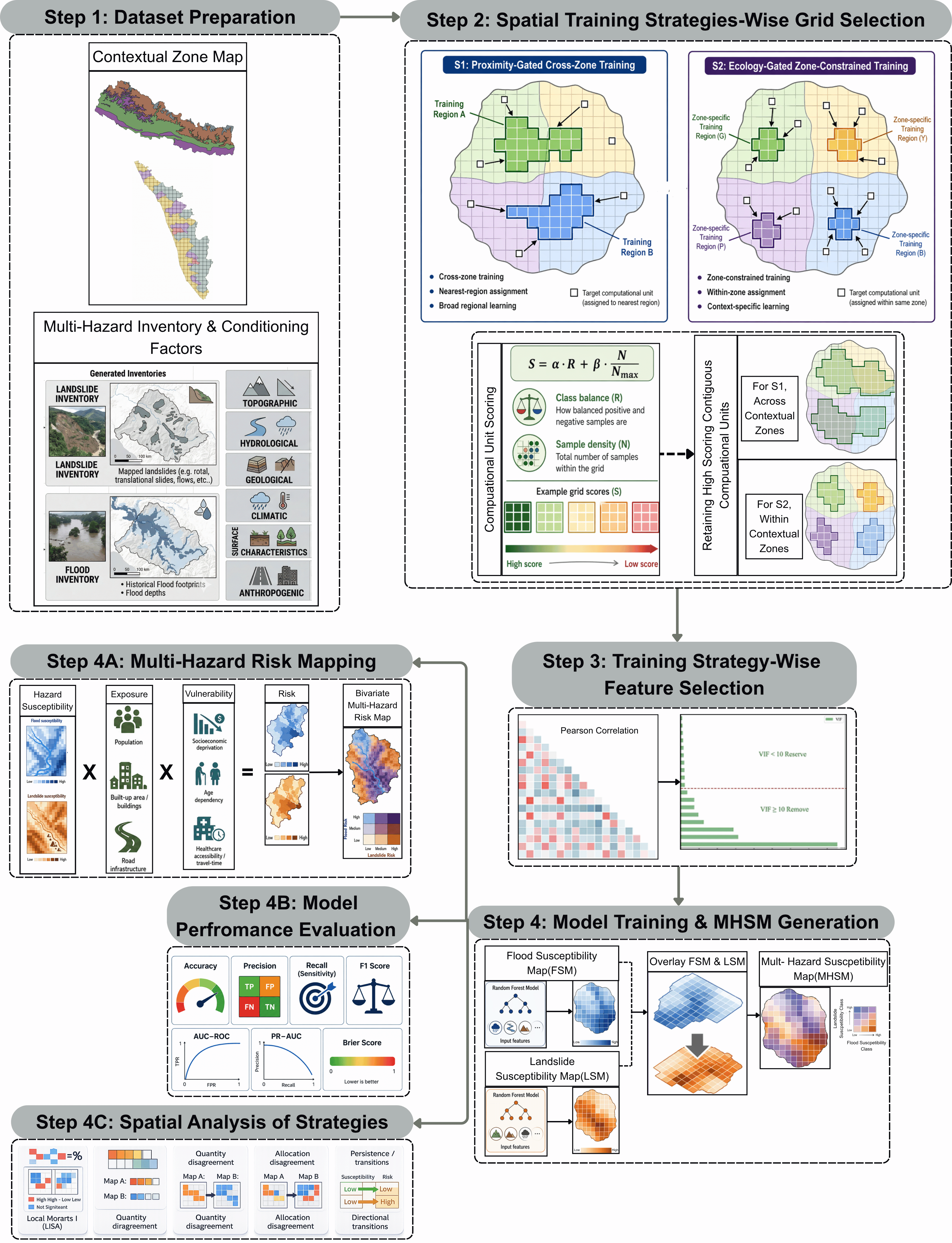}
    \caption{Overall methodological workflow for spatial heterogeneity-aware flood--landslide susceptibility and risk mapping.}
    \label{fig:methodological_workflow}
\end{figure}

The workflow comprises dataset preparation, strategy-specific training-region selection, predictor screening, susceptibility modelling, multi-hazard risk mapping, model-performance evaluation, and spatial comparison of the resulting susceptibility and risk maps. The detailed workflow diagram is given in Figure~\ref{fig:methodological_workflow}.

\subsection{Hazard Occurrence and Non-occurrence Sampling} \label{sec:sample_generation}
Occurrence (positive) and non-occurrence (negative) samples are curated separately for flood and landslide modelling in Kerala and Nepal. To improve the spatial representation of the available inventories, kernel density estimation (KDE) with a Gaussian kernel is applied to the recorded hazard locations~\citep{fan2023comparison,li2025evaluation}. Additional occurrence locations are sampled from the resulting density surfaces and combined with the original inventory records. A minimum spacing of 100~m is maintained among the generated locations to reduce excessive cluttering in areas with dense occurrence records.

As verified non-occurrence inventories are unavailable, non-occurrence locations are identified using a one-class support vector machine (OC-SVM)~\citep{ye2023generating,chen2021one}. The procedure is applied independently to the flood and landslide inventories of each study region. Background samples outside the spatial concentration of the recorded hazards are retained as non-occurrences if they are at least 100~m from known occurrence locations.

The resulting hazard-specific occurrence and non-occurrence samples are used to evaluate candidate training regions and develop the susceptibility models.

\subsection{Spatial Partitioning and Candidate-region Scoring} \label{sec:spatial_partitioning_scoring}
The generation of regions using computational units and contextual zones is referred to as a two-level spatial partitioning framework, as adapted from a previous study~\citep{mundayatt2024scaling}. The spatial partitioning accounts for spatial heterogeneity in the relationships between hazard occurrence and the conditioning factors, as explained in Section~\ref{sec:units-zones}. The framework combines regular spatial neighborhoods with region-specific contextual zones and provides the basis for the two spatial training strategies described in Section~\ref{sec:spatial_training_strategies}.

Kerala and Nepal are divided into regular 15~km $\times$ 15~km grid cells, in the form of a lattice. These grid cells form the basic spatial units of the framework. Each grid cell is then intersected with the corresponding contextual-zone layer, and portions of a cell falling within different zones are retained as separate grid--zone units. The regular grid represents local spatial contiguity, whereas the contextual zones distinguish broader environmental settings. The zone classes are used to organize training-region selection and model assignment but are not included as conditioning factors.

Adjacency is defined using queen contiguity, whereby polygons sharing an edge or a corner are considered neighbors. Separate adjacency structures are constructed for the regular grid cells and the grid--zone units, and contiguous candidate regions are generated using depth-first search. The occurrence and non-occurrence samples within the constituent cells or units of each candidate region are aggregated for evaluation.

Each candidate region is assigned a weighted average score $S$ combining class balance and sample availability:
\begin{equation}
\begin{aligned}
R &= \frac{\min(H_{\mathrm{o}},H_{\mathrm{n}})}
{\max(H_{\mathrm{o}},H_{\mathrm{n}})},\\
N &= H_{\mathrm{o}}+H_{\mathrm{n}}, \\
S &= \alpha R+\beta\frac{N}{N_{\max}},
\end{aligned}
\label{eq:training_region_score}
\end{equation}
\noindent where $H_{\mathrm{o}}$ and $H_{\mathrm{n}}$ denote the numbers of hazard occurrence and non-occurrence samples within a candidate region, respectively. The class-balance term $R$ approaches 1 as the two classes become more evenly represented, while $N$ denotes the total sample count. The normalized term $N/N_{\max}$ expresses sample availability relative to the largest eligible candidate within the corresponding strategy-specific comparison group. $\alpha$ and $\beta$ are the weights for $R$ and normalized $N$, respectively.

In this study, the weights are fixed at $\alpha=0.55$ and $\beta=0.45$ for both hazards and study regions. The greater weight assigned to class balance reduces the likelihood of selecting sample-rich candidates dominated by one class, while the normalized sample-count term favors candidates with stronger sample support. Eligible candidates must contain both occurrence and non-occurrence samples, although no additional minimum sample-count threshold is imposed.

Within each strategy-specific comparison group, candidates are ranked in descending order of $S$. Selection proceeds sequentially, and any lower-ranked candidate overlapping a previously selected training region is excluded.

\subsection{Proposed Spatial Training Strategies and Model Assignment} \label{sec:spatial_training_strategies}
Two spatial training strategies are implemented to examine how training-region configuration and contextual restrictions on training-region selection and model assignment influence susceptibility modelling. Proximity-gated cross-zone training (S1) permits model assignment across contextual-zone boundaries based on geographic proximity, whereas ecology-gated zone-constrained training (S2) restricts both training-region selection and model assignment to the same contextual zone.

Each selected training region is used to develop a separate susceptibility model. For each target grid--zone unit $i$, the nearest eligible training region is identified as

\begin{equation}
\hat{k}_{i}
=
\arg\min_{k\in\mathcal{C}_{i}}
\sqrt{(x_{i}-x_{k})^{2}+(y_{i}-y_{k})^{2}},
\label{eq:model_assignment}
\end{equation}
\noindent where $\hat{k}_{i}$ denotes the training region assigned to target grid--zone unit $i$, $\mathcal{C}_{i}$ is the strategy-specific set of eligible training regions, $(x_i,y_i)$ is the centroid of the target grid--zone unit, and $(x_k,y_k)$ is the centroid of the combined geometry of training region $k$. The grid--zone computational units are defined by intersecting the regular grid cells with the contextual zones; and the training regions selected under S1 and S2 for flood and landslide susceptibility modelling are shown in Figure~\ref{fig:study_area}.

\subsubsection{S1: Proximity-gated Cross-zone Training}
Under S1, each candidate training region comprises eight contiguous regular grid cells and can extend across contextual-zone boundaries. Candidate regions are generated and evaluated separately for each hazard and study region. For the sample-availability term in Equation~\ref{eq:training_region_score}, $N_{\max}$ is defined as the largest total sample count among all eligible eight-cell candidates for the corresponding hazard and study region.

For each hazard and study region, the two highest-ranked non-overlapping candidates are selected. For every target grid--zone unit, both selected S1 training regions are included in the eligible set $\mathcal{C}_{i}$, irrespective of contextual-zone membership. The model developed for the geographically nearest training region is assigned to the target unit. Thus, S1 produces two models for each hazard and study region.

\subsubsection{S2: Ecology-gated Zone-constrained Training}
Under S2, each candidate training region comprises two contiguous grid--zone units belonging to the same contextual zone. Candidate regions are generated and evaluated separately for each contextual zone, hazard, and study region. Accordingly, $N_{\max}$ is defined as the largest total sample count among the eligible two-unit candidates within each corresponding group.

Within each contextual zone, the two highest-ranked non-overlapping candidates are selected. For a target grid--zone unit, the eligible set $\mathcal{C}_{i}$ is restricted to the two training regions selected within its contextual zone. The model developed for the geographically nearest eligible training region is then assigned to the target unit. Thus, model development and assignment remain within the same contextual zone. Because each study region contains four contextual classes, S2 produces eight models for each hazard and study region. 

\subsection{Predictor Screening}\label{sec:predictor_screening}
The goal for predictor screening is to determine the optimal set of predictors or variables for training an ML model. Candidate conditioning factors are screened separately for each hazard and study region. Under S1, screening is conducted at the study-region level, and the resulting predictor set is used for both selected training regions. Under S2, screening is conducted independently within each contextual zone, and the resulting predictor set is used for both training regions selected within that zone.

Pearson correlation and variance inflation factor (VIF) analyses are applied only to continuous predictors. When the absolute Pearson correlation coefficient between two predictors exceeds 0.9, the predictor considered less physically relevant to the corresponding hazard is removed. VIF is then calculated for the remaining continuous predictors and assessed against a threshold of 13. Predictors exceeding this threshold are reviewed individually, and their retention or removal is determined according to their relative physical relevance to flood or landslide occurrence. The VIF assessment is conducted once rather than iteratively.

Categorical predictors, including LULC, lithology, soil type, and flow direction, are not included in the correlation and VIF analyses and are retained in the corresponding predictor sets. The retained predictors are subsequently used to develop the susceptibility models.

\subsection{Flood and landslide susceptibility modelling}\label{sec:susceptibility_modelling}
A spatial holdout design is adopted for model development and evaluation. Hazard occurrence and non-occurrence samples within each selected training region are divided using stratified random sampling into training, validation, and calibration subsets in proportions of 70\%, 20\%, and 10\%, respectively. Stratification preserves the relative proportions of the two classes. The validation subset is used for model selection, while the calibration subset is reserved for probability calibration.

Samples outside the selected training regions form the spatially held-out test sets. Under S1, samples outside the two selected cross-zone training regions constitute a common test set for the corresponding hazard and study region. Under S2, test samples are defined separately within each contextual zone and comprise the samples remaining after the two selected training regions are excluded. These test samples are not used for model selection or probability calibration.

Flood and landslide susceptibility are modelled separately using Random Forest classifiers. A separate model is developed for each selected training region using the predictor set retained at the corresponding screening level. Model hyperparameters are optimized through a preliminary parameter search followed by 20 iterations of Bayesian optimization, with validation AUC--ROC used as the optimization criterion. The selected models are subsequently calibrated on the held-out calibration subset using sigmoid (Platt) calibration.

Predictive performance is assessed on the spatially held-out test samples after each sample is assigned to an eligible model according to the corresponding S1 or S2 assignment rule. Performance is evaluated separately for each hazard, study region, and training strategy using accuracy, precision, recall, F1-score, area under the receiver operating characteristic curve (AUC--ROC), area under the precision--recall curve (PR--AUC), and Brier score. Accuracy, precision, recall, and F1-score quantify classification performance at the selected threshold; AUC--ROC and PR--AUC evaluate discrimination across probability thresholds; and the Brier score measures the accuracy of the calibrated probability estimates, with lower values indicating better probabilistic performance. Under S1, the metrics are calculated separately for the test samples assigned to each selected training-region model; under S2, they are calculated separately within each contextual zone. The values reported for each hazard and study region are the arithmetic means across the corresponding evaluation units.

\subsection{Model Interpretation}\label{sec:model_interpretation}
SHapley Additive exPlanations (SHAP) are used to examine the contributions of the conditioning factors to flood and landslide susceptibility predictions. The analysis is conducted on the uncalibrated Random Forest models developed for the selected training regions using class-balanced samples.

SHAP values are calculated for the occurrence class, and global factor importance is quantified using the mean absolute SHAP value. For categorical factors represented by multiple one-hot encoded variables, the absolute contributions of the associated variables are summed to obtain a single importance value for the original factor.

Under S1, factor importance is calculated for the two selected models within each hazard and study region and then averaged. Under S2, the same procedure is applied to the two models within each contextual zone. The five highest-ranked factors are reported to compare importance patterns across hazards, training strategies, contextual zones, and study regions.

SHAP dependence plots are generated for the highest-ranked factor in each hazard--region combination under S1 and each hazard--region--zone combination under S2, using 2000 class-balanced observations per model. For continuous factors, predictor values are plotted against their corresponding SHAP values. For categorical factors, SHAP values are aggregated at the level of the original factor and examined by category.

\subsection{Multi-hazard Susceptibility and Risk Mapping}\label{sec:multi_hazard_susceptibility_risk_mapping}
Regional flood and landslide susceptibility maps under S1 and S2 are obtained from the models assigned to the target grid--zone units. For each hazard and study region, a common set of Jenks natural-break thresholds is derived from the pooled S1 and S2 susceptibility distributions and used to define low, medium, and high susceptibility classes. The resulting flood and landslide classes are combined to form a nine-class bivariate multi-hazard susceptibility map for each strategy.

Risk is estimated by integrating the continuous susceptibility surfaces with composite exposure and vulnerability indices. Exposure comprises population, built-up surface, and road density, whereas vulnerability comprises dependent population share, healthcare travel time, and socioeconomic deprivation. Before index construction, the indicators are oriented so that higher values indicate greater exposure or vulnerability and are normalized to a common range from 0 to 1.

The exposure and vulnerability indicators are weighted separately using the Criteria Importance Through Intercriteria Correlation (CRITIC) method~\citep{diakoulaki1995determining}. The weight $w_j$ for the indicator $j$ is given as:
\begin{equation}
w_j=\frac{C_j}{\sum_{\ell=1}^{p}C_\ell},
\qquad
\text{ for } C_j=\sigma_j\sum_{k=1}^{p}(1-r_{jk}),
\label{eq:critic}
\end{equation}
\noindent where $\sigma_j$ is the standard deviation of normalized indicator $j$, $r_{jk}$ is the correlation coefficient between indicators $j$ and $k$, and $p$ is the number of indicators in the corresponding index. The composite exposure ($E$) and vulnerability ($V$) indices at location $x$ are defined as:
\begin{equation}
E(x)=\sum_{j=1}^{m}w^{E}_{j}z^{E}_{j}(x),
\qquad
V(x)=\sum_{k=1}^{n}w^{V}_{k}z^{V}_{k}(x),
\label{eq:exposure_vulnerability}
\end{equation}
\noindent where $z^{E}_{j}(x)$ and $z^{V}_{k}(x)$ are the normalized exposure and vulnerability indicators, $w^{E}_{j}$ and $w^{V}_{k}$ are their respective CRITIC weights, and $m$ and $n$ denote the numbers of indicators in the two indices. The weights are derived independently for Kerala and Nepal, while the same region-specific indices are used under S1 and S2.

Flood and landslide risk under strategy $s$ is defined as:
\begin{equation}
\begin{aligned}
R_{\mathrm{F}}^{(s)}(x)
&=S_{\mathrm{F}}^{(s)}(x)\,E(x)\,V(x),\\
R_{\mathrm{L}}^{(s)}(x)
&=S_{\mathrm{L}}^{(s)}(x)\,E(x)\,V(x),
\end{aligned}
\label{eq:hazard_risk}
\end{equation}
\noindent where $R_{\mathrm{F}}^{(s)}(x)$ and $R_{\mathrm{L}}^{(s)}(x)$ are the flood- and landslide-risk values, and $S_{\mathrm{F}}^{(s)}(x)$ and $S_{\mathrm{L}}^{(s)}(x)$ are the corresponding continuous susceptibility values. All components range from 0 to 1.

For each hazard and each study region, common Jenks natural-break thresholds are derived from the pooled S1 and S2 risk distributions and used to define three risk classes, namely, low, medium, and high risk. The resulting flood- and landslide-risk classes are combined to form a final nine-class bivariate multi-hazard risk map for each strategy.

\subsection{Spatial Comparison of Susceptibility and Risk Maps} \label{sec:spatial_comparison}
The spatial outputs of S1 and S2 are compared in terms of local spatial association, agreement between classified risk maps, and changes associated with the integration of exposure and vulnerability. The analyses are conducted in both study regions and, where applicable, separately for flood and landslide.

\subsubsection{Local Spatial Association of Susceptibility}
Local Moran's~$I$ is used to assess local spatial association in the continuous flood and landslide susceptibility surfaces produced under S1 and S2. Before analysis, the susceptibility surfaces are aggregated to 1000~m by calculating the mean susceptibility within each analysis cell. Local Moran's~$I$ is then calculated using row-standardized rook-contiguity spatial weights. Statistical significance is assessed at $p<0.05$ using 999 permutations.

Significant associations are classified as high--high and low--low clusters or high--low and low--high spatial outliers, while all remaining cells are classified as non-significant.

\subsubsection{Agreement between Classified Risk Maps}
Agreement between S1 and S2 is assessed for three scenarios, namely, the flood-risk, landslide-risk, and bivariate multi-hazard risk maps over their common spatial extent. Class-wise overlap is quantified using the Jaccard index:
\begin{equation}
J_c=
\frac{|A_c\cap B_c|}
{|A_c\cup B_c|},
\label{eq:classwise_jaccard}
\end{equation}
\noindent where $A_c$ and $B_c$ denote the spatial units assigned to class $c$ under S1 and S2, respectively. Overall agreement is summarized using the macro and union-weighted Jaccard indices:
\begin{equation}
\begin{aligned}
J_{\mathrm{macro}}
&=
\frac{1}{C}\sum_{c=1}^{C}J_c,\\
J_{\mathrm{uw}}
&=
\frac{
\displaystyle\sum_{c=1}^{C}|A_c\cap B_c|
}{
\displaystyle\sum_{c=1}^{C}|A_c\cup B_c|
},
\end{aligned}
\label{eq:jaccard_summary}
\end{equation}
\noindent where $C$ is 3 for the individual hazard-risk maps and 9 for the bivariate multi-hazard risk maps. The macro index gives equal weight to all classes, whereas the union-weighted index accounts for differences in their spatial extent.

Categorical disagreement is further separated into quantity and allocation components. Let $p_{ij}$ denote the proportion of the common mapped area assigned to class $i$ under S1 and class $j$ under S2. Overall agreement, quantity disagreement, and allocation disagreement are calculated as:
\begin{equation}
\begin{aligned}
A_{\mathrm{o}}
&=\sum_i p_{ii},\\
Q
&=\frac{1}{2}\sum_i
\left|p_{i+}-p_{+i}\right|,\\
D_{\mathrm{a}}
&=1-A_{\mathrm{o}}-Q,
\end{aligned}
\label{eq:quantity_allocation}
\end{equation}
\noindent where $p_{i+}$ and $p_{+i}$ are the marginal class proportions under S1 and S2, respectively. Higher Jaccard and $A_{\mathrm{o}}$ values indicate stronger spatial agreement, whereas $Q$ and $D_{\mathrm{a}}$ quantify disagreement arising from differences in class proportions and spatial allocation, respectively.

\subsubsection{Susceptibility-to-risk Correspondence and Transitions} \label{sec:susceptibility_risk_correspondence}
Susceptibility-to-risk correspondence is assessed for each hazard, study region, and training strategy over the common spatial extent of the classified maps. Overall persistence is defined as the proportion of cells retaining the same low, medium, or high class following the integration of exposure and vulnerability. Class-specific persistence represents the proportion of cells within each susceptibility class that remain in the corresponding risk class. Changes to a higher or lower class on the corresponding risk map are identified as upward and downward transitions, respectively. Class persistence and upward and downward transitions are denoted by $P_{\mathrm{c}}$, $R_{\mathrm{up}}$, and $R_{\mathrm{down}}$, respectively.

For the bivariate maps, persistence represents retention of the same flood--landslide class combination between the susceptibility and risk maps. Spatial correspondence is further assessed using the Jaccard, overall agreement, and quantity--allocation disagreement measures defined in the preceding subsection, with the susceptibility and risk maps replacing S1 and S2 in the corresponding formulations.

\section{Results} \label{sec:result}
This section presents the strategy-specific predictor-screening outcomes, predictive performance, and SHAP-based interpretation of the flood and landslide susceptibility models. It then examines the spatial distribution of the resulting bivariate susceptibility and risk classes, local spatial association, agreement between the S1 and S2 risk maps, and susceptibility-to-risk correspondence and transitions in Kerala and Nepal.

\subsection{Strategy-specific Predictor-screening Outcomes} \label{sec:predictor_screening_outcomes}
\begin{figure}[!p]
    \centering
    \includegraphics[
        width=0.8\textwidth,
        height=\textheight,
        keepaspectratio
    ]{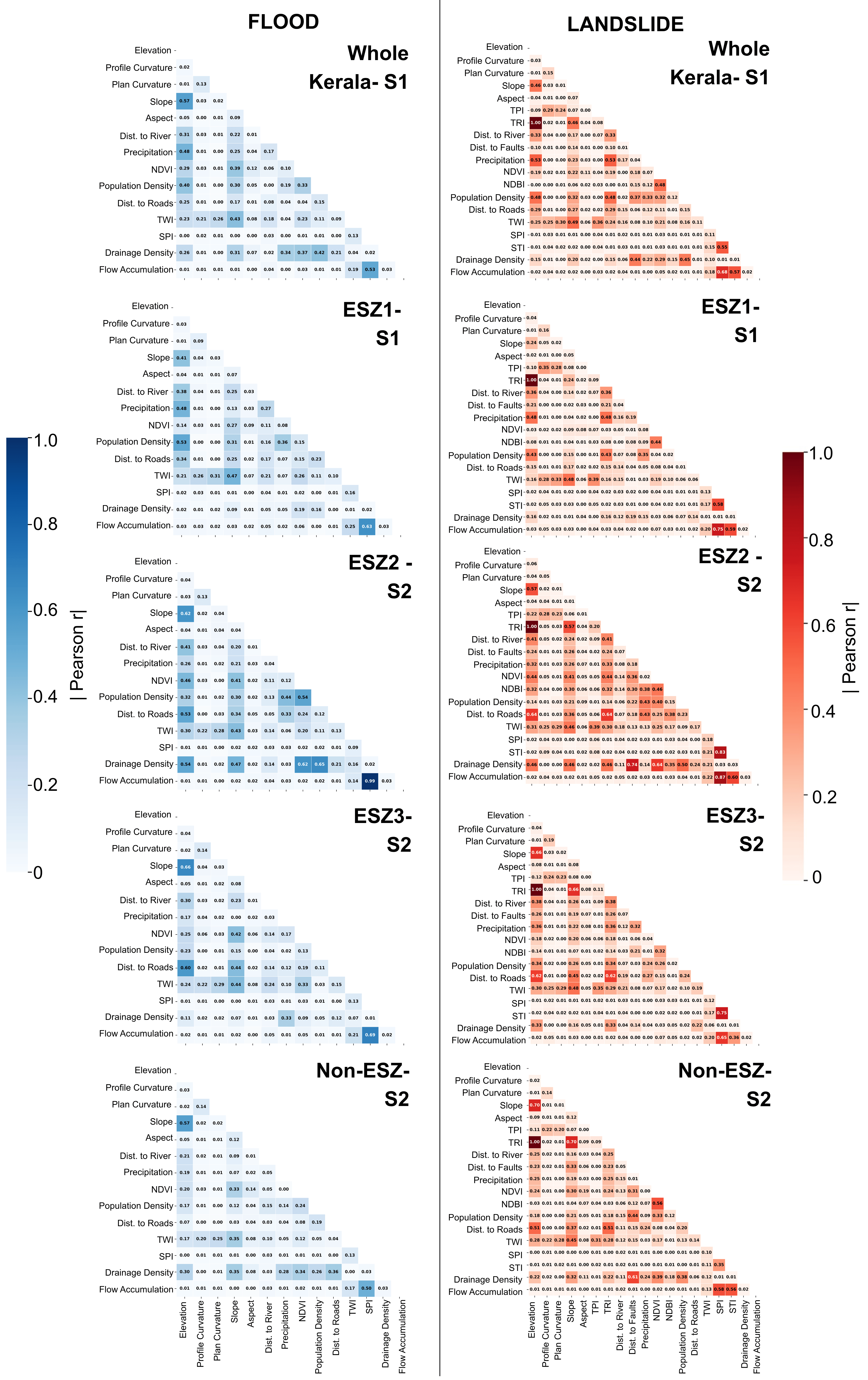}
    \caption{Pearson correlation matrices of continuous candidate predictors across Kerala spatial scopes.}
    \label{fig:corr_matrices_K}
\end{figure}

\begin{figure}[!p]
    \centering
    \includegraphics[
        width=0.8\textwidth,
        height=\textheight,
        keepaspectratio
    ]{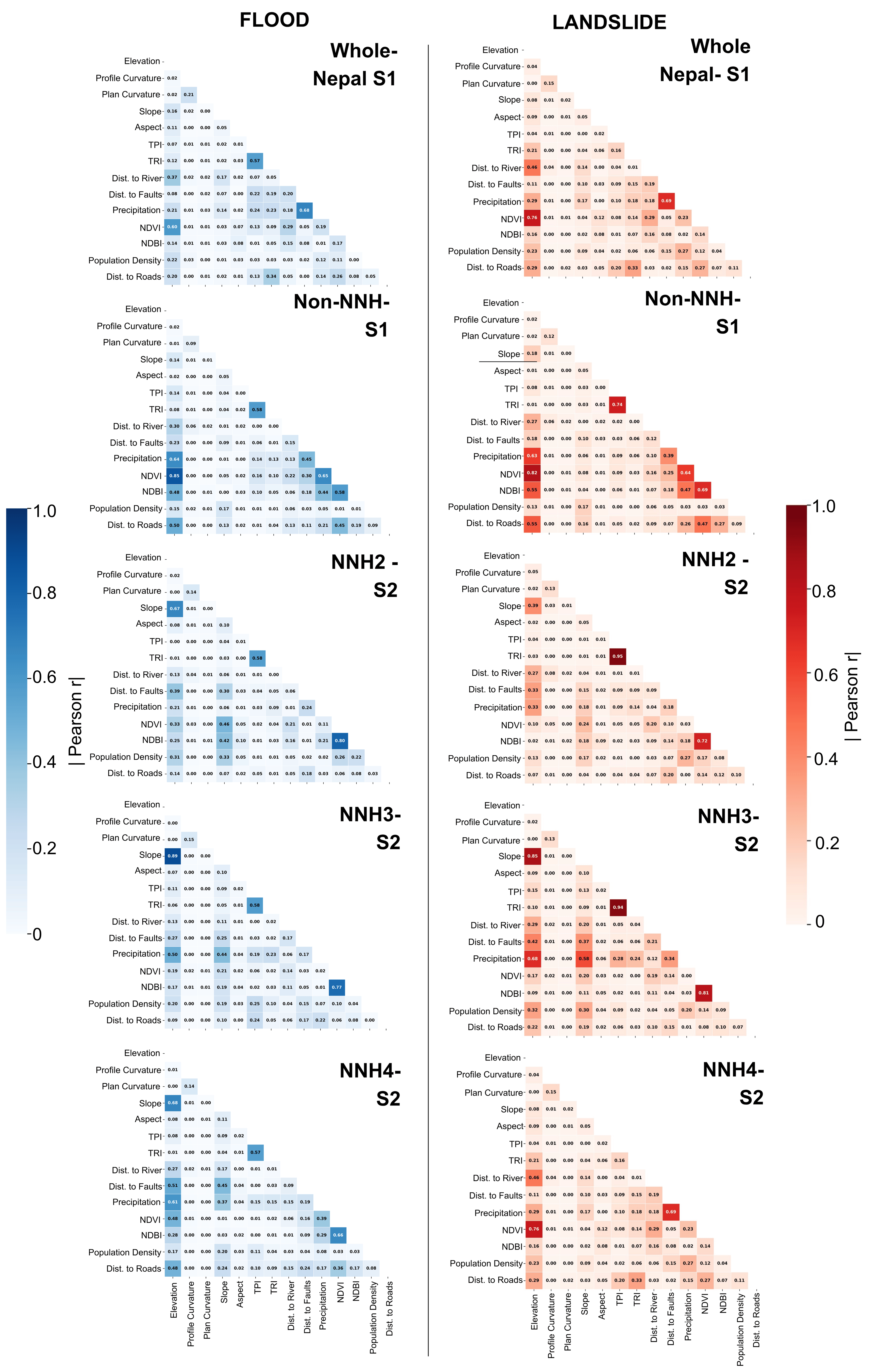}
    \caption{Pearson correlation matrices of continuous candidate predictors across Nepal spatial scopes.}
    \label{fig:corr_matrices_N}
\end{figure}

As described in Section~\ref{sec:predictor_screening}, predictor screening yields different conditioning-factor sets under S1 and S2 using correlation and VIF analyses, as shown in Figures~\ref{fig:corr_matrices_K} and~\ref{fig:corr_matrices_N} for Kerala and Nepal, respectively. This involves screening the entire region for S1 and specific contextual zones for S2.

The correlation and VIF results for floods and landslides are presented in Table~\ref{tab:kerala_vif} for Kerala, and Table~\ref{tab:nepal_vif} for Nepal. The retained factors are summarized in Table~\ref{tab:feature_selection_outcomes}.

In Kerala, profile curvature, plan curvature, NDVI, and NDBI are not retained in any final predictor set. Slope, flow direction, soil type, lithology, and LULC form a common core retained for both hazards across the whole-state and contextual-zone-specific scopes. Elevation is consistently retained only for flood modelling, whereas TRI and distance to faults are retained only for landslide modelling. The principal zone-specific differences involve distance to rivers, drainage density, precipitation, population density, and distance to roads. TWI and SPI are retained for flood modelling in individual ESZs, while aspect is retained for landslide modelling only in ESZ2.

Nepal shows greater consistency between the whole-country and NNH-specific predictor sets. Profile curvature, plan curvature, TPI, and NDVI are not retained in any final set, while slope, aspect, distance to rivers, distance to faults, precipitation, LULC, population density, and distance to roads are retained for both hazards across all spatial scopes. Zone-specific variation is confined mainly to elevation, TRI, and NDBI. Overall, the Kerala predictor sets exhibit broader variation between hazards and contextual zones, particularly among hydrological, climatic, and anthropogenic factors, whereas the Nepal sets retain a more stable predictor core.

\subsection{Model Performance Comparison} \label{sec:model_performance_comparison}
Across both study regions and hazards, S1 achieves higher mean accuracy, precision, recall, F1-score, AUC--ROC, and PR--AUC than S2, as shown in Table~\ref{tab:model_performance_comparison}. The Brier-score pattern differs between the study regions: S1 produces slightly lower values in Kerala, whereas S2 produces lower values in Nepal.

In Kerala, S1 records higher mean performance for both hazards. For flood susceptibility, the largest differences occur in precision, F1-score, AUC--ROC, and PR--AUC. For landslide susceptibility, the improvements are most evident in precision, recall, and F1-score, while the differences in accuracy and AUC--ROC are smaller. The Brier scores differ only slightly between the strategies, with S1 producing lower values for both hazards.

The greatest separation between the strategies occurs for flood susceptibility in Nepal, where S1 substantially improves precision, F1-score, AUC--ROC, and PR--AUC. Landslide susceptibility shows lower overall performance than flood susceptibility under both strategies. Recall remains high, but the lower precision and AUC-based metrics indicate weaker distinction between occurrence and non-occurrence samples. S2 nevertheless produces lower mean Brier scores for both Nepal hazards. Thus, S1 provides stronger classification and discrimination performance, whereas S2 provides better probabilistic accuracy in Nepal.

In what follows, explanability of model performance and downstream computation of risk maps using the computed sustainability maps are demonstrated.

\FloatBarrier

\FloatBarrier

\begin{table}[!t]
\centering
\caption{VIF values of conditioning factors for hazard susceptibility maps in Kerala.}
\label{tab:kerala_vif}
\centering
\scriptsize
\setlength{\tabcolsep}{3pt}
\renewcommand{\arraystretch}{1.10}

\begin{tabular}{@{}llccccc@{}}
\toprule
\textbf{Factor Group}
& \textbf{Conditioning Factor}
& \shortstack{\textbf{Whole}\\\textbf{Kerala}}
& \textbf{ESZ1}
& \textbf{ESZ2}
& \textbf{ESZ3}
& \shortstack{\textbf{Non-}\\\textbf{ESZ}} \\
\midrule

\multicolumn{7}{c}{\textbf{\textit{Flood Susceptibility Map}}} \\
\midrule

\multirow{7}{*}{Topographic}
& Elevation
& 3.15 & 4.83 & 7.44 & 3.89 & 2.71 \\

& Profile curvature
& \underline{3216.24}
& \underline{5505.48}
& \underline{2788.78}
& \underline{3496.23}
& \underline{2623.10} \\

& Plan curvature
& \underline{3319.39}
& \underline{5759.18}
& \underline{2746.84}
& \underline{3607.16}
& \underline{2690.87} \\

& Slope
& 3.63 & 4.16 & 4.60 & 4.42 & 3.55 \\

& Aspect
& 3.62 & 3.84 & 3.62 & 3.80 & 3.48 \\

& TPI
& -- & -- & -- & -- & -- \\

& TRI
& -- & -- & -- & -- & -- \\

\midrule

\multirow{6}{*}{Hydrological}
& Distance to rivers
& 2.91 & 3.31 & 3.81 & 3.08 & 2.72 \\

& TWI
& 9.73 & 9.41 & 10.15 & 9.15 & 10.82 \\

& Flow accumulation
& 1.43 & 1.75 & -- & 1.99 & 1.35 \\

& Drainage density
& 8.13 & 9.05 & 9.44 & \underline{29.83} & 5.61 \\

& SPI
& 1.39 & 1.67 & 1.02 & 1.92 & 1.33 \\

& STI
& -- & -- & -- & -- & -- \\

\midrule

Geological
& Distance to faults
& -- & -- & -- & -- & -- \\

\midrule

Climatic
& Precipitation
& 9.01 & 9.40 & 9.57 & 9.36 & 9.20 \\

\midrule

\multirow{2}{*}{\shortstack[l]{Surface\\characteristics}}
& NDVI
& \underline{66.56}
& \underline{109.35}
& \underline{140.93}
& \underline{78.62}
& \underline{53.39} \\

& NDBI
& -- & -- & -- & -- & -- \\

\midrule

\multirow{2}{*}{Anthropogenic}
& Population density
& 4.10 & 6.02 & 9.49 & 8.22 & 4.85 \\

& Distance to roads
& 2.20 & 2.09 & 2.99 & 3.00 & 2.52 \\

\midrule

\multicolumn{7}{c}{\textbf{\textit{Landslide Susceptibility Map}}} \\
\midrule

\multirow{7}{*}{Topographic}
& Elevation
& -- & -- & -- & -- & -- \\

& Profile curvature
& \underline{4622.93}
& \underline{8842.32}
& \underline{6476.51}
& \underline{5826.00}
& \underline{4622.93} \\

& Plan curvature
& \underline{4013.21}
& \underline{8125.00}
& \underline{5040.95}
& \underline{5274.38}
& \underline{4013.21} \\

& Slope
& 4.79 & 5.26 & 4.96 & 5.92 & 4.79 \\

& Aspect
& 3.85 & 3.84 & 3.67 & 4.18 & 3.85 \\

& TPI
& \underline{545.17}
& \underline{290.48}
& \underline{757.36}
& \underline{330.72}
& \underline{545.17} \\

& TRI
& 3.28 & 6.84 & 6.21 & 5.39 & 3.28 \\

\midrule

\multirow{6}{*}{Hydrological}
& Distance to rivers
& 3.21 & 3.55 & 4.25 & 3.90 & 3.21 \\

& TWI
& 10.51
& 10.35
& \underline{13.42}
& 10.11
& 10.51 \\

& Flow accumulation
& 1.93 & 2.54 & 5.68 & 1.97 & 1.93 \\

& Drainage density
& \underline{13.42}
& \underline{46.40}
& 9.60
& \underline{39.28}
& 9.42 \\

& SPI
& 1.51 & 2.45 & 11.35 & 3.81 & 1.51 \\

& STI
& 1.46 & 1.65 & 4.23 & 2.55 & 1.46 \\

\midrule

Geological
& Distance to faults
& 5.30 & 2.66 & 9.06 & 4.52 & 5.30 \\

\midrule

Climatic
& Precipitation
& 8.53
& \underline{30.33}
& \underline{23.80}
& \underline{26.97}
& 8.76 \\

\midrule

\multirow{2}{*}{\shortstack[l]{Surface\\characteristics}}
& NDVI
& \underline{90.14}
& \underline{167.29}
& \underline{216.45}
& \underline{191.19}
& \underline{90.14} \\

& NDBI
& \underline{96.85}
& \underline{105.39}
& \underline{78.32}
& \underline{110.69}
& \underline{96.85} \\

\midrule

\multirow{2}{*}{Anthropogenic}
& Population density
& 5.25
& 5.17
& \underline{53.17}
& 6.43
& 5.25 \\

& Distance to roads
& 2.65 & 2.18 & 4.15 & 3.52 & 2.65 \\

\bottomrule
\end{tabular}

\vspace{3pt}

{\footnotesize
\textit{Note:} -- indicates values excluded during the Pearson correlation stage before consideration for the VIF stage, and the \underline{underlined} values exceed the adopted VIF threshold of 13. 
\par}
\end{table}

\begin{table}[ht]
\centering
\caption{VIF values of conditioning factors for hazard susceptibility maps in Nepal.}
\label{tab:nepal_vif}
\centering
\scriptsize
\setlength{\tabcolsep}{3pt}
\renewcommand{\arraystretch}{1.10}

\begin{tabular}{@{}llccccc@{}}
\toprule
\textbf{Factor Group}
& \textbf{Conditioning Factor}
& \shortstack{\textbf{Whole}\\\textbf{Nepal}}
& \textbf{Non-NNH}
& \textbf{NNH2}
& \textbf{NNH3}
& \textbf{NNH4} \\
\midrule

\multicolumn{7}{c}{\textbf{\textit{Flood Susceptibility Map}}} \\
\midrule

\multirow{7}{*}{Topographic}
& Elevation
& 7.56
& \vifhigh{221.52}
& \vifhigh{53.84}
& 6.35
& 7.56 \\

& Profile curvature
& \vifhigh{13814.49}
& \vifhigh{46583.82}
& \vifhigh{63640.95}
& \vifhigh{24114.51}
& \vifhigh{13814.49} \\

& Plan curvature
& \vifhigh{12533.61}
& \vifhigh{44721.86}
& \vifhigh{60242.81}
& \vifhigh{21206.99}
& \vifhigh{12533.61} \\

& Slope
& 6.52
& 4.74
& 6.44
& 5.70
& 6.52 \\

& Aspect
& 3.37
& 4.47
& 4.11
& 3.87
& 3.37 \\

& TPI
& \vifhigh{31358.64}
& \vifhigh{110683.06}
& \vifhigh{137711.54}
& \vifhigh{52824.27}
& \vifhigh{31358.64} \\

& TRI
& 2.77
& 2.04
& 3.43
& 5.58
& 2.77 \\

\midrule

Hydrological
& Distance to rivers
& 2.99
& 4.74
& 3.80
& 3.05
& 2.99 \\

\midrule

Geological
& Distance to faults
& 3.56
& 10.70
& 5.11
& 3.81
& 3.56 \\

\midrule

Climatic
& Precipitation
& \vifhigh{32.24}
& \vifhigh{28.55}
& \vifhigh{16.05}
& \vifhigh{33.39}
& \vifhigh{32.23} \\

\midrule

\multirow{2}{*}{\shortstack[l]{Surface\\characteristics}}
& NDVI
& \vifhigh{260.64}
& \vifhigh{163.41}
& \vifhigh{149.17}
& \vifhigh{247.70}
& \vifhigh{260.64} \\

& NDBI
& \vifhigh{337.78}
& \vifhigh{13.74}
& \vifhigh{43.53}
& \vifhigh{250.33}
& \vifhigh{337.78} \\

\midrule

\multirow{2}{*}{Anthropogenic}
& Population density
& 1.63
& 2.90
& 2.56
& 1.87
& 1.63 \\

& Distance to roads
& 2.06
& 4.43
& 2.81
& 2.17
& 2.06 \\

\midrule

\multicolumn{7}{c}{\textbf{\textit{Landslide Susceptibility Map}}} \\
\midrule

\multirow{7}{*}{Topographic}
& Elevation
& \underline{13.78}
& \underline{240.49}
& \underline{36.75}
& 8.17
& \underline{13.05} \\

& Profile curvature
& \underline{49562.20}
& \underline{61213.10}
& \underline{87531.19}
& \underline{54343.12}
& \underline{13589.98} \\

& Plan curvature
& \underline{43067.44}
& \underline{63829.25}
& \underline{73287.57}
& \underline{47421.80}
& \underline{12644.79} \\

& Slope
& 6.42
& 5.14
& 7.36
& 6.88
& 7.82 \\

& Aspect
& 4.01
& 4.52
& 4.06
& 4.14
& 3.75 \\

& TPI
& \underline{128718.36}
& \underline{152844.50}
& \underline{209742.00}
& \underline{161414.40}
& -- \\

& TRI
& 2.36
& 2.12
& 3.44
& --
& 1.96 \\

\midrule

Hydrological
& Distance to rivers
& 3.54
& 4.94
& 3.85
& 3.48
& 3.60 \\

\midrule

Geological
& Distance to faults
& 3.68
& 12.09
& 6.29
& 4.01
& 2.93 \\

\midrule

Climatic
& Precipitation
& \underline{18.57}
& \underline{31.89}
& \underline{18.20}
& \underline{23.63}
& \underline{21.17} \\

\midrule

\multirow{2}{*}{\shortstack[l]{Surface\\characteristics}}
& NDVI
& \underline{154.67}
& \underline{155.76}
& \underline{159.84}
& \underline{237.18}
& \underline{208.13} \\

& NDBI
& \underline{76.17}
& \underline{14.59}
& \underline{56.62}
& \underline{187.48}
& \underline{179.53} \\

\midrule

\multirow{2}{*}{Anthropogenic}
& Population density
& 1.65
& 1.85
& 2.70
& 1.78
& 1.38 \\

& Distance to roads
& 2.55
& 4.42
& 2.70
& 2.23
& 2.36 \\

\bottomrule
\end{tabular}

\vspace{3pt}

{\footnotesize
\textit{Note:} -- indicates values excluded during the Pearson correlation stage before consideration for the VIF stage, and the \underline{underlined} values exceed the adopted VIF threshold of 13.
\par}
\end{table}

\begin{table}[htp]
\caption{Conditioning factors retained for hazard susceptibility map generation in the study areas of Kerala and Nepal.}
\label{tab:feature_selection_outcomes}
\centering
\scriptsize
\setlength{\tabcolsep}{2pt}
\renewcommand{\arraystretch}{1.10}

\begin{tabular*}{\linewidth}{@{\extracolsep{\fill}}lcccccccccc@{}}
\toprule
& \multicolumn{5}{c}{\textbf{Kerala}}
& \multicolumn{5}{c}{\textbf{Nepal}} \\
\cmidrule(lr){2-6}
\cmidrule(lr){7-11}

\bf Conditioning Factor
& \rotatebox{90}{\textbf{Whole}}
& \rotatebox{90}{\textbf{ESZ1}}
& \rotatebox{90}{\textbf{ESZ2}}
& \rotatebox{90}{\textbf{ESZ3}}
& \rotatebox{90}{\textbf{Non-ESZ}}
& \rotatebox{90}{\textbf{Whole}}
& \rotatebox{90}{\textbf{Non-NNH}}
& \rotatebox{90}{\textbf{NNH2}}
& \rotatebox{90}{\textbf{NNH3}}
& \rotatebox{90}{\textbf{NNH4}} \\
\midrule \\

\multicolumn{11}{@{}l}{\textit{Topographic}} \\

Elevation
& F & F & F & F & F
& B & -- & L & B & B \\

Profile curvature
& -- & -- & -- & -- & --
& -- & -- & -- & -- & -- \\

Plan curvature
& -- & -- & -- & -- & --
& -- & -- & -- & -- & -- \\

Slope
& B & B & B & B & B
& B & B & B & B & B \\

Aspect
& -- & -- & L & -- & --
& B & B & B & B & B \\

TPI
& -- & -- & -- & -- & --
& -- & -- & -- & -- & -- \\

TRI
& L & L & L & L & L
& B & B & B & F & B \\

\midrule \\

\multicolumn{11}{@{}l}{\textit{Hydrological}} \\

TWI
& -- & -- & -- & F & --
& NA & NA & NA & NA & NA \\

Flow accumulation
& -- & -- & -- & -- & --
& NA & NA & NA & NA & NA \\

Flow direction
& B & B & B & B & B
& NA & NA & NA & NA & NA \\

Drainage density
& F & F & B & -- & B
& NA & NA & NA & NA & NA \\

SPI
& -- & F & -- & -- & --
& NA & NA & NA & NA & NA \\

Distance to rivers
& B & L & B & B & F
& B & B & B & B & B \\

STI
& -- & -- & -- & -- & --
& NA & NA & NA & NA & NA \\

\midrule \\

\multicolumn{11}{@{}l}{\textit{Geological and soil}} \\

Soil type
& B & B & B & B & B
& NA & NA & NA & NA & NA \\

Lithology
& B & B & B & B & B
& NA & NA & NA & NA & NA \\

Distance to faults
& L & L & L & L & L
& B & B & B & B & B \\

\midrule \\

\multicolumn{11}{@{}l}{\textit{Climatic}} \\

Precipitation
& B & F & F & F & B
& B & B & B & B & B \\

\midrule \\

\multicolumn{11}{@{}l}{\shortstack[l]{\textit{Land-surface}\\\textit{characteristics}}} \\

LULC
& B & B & B & B & B
& B & B & B & B & B \\

NDVI
& -- & -- & -- & -- & --
& -- & -- & -- & -- & -- \\

NDBI
& -- & -- & -- & -- & --
& -- & B & F & L & -- \\

\midrule \\

\multicolumn{11}{@{}l}{\textit{Anthropogenic}} \\

Population density
& B & B & F & B & B
& B & B & B & B & B \\

Distance to roads
& -- & L & B & L & --
& B & B & B & B & B \\

\bottomrule
\end{tabular*}

\vspace{2pt}

{\footnotesize 
\textit{Note:} B, F, and L indicate retention for both hazards, flood only, and landslide only, respectively; -- indicates non-retention and NA indicates that the factor is not considered.
\par}
\end{table}

\clearpage

\begin{table}[htp]
\caption{Mean predictive performance across the strategy-specific evaluation units under proposed training strategies, S1 and S2, for flood and landslide susceptibility modelling in Kerala and Nepal.}
\label{tab:model_performance_comparison}
\centering
\scriptsize
\setlength{\tabcolsep}{5pt}
\renewcommand{\arraystretch}{1.15}

\begin{tabular*}{\linewidth}{@{\extracolsep{\fill}}llcccc@{}}
\toprule
\textbf{Region} & \textbf{Metric}
& \multicolumn{2}{c}{\textbf{Flood}}
& \multicolumn{2}{c}{\textbf{Landslide}} \\
\cmidrule(lr){3-4}
\cmidrule(lr){5-6}
& & \textbf{S1} & \textbf{S2}
& \textbf{S1} & \textbf{S2} \\
\midrule

\multirow{7}{*}{\bf Kerala}
& Accuracy $\uparrow$
& \textbf{0.683} & 0.634
& \textbf{0.747} & 0.720 \\

& Precision $\uparrow$
& \textbf{0.628} & 0.490
& \textbf{0.682} & 0.558 \\

& Recall $\uparrow$
& \textbf{0.899} & 0.827
& \textbf{0.924} & 0.815 \\

& F1-score $\uparrow$
& \textbf{0.739} & 0.605
& \textbf{0.785} & 0.650 \\

& Brier score $\downarrow$
& \textbf{0.199} & 0.203
& \textbf{0.182} & 0.192 \\

& AUC--ROC $\uparrow$
& \textbf{0.789} & 0.658
& \textbf{0.792} & 0.732 \\

& PR--AUC $\uparrow$
& \textbf{0.778} & 0.530
& \textbf{0.731} & 0.597 \\

\midrule

\multirow{7}{*}{\bf Nepal}
& Accuracy $\uparrow$
& \textbf{0.857} & 0.789
& \textbf{0.558} & 0.468 \\

& Precision $\uparrow$
& \textbf{0.811} & 0.503
& \textbf{0.533} & 0.400 \\

& Recall $\uparrow$
& \textbf{0.932} & 0.697
& \textbf{0.960} & 0.930 \\

& F1-score $\uparrow$
& \textbf{0.867} & 0.559
& \textbf{0.685} & 0.526 \\

& Brier score $\downarrow$
& 0.194 & \textbf{0.179}
& 0.250 & \textbf{0.229} \\

& AUC--ROC $\uparrow$
& \textbf{0.886} & 0.728
& \textbf{0.638} & 0.555 \\

& PR--AUC $\uparrow$
& \textbf{0.823} & 0.512
& \textbf{0.616} & 0.416 \\

\bottomrule
\end{tabular*}

\vspace{3pt}
{\footnotesize
\textit{Note:} \textbf{Boldface} indicates the better-performing strategy for each region, hazard, and metric.
\par}

\end{table}

\subsection{SHAP-based Interpretation of Susceptibility Models} \label{sec:shap_analysis}
\begin{figure}[htp]
    \centering
    \includegraphics[
        width=\linewidth,
    ]{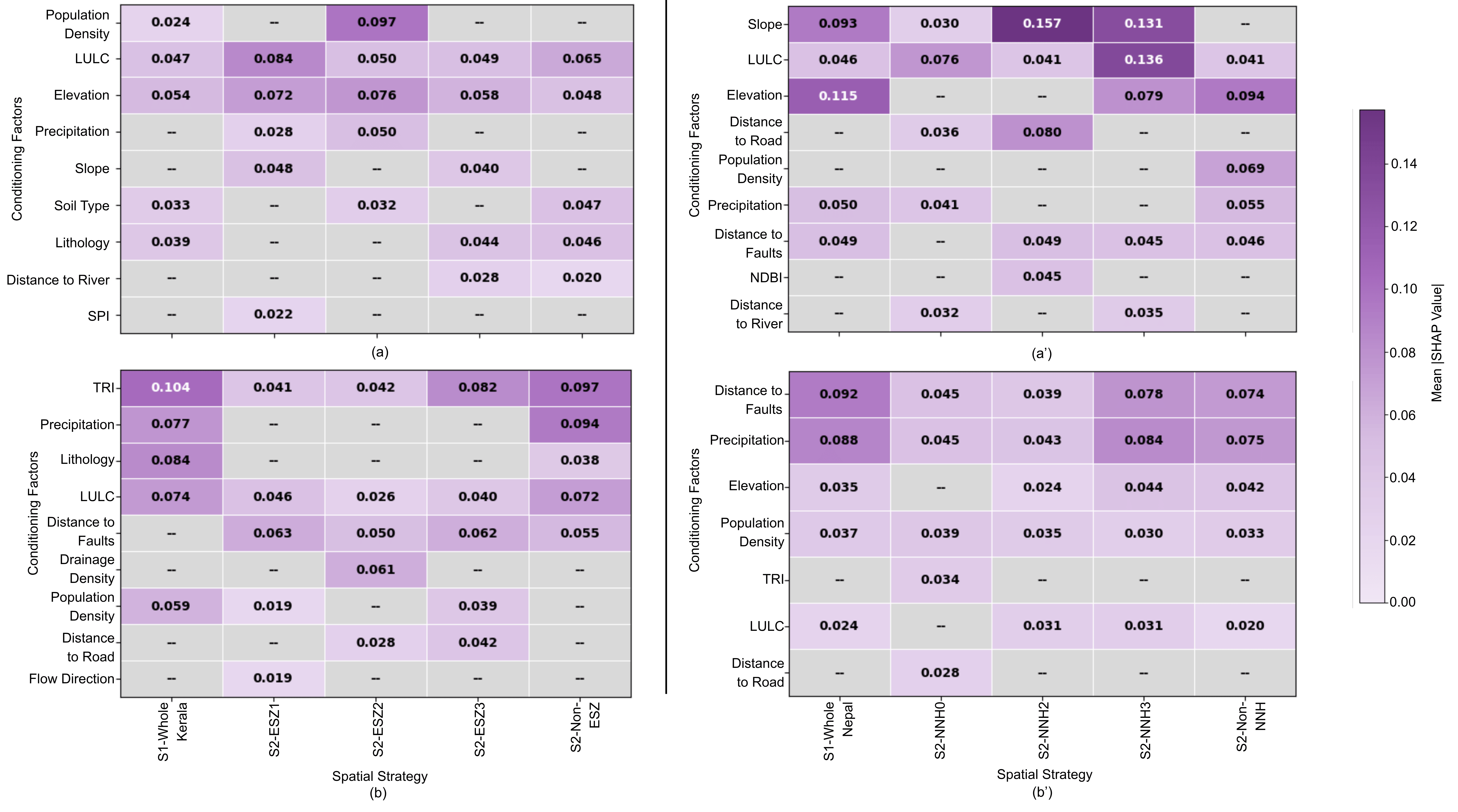}
    \caption{Top-five mean absolute SHAP values for the flood and landslide susceptibility models in Kerala and Nepal for the best optimal strategy between S1 and S2. Panels (a) and (b) show the Kerala flood and landslide models, respectively, while panels (a$'$) and (b$'$) show the Nepal flood and landslide models, respectively. Grey cells indicate that the predictor is not ranked among the top-five SHAP factors for the corresponding susceptibility mapping model.}
    \label{fig:shap}
\end{figure}

SHAP analysis reveals differences in the contributions and rankings of the retained conditioning factors between S1 and S2, as shown in Figure~\ref{fig:shap}. Kerala exhibits greater variation in the leading predictors across the S2 contextual zones, whereas the Nepal landslide models retain a more consistent importance pattern.

For Kerala flood susceptibility, elevation has the highest mean absolute SHAP value under S1. Under S2, LULC ranks first in ESZ1 and Non-ESZ, population density in ESZ2, and elevation in ESZ3. Elevation and LULC are the only factors appearing among the five highest-ranked predictors in every Kerala flood configuration. For landslide susceptibility, TRI ranks first under S1 and remains the leading predictor in ESZ3 and Non-ESZ. Distance to faults ranks first in ESZ1, whereas drainage density ranks first in ESZ2. TRI and LULC occur among the five highest-ranked predictors in all Kerala landslide configurations.

For Nepal flood susceptibility, elevation ranks first under S1 and NNH4, LULC in Non-NNH and NNH3, and slope in NNH2. The mean absolute SHAP value of slope in NNH2 is the largest among the Nepal flood configurations. Nepal landslide susceptibility shows less variation in predictor ranking, with distance to faults and precipitation occupying the two highest ranks under S1 and across the S2 configurations. Population density also occurs among the five leading predictors in every Nepal landslide configuration.

Several predictors retained despite exceeding the adopted VIF threshold remain influential in the fitted models. Precipitation ranks among the two leading predictors in every Nepal landslide configuration, while elevation occurs among the five leading predictors in the whole-Nepal, NNH2, and NNH4 landslide models. NDBI appears among the five leading predictors only for flood susceptibility in NNH2, indicating a more spatially restricted contribution.

\begin{figure}[hp]
    \centering
    \includegraphics[
        width=0.7\linewidth,
    ]{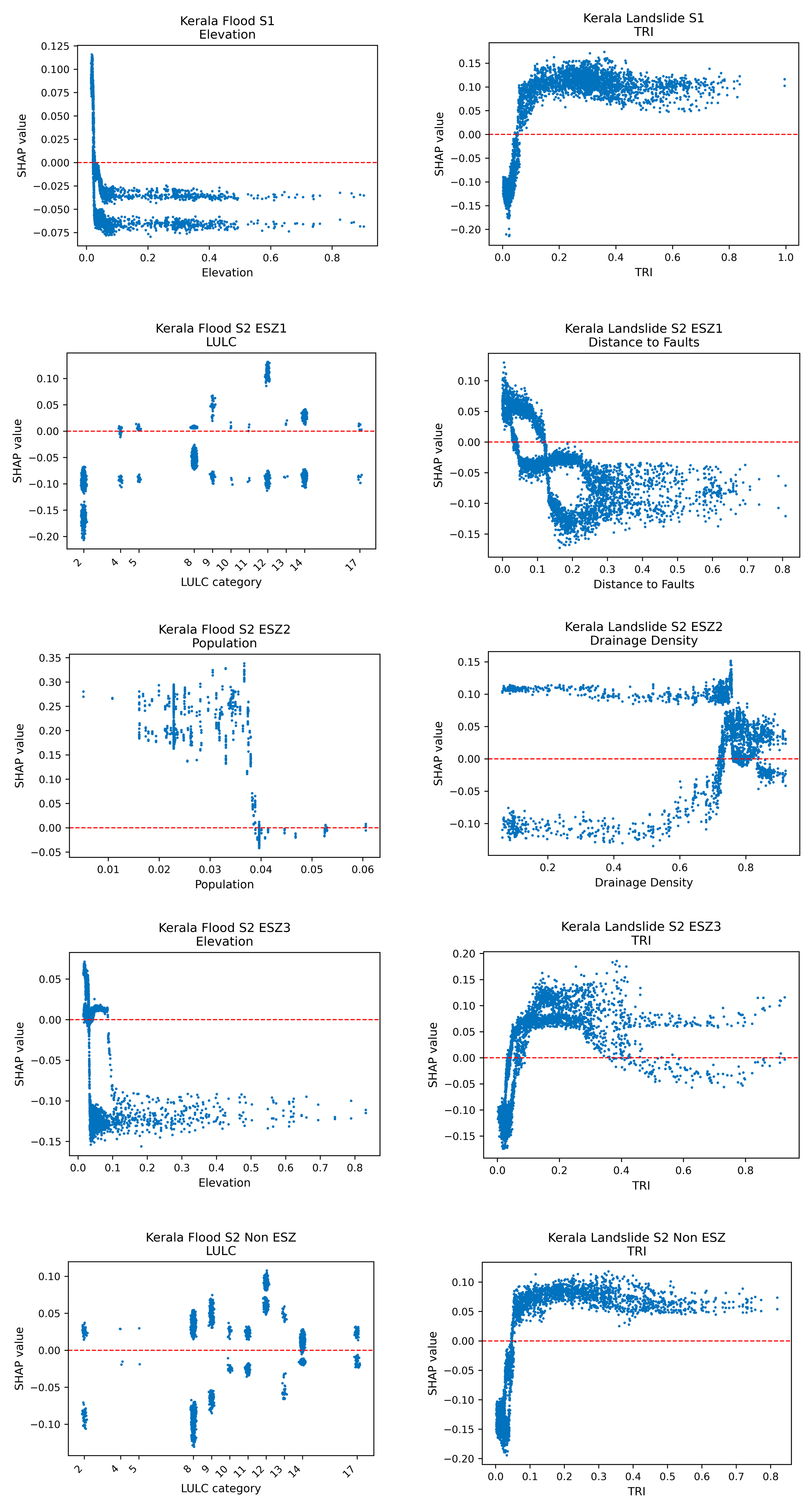}
    \caption{SHAP dependence plots for the dominant predictor in each susceptibility map in Kerala for (entire area, S1) and (zone-wise, S2 models) for (Left) flood and (Right) landslide, with zero SHAP contribution as a red dashed line.}
    \label{fig:kerala_shap_dependence}
\end{figure}

\begin{figure}[hp]
    \centering
    \includegraphics[
        width=0.7\linewidth,
    ]{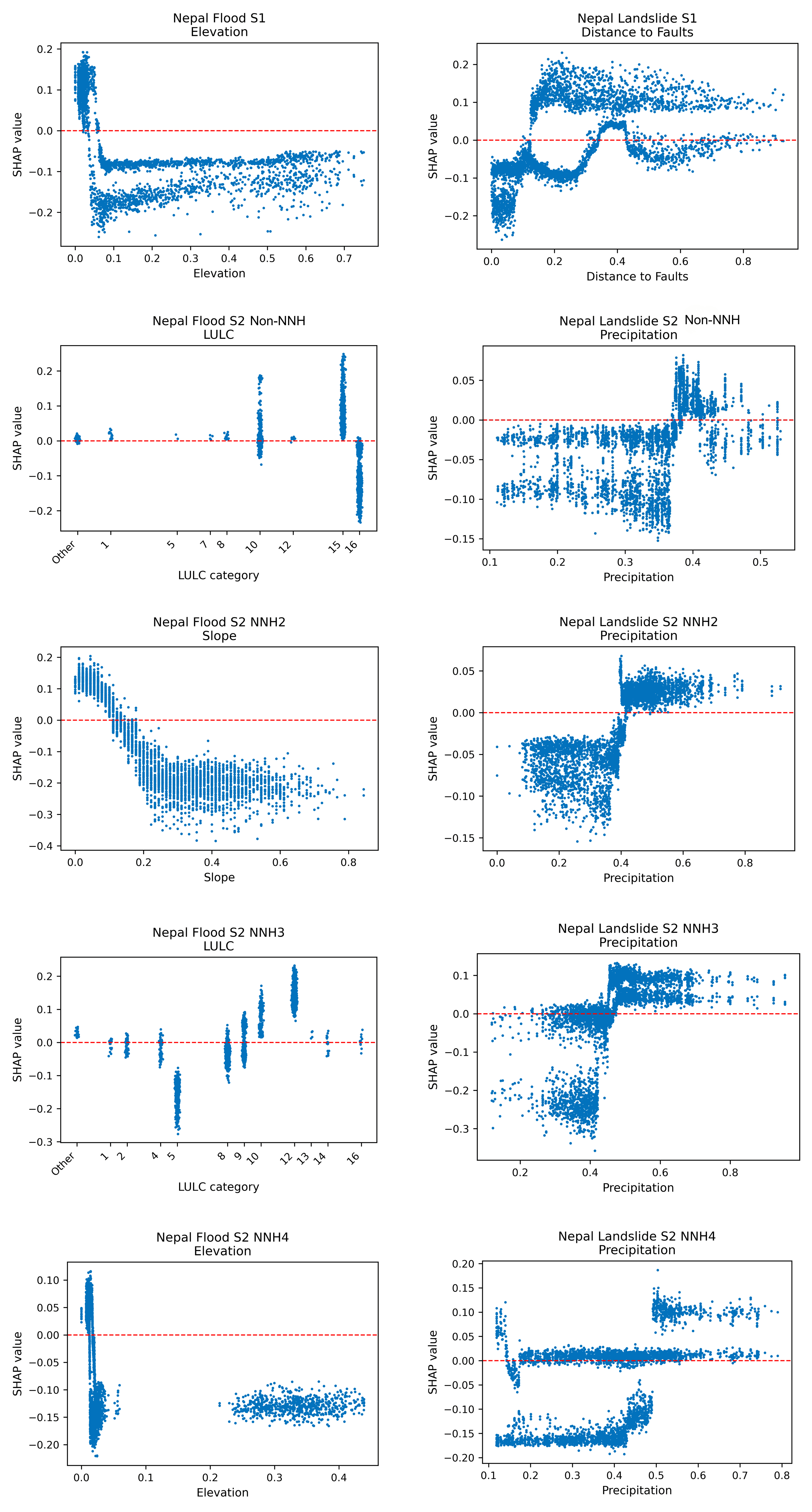}
    \caption{SHAP dependence plots for the dominant predictor in each susceptibility map in Nepal for (entire area, S1) and (zone-wise, S2 models) for (Left) flood and (Right) landslide, with zero SHAP contribution as a red dashed line.}
    \label{fig:nepal_shap_dependence}
\end{figure}

The dependence plots reveal nonlinear, threshold-like, and category-specific relationships between the leading predictors and model output, as shown in Figures~\ref{fig:kerala_shap_dependence} and~\ref{fig:nepal_shap_dependence} for Kerala and Nepal, respectively. For Kerala flood susceptibility, elevation under S1 and ESZ3 has positive contributions at the lowest observed values and predominantly negative contributions at higher values. Among the MODIS land-cover classes, croplands have the largest positive contributions in ESZ1 and Non-ESZ. In ESZ2, population density has positive contributions over the lower observed range and near-zero or negative contributions at higher values.

For Kerala landslide susceptibility, low TRI values have negative contributions under S1, ESZ3, and Non-ESZ, followed by a transition to positive contributions at low-to-intermediate values. The contributions generally stabilize or decline at higher TRI values, although ESZ3 exhibits a broader mixture of positive and negative responses. In ESZ1, lower values of distance to faults have positive contributions, whereas higher values are predominantly negative. Drainage density in ESZ2 exhibits a non-monotonic pattern, with alternating positive and negative contributions across the observed range.

For Nepal flood susceptibility, elevation under S1 and NNH4 has positive contributions at the lowest observed values and negative contributions across most of the higher range. In NNH2, low slope values have positive contributions, followed by negative contributions at intermediate and higher values. The LULC responses vary among the S2 zones. Permanent snow and ice have the largest positive contributions in Non-NNH, whereas barren land is predominantly associated with negative contributions. In NNH3, croplands have the largest positive contributions, while mixed forests show predominantly negative contributions.

For Nepal landslide susceptibility, distance to faults under S1 has negative contributions at lower observed values, positive contributions at intermediate values, and weaker or mixed contributions at higher values. Under S2, precipitation generally shifts from negative to positive contributions across the intermediate observed range in Non-NNH, NNH2, and NNH3. In NNH4, precipitation responses form positive, negative, and near-zero bands across the observed range.

\subsection{Spatial Distribution of Multi-hazard Susceptibility and Risk Maps} \label{sec:spatial_distribution_multihazard}

\begin{figure}[p]
    \centering
    \includegraphics[width=0.8\textwidth]{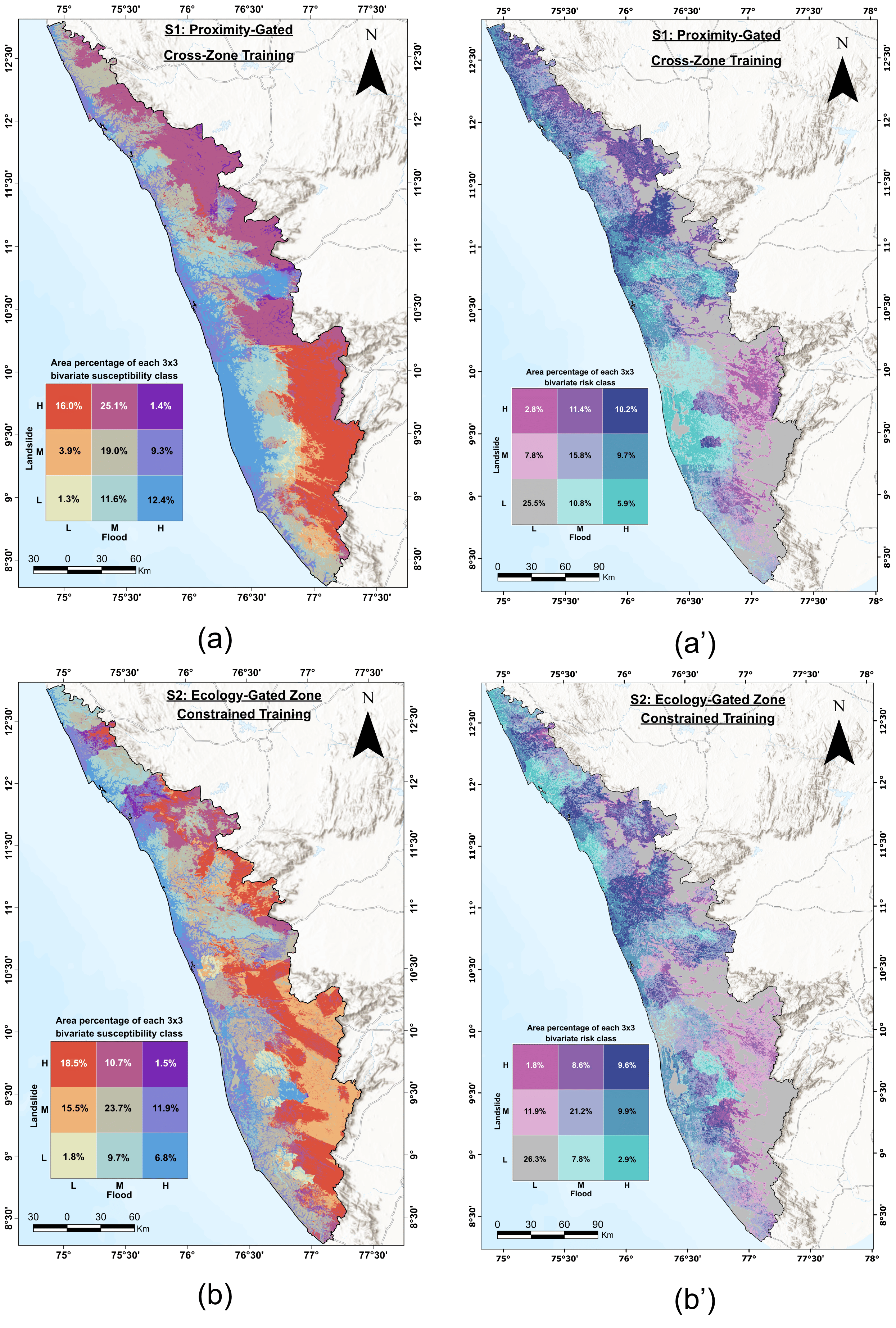}
    \caption{Bivariate flood--landslide susceptibility and risk maps for Kerala under the two spatial training strategies. Panels (a) and (a$'$) show the S1 susceptibility and risk maps, respectively, whereas panels (b) and (b$'$) show the corresponding S2 maps. The inset matrices report the percentage of the study area within each combination of low, medium, and high flood and landslide classes.}
    \label{fig:kerala_bivariate_maps}
\end{figure}

\begin{figure}[p]
    \centering
    \includegraphics[width=0.8\textwidth]{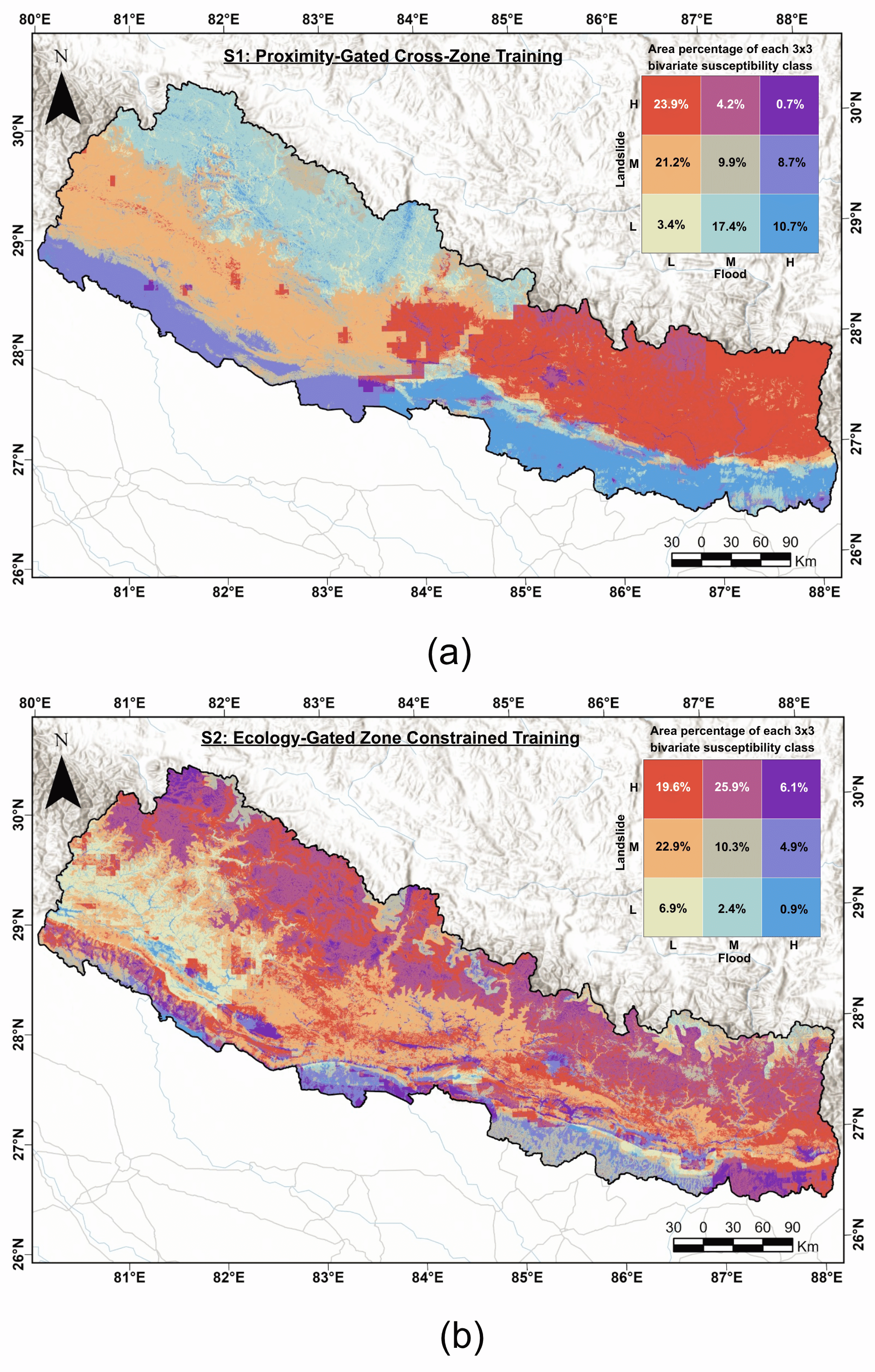}
    \caption{Bivariate flood--landslide susceptibility maps for Nepal under (a) S1 and (b) S2. The inset matrices report the percentage of the study area within each combination of low, medium, and high flood and landslide susceptibility classes.}
    \label{fig:nepal_bivariate_susc}
\end{figure}

\begin{figure}[p]
    \centering
    \includegraphics[width=0.8\textwidth]{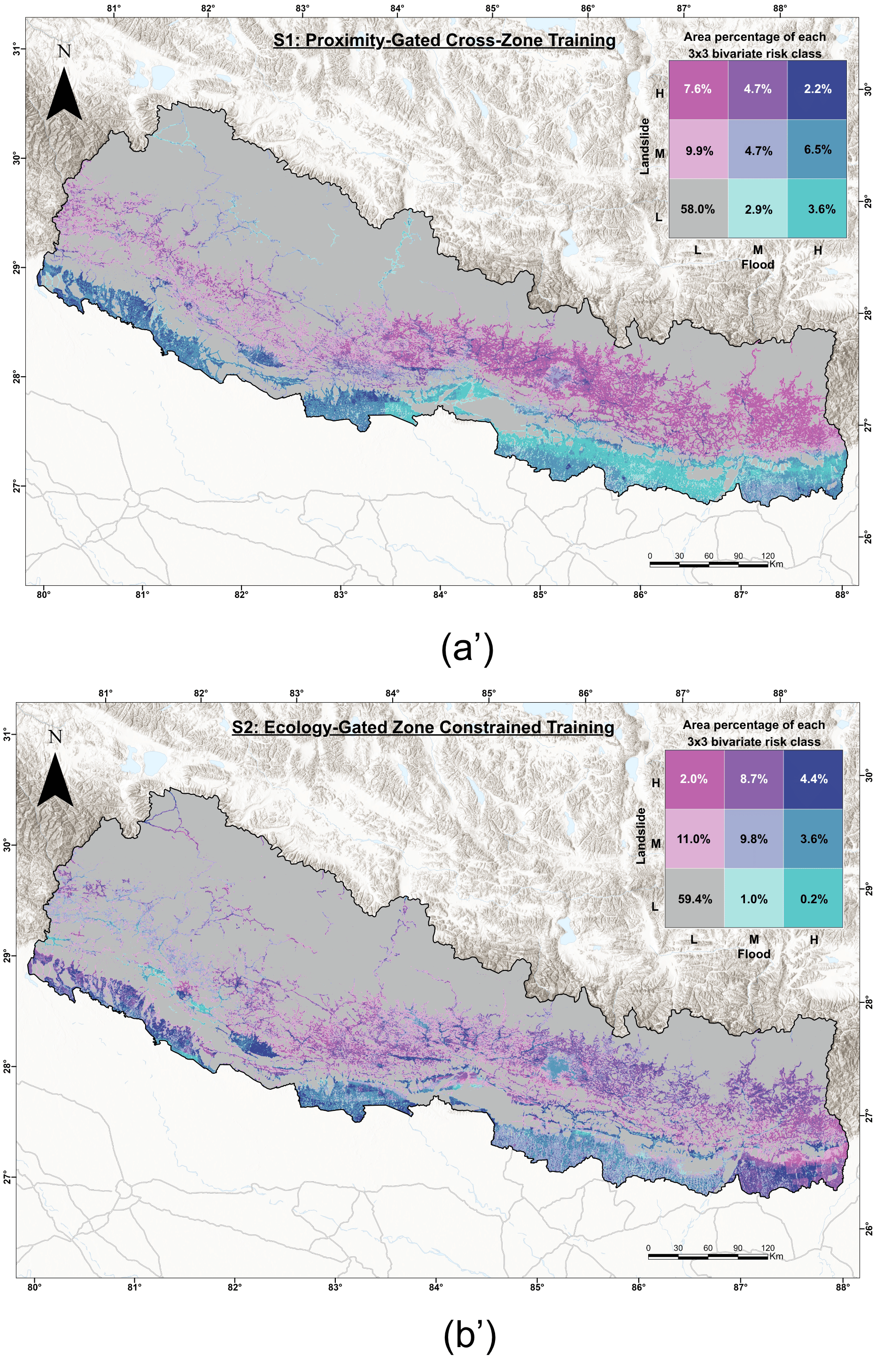}
    \caption{Bivariate flood--landslide risk maps for Nepal under (a$'$) S1 and (b$'$) S2. The inset matrices report the percentage of the study area within each combination of low, medium, and high flood and landslide risk classes.}
    \label{fig:nepal_bivariate_risk}
\end{figure}

The bivariate susceptibility and risk maps are computed using Jenks natural-break thresholds (Section~\ref{sec:multi_hazard_susceptibility_risk_mapping}) and susceptibility--exposure--vulnerability values (Equation~\ref{eq:hazard_risk}), respectively. These maps show contrasting spatial patterns between S1 and S2 for Kerala in Figure~\ref{fig:kerala_bivariate_maps}, and for Nepal in Figures~\ref{fig:nepal_bivariate_susc}, and~\ref{fig:nepal_bivariate_risk}. These results also show unique patterns in Kerala and Nepal. Kerala exhibits a more interspersed distribution of flood- and landslide-dominated combinations, whereas Nepal shows a clearer separation between flood-dominated combinations in the southern lowlands and landslide-dominated combinations across the hill and mountain regions.

In Kerala, medium flood--high landslide susceptibility forms the largest class under S1, covering 25.1\% of the region, followed by medium flood--medium landslide susceptibility at 19.0\%. Landslide-dominated combinations are widespread across the eastern uplands, while high-flood combinations are concentrated mainly in the western lowlands and along river corridors. Under S2, medium flood--medium landslide susceptibility becomes the largest class at 23.7\%, while low flood paired with medium or high landslide susceptibility occupies a larger share than under S1. High-landslide combinations are broadly distributed across the eastern ESZ1 uplands, whereas flood-dominated and intermediate combinations are more extensive within the western Non-ESZ belt. High flood--high landslide susceptibility remains limited under both strategies, accounting for 1.4\% under S1 and 1.5\% under S2.

Following the integration of exposure and vulnerability and classification of the resulting risk surfaces, the Kerala risk maps contain larger areal shares of both low flood--low landslide and high flood--high landslide classes than the susceptibility maps. Low flood--low landslide risk forms the largest class under S1 and S2, covering 25.5\% and 26.3\% of the region, respectively. High flood--high landslide risk accounts for 10.2\% under S1 and 9.6\% under S2, compared with 1.4\% and 1.5\% in the corresponding susceptibility maps. These differences reflect redistribution of the classified combinations following the integration of exposure and vulnerability rather than increases in the continuous risk values relative to the corresponding susceptibility values. Medium flood--medium landslide risk also covers a larger area under S2 than under S1. Areas classified as high for both flood and landslide risk occur mainly as discontinuous clusters within broader low- and medium-risk distributions.

In Nepal, low flood--high landslide susceptibility forms the largest class under S1 at 23.9\%, followed by low flood--medium landslide susceptibility at 21.2\%. These combinations extend across much of the hill and mountain terrain, while high-flood combinations are concentrated mainly in the southern lowlands. Under S2, medium flood--high landslide susceptibility becomes the largest class at 25.9\%, and high flood--high landslide susceptibility occupies 6.1\%, compared with 0.7\% under S1. High-flood combinations under S2 are concentrated mainly within the southern NNH4 belt, whereas high-landslide combinations paired with low or medium flood susceptibility extend across much of the NNH2 and NNH3 mountain and hill zones.

The Nepal risk maps are dominated by low flood--low landslide risk, which covers 58.0\% under S1 and 59.4\% under S2. Landslide-dominated combinations that occupy extensive areas in the susceptibility maps account for smaller areal shares in the risk maps. Risk combinations other than low flood--low landslide are concentrated mainly in the southern lowlands and selected central valleys. Low flood--high landslide risk covers a larger area under S1, whereas medium flood--high landslide risk is more extensive under S2. Low flood--low landslide risk is widespread across NNH2 and NNH3, while higher flood-related combinations occur mainly within NNH4 and selected central valleys. Across both regions, the two strategies retain the broad contrast between flood-prone lowlands and landslide-prone uplands but differ in the extent and spatial distribution of the combined susceptibility and risk classes.

\subsection{Spatial Analysis of Susceptibility--Risk Maps Using Autocorrelation, Agreement, and Transitions} \label{sec:spatial_consistency_and_transitions}

\begin{figure}[htp]
    \centering
    \includegraphics[width=\textwidth]{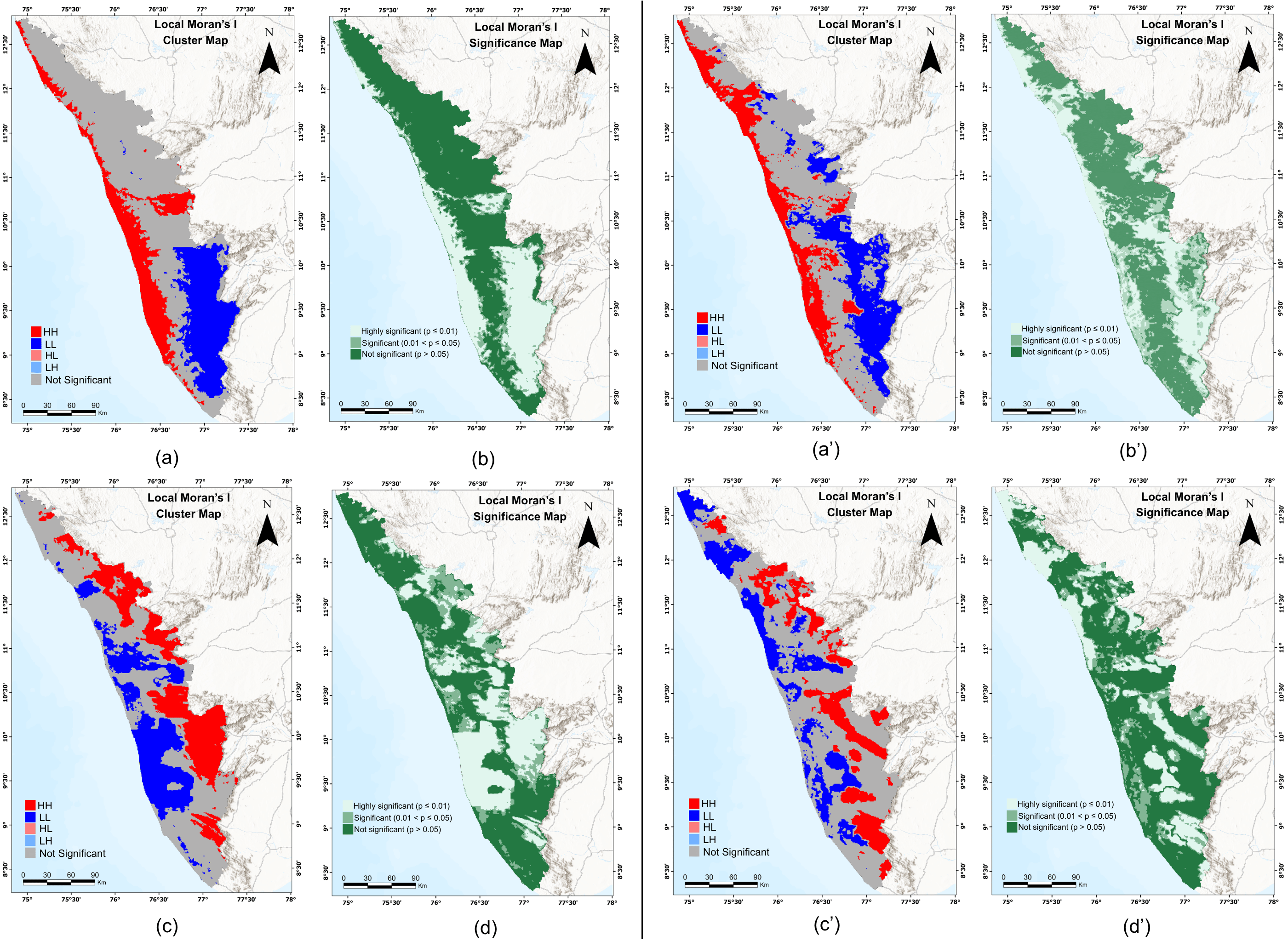}
    \caption{Local Moran's~$I$ cluster and significance maps for flood and landslide susceptibility in Kerala under S1 and S2. Panels (a) and (b) show the S1 flood cluster and significance maps, respectively, whereas panels (a$'$) and (b$'$) show the corresponding S2 flood maps. Panels (c) and (d) show the S1 landslide cluster and significance maps, respectively, whereas panels (c$'$) and (d$'$) show the corresponding S2 landslide maps.}
    \label{fig:kerala_lisa}
\end{figure}

\begin{figure}[hp]
    \centering
    \includegraphics[width=0.8\textwidth]{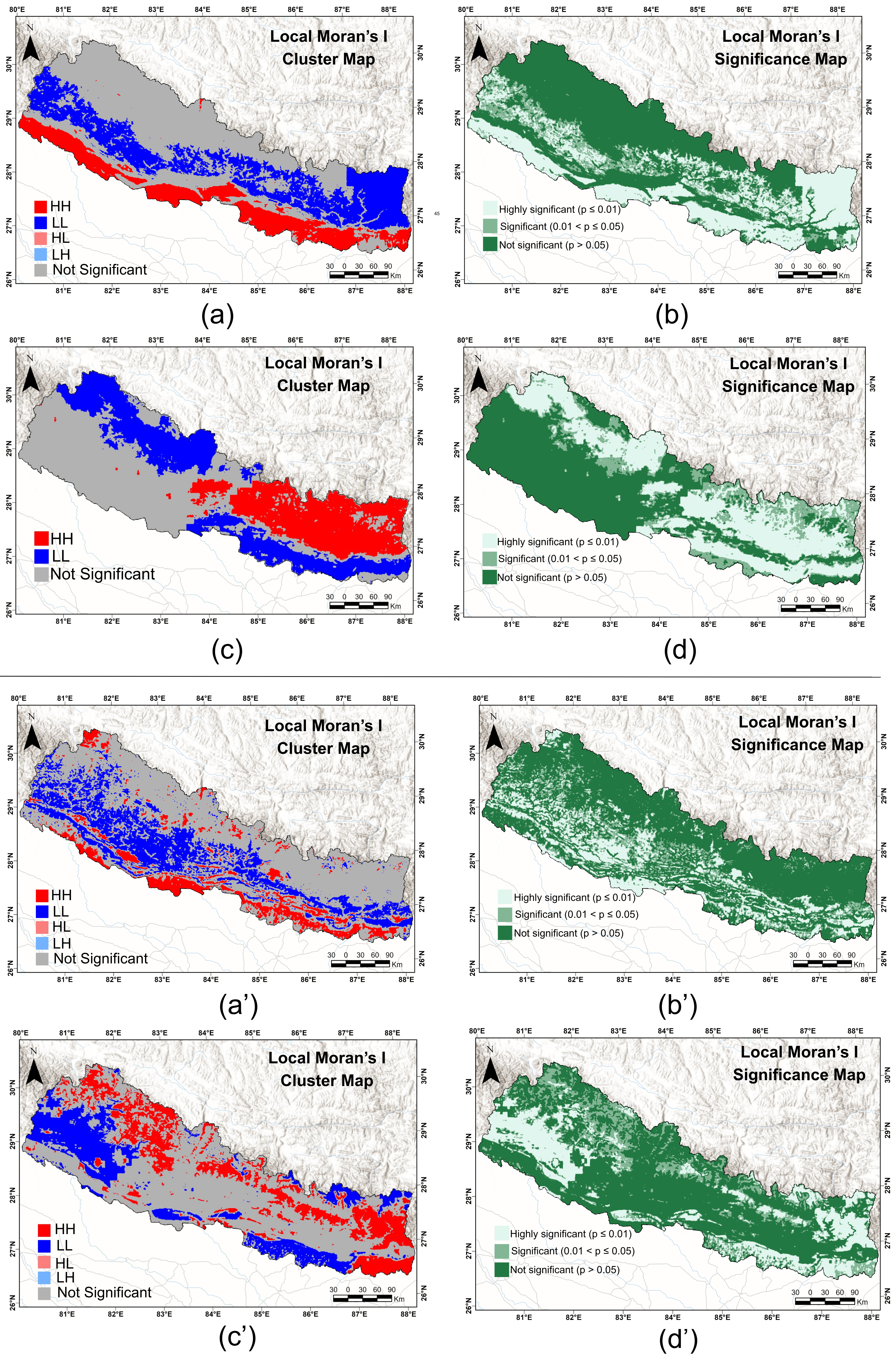}
    \caption{Local Moran's~$I$ cluster and significance maps for flood and landslide susceptibility in Nepal under S1 and S2. Panels (a) and (b) show the S1 flood cluster and significance maps, respectively, whereas panels (a$'$) and (b$'$) show the corresponding S2 flood maps. Panels (c) and (d) show the S1 landslide cluster and significance maps, respectively, whereas panels (c$'$) and (d$'$) show the corresponding S2 landslide maps.}
    \label{fig:nepal_lisa}
\end{figure}

Spatial statistical analyses involve Local Indicators of Spatial Autocorrelation (LISA) metrics, namely, local Moran's $I$, agreement between S1- and S2-risk maps using the Jaccard index, and directional susceptibility-to-risk transitions using correspondence analysis.

\subsubsection{Spatial Autocorrelation}
Local Moran's~$I$ identifies high--high and low--low susceptibility clusters in both study regions, while high--low and low--high spatial outliers occupy comparatively limited areas, as shown in Figures~\ref{fig:kerala_lisa} and~\ref{fig:nepal_lisa}.

In Kerala, the S1 flood map shows high--high clusters across much of the western and central lowland belt, while low--low clusters are concentrated mainly in the southeastern uplands. Under S2, the high--high and low--low flood clusters exhibit a more interspersed arrangement across the state. The landslide maps show a similar strategy-related contrast, with high--high and low--low clusters more closely interspersed under S2 than under S1. Significant local associations occur within selected belts and patches rather than uniformly across Kerala.

Nepal exhibits a clearer belt-like arrangement of local spatial association. Under S1, high--high flood clusters follow much of the southern lowland margin, whereas low--low clusters extend across the central and northern terrain. Under S2, high--high clusters occupy a larger part of the central and eastern lowland belt, while low--low clusters remain extensive farther north. The landslide maps contain elongated high--high and low--low cluster zones across the hill and mountain regions, with a more interspersed arrangement under S2. The significance maps indicate that statistically significant associations are concentrated within selected belts rather than distributed uniformly across Nepal.

\subsubsection{Agreement between S1 and S2 Risk Maps}
Agreement between S1 and S2 is consistently higher for the individual-hazard risk maps than for the nine-class bivariate maps, as shown in Table~\ref{tab:risk_agreement_summary}. In Kerala, flood risk shows the highest overall agreement at 82.53\%, compared with 60.80\% for landslide risk and 52.08\% for the bivariate risk map. In Nepal, overall agreement exceeds 81\% for both individual hazards and reaches 71.11\% for the bivariate map.

Allocation disagreement exceeds quantity disagreement in every S1--S2 comparison, indicating that differences between the strategies are associated mainly with the spatial locations of the classified risk areas rather than with their total class proportions. This pattern is strongest for the Kerala bivariate map, for which allocation disagreement reaches 37.38\%.

\begin{table}[tp]
\caption{S1--S2 agreement for classified risk maps in Kerala and Nepal.}
\label{tab:risk_agreement_summary}
\centering
\scriptsize
\setlength{\tabcolsep}{6pt}
\renewcommand{\arraystretch}{1.15}

\begin{tabular*}{\linewidth}{@{\extracolsep{\fill}}l ccc ccc@{}}
\toprule
& \multicolumn{3}{c}{\bf Kerala}
& \multicolumn{3}{c}{\bf Nepal}
\\
\cmidrule(lr){2-3}
\cmidrule(lr){3-4}
\cmidrule(lr){5-6}
\cmidrule(lr){6-7}

\bf Metric
& \bf Flood & \bf Landslide & \bf Multi-hazard
& \bf Flood & \bf Landslide & \bf Multi-hazard
\\ \midrule
$A_{\mathrm{o}}$ (\%) $\uparrow$
& \bf 82.53 & 60.80 & 52.08
& 81.63 & \bf 82.01 & 71.11 \\

$J_{\mathrm{macro}}$ $\uparrow$
& \bf 0.700 & 0.430 & 0.281
& 0.529 & \bf 0.598 & 0.260 \\

$J_{\mathrm{uw}}$ $\uparrow$
& \bf 0.703 & 0.437 & 0.352
& 0.690 & \bf 0.695 & 0.552 \\

$Q$ (\%) $\downarrow$
& \bf 3.93 & 9.66 & 10.54
& 7.11 & \bf 3.86 & 13.75 \\

$D_{\mathrm{a}}$ (\%) $\downarrow$
& \bf 13.54 & 29.54 & 37.38
& \bf 11.26 & 14.13 & 15.14 \\

\bottomrule
\end{tabular*}

\vspace{3pt}
{\footnotesize
\textit{Note:} $A_{\mathrm{o}}$ = overall agreement; $J_{\mathrm{macro}}$ = macro Jaccard; $J_{\mathrm{uw}}$ = union-weighted Jaccard;

$Q$ = quantity disagreement; $D_{\mathrm{a}}$ = allocation disagreement.

\textbf{Boldface} indicates the best value for each metric for each study area, across single and multi-hazard analyses. 
\par}
\end{table}

Class-wise agreement is generally greater for low-risk classes than for medium- and high-risk classes. In Nepal, the low-risk Jaccard values are 0.854 for flood risk and 0.880 for landslide risk. The low flood--low landslide class also shows the strongest bivariate agreement in Kerala and Nepal, with Jaccard values of 0.809 and 0.950, respectively. Several mixed bivariate classes show weak overlap, including high flood--low landslide risk, for which the Jaccard values are 0.026 in Kerala and 0.011 in Nepal.

The difference between the macro and union-weighted Jaccard indices is particularly evident for the Nepal bivariate maps. The higher union-weighted value indicates that agreement is concentrated in spatially extensive classes, especially low flood--low landslide risk, whereas the less extensive mixed classes exhibit substantially lower overlap.

Overall, the two strategies retain the principal regional hazard structures but differ in their local cluster configurations and in the spatial agreement of the classified risk maps. This is further reinforced in the Jaccard overlap-based agreement for individual hazards in Table~\ref{tab:classwise_jaccard_individual}, and bivariate risk class combinations in Table~\ref{tab:classwise_jaccard_bivariate}.

\begin{table}[tp]
\caption{Class-wise Jaccard values for S1--S2 agreement in risk maps of individual hazards.}
\label{tab:classwise_jaccard_individual}
\centering
\scriptsize
\setlength{\tabcolsep}{6pt}
\renewcommand{\arraystretch}{1.15}

\begin{tabular*}{\linewidth}{@{\extracolsep{\fill}}l ccc ccc@{}}
\toprule
& \multicolumn{3}{c}{\bf Kerala}
& \multicolumn{3}{c}{\bf Nepal}
\\
\cmidrule(lr){2-3}
\cmidrule(lr){3-4}
\cmidrule(lr){5-6}
\cmidrule(lr){6-7}

\bf Hazard $\downarrow$ \textbackslash Risk Class $\rightarrow$
& \bf Low & \bf Medium & \bf High
& \bf Low & \bf Medium & \bf High
\\ \midrule

\bf Flood
& \bf 0.814 & 0.630 & 0.658
& \bf 0.854 & 0.306 & 0.426 \\

\bf Landslide
& \bf 0.549 & 0.374 & 0.368
& \bf 0.880 & 0.453 & 0.460 \\

\bottomrule
\end{tabular*}

\vspace{3pt}
{\footnotesize
\textit{Note:} \textbf{Boldface} indicates the best value for each hazard for each study area across different risk classes. 
\par}

\vspace{2em}

\centering
\caption{Class-wise Jaccard values for S1--S2 agreement in bivariate flood--landslide risk maps.}
\label{tab:classwise_jaccard_bivariate}
\centering
\scriptsize
\setlength{\tabcolsep}{6pt}
\renewcommand{\arraystretch}{1.15}

\begin{tabular*}{\linewidth}{@{\extracolsep{\fill}}l ccc ccc@{}}
\toprule
& \multicolumn{3}{c}{\bf Kerala}
& \multicolumn{3}{c}{\bf Nepal}
\\
\cmidrule(lr){2-3}
\cmidrule(lr){3-4}
\cmidrule(lr){5-6}
\cmidrule(lr){6-7}

\bf Landslide $\downarrow$ \textbackslash Flood $\rightarrow$
&&& &&& \\
\bf Risk Classes
& \bf Low & \bf Medium & \bf High
& \bf Low & \bf Medium & \bf High
\\ \midrule

\bf Low
& \bf 0.809 & 0.180 & 0.026
& \bf 0.950 & 0.106 & 0.011 \\

\bf Medium
& \it 0.347 & 0.269 & 0.219
& \it 0.376 & 0.129 & 0.182 \\

\bf High
& 0.084 & 0.236 & 0.357
& 0.096 & 0.221 & 0.267 \\

\bottomrule
\end{tabular*}

\vspace{3pt}
{\footnotesize
\textit{Note:} \textbf{Boldface} and \textbf{italicized} indicate top-2 best values for each study area across all bivariate risk class combinations. 
\par}
\end{table}

\begin{table}[tbp]
\caption{Susceptibility-to-risk correspondence after exposure--vulnerability integration.}
\label{tab:susceptibility_risk_correspondence}
\centering
\scriptsize
\setlength{\tabcolsep}{4.5pt}
\renewcommand{\arraystretch}{1.12}

\begin{tabular*}{\linewidth}{@{\extracolsep{\fill}}l ccc ccc@{}}
\toprule
& \multicolumn{3}{c}{\bf Kerala}
& \multicolumn{3}{c}{\bf Nepal}
\\
\cmidrule(lr){2-3}
\cmidrule(lr){3-4}
\cmidrule(lr){5-6}
\cmidrule(lr){6-7}

\bf Metric
& \bf Flood & \bf Landslide & \bf Multi-hazard
& \bf Flood & \bf Landslide & \bf Multi-hazard
\\ \midrule

\multicolumn{7}{c}{\bf S1 Training Strategy}
\\ \midrule

$A_{\mathrm{o}}$ (\%) $\uparrow$
& \bf 57.02 & 52.96 & 34.19
& \bf 60.57 & 50.68 & 26.63 \\

$J_{\mathrm{macro}}$ $\uparrow$
& \bf 0.399 & 0.362 & 0.197
& \bf 0.409 & 0.327 & 0.179 \\

$J_{\mathrm{uw}}$ $\uparrow$
& \bf 0.399 & 0.360 & 0.206
& \bf 0.434 & 0.339 & 0.154 \\

$Q$ (\%) $\downarrow$
& \bf 17.63 & 18.06 & 37.37
& \bf 26.88 & 32.89 & 56.59 \\

$D_{\mathrm{a}}$ (\%) $\downarrow$
& \bf 25.35 & 28.98 & 28.45
& \bf 12.55 & 16.43 & 16.78 \\

\midrule

\multicolumn{7}{c}{\bf S2 Training Strategy}
\\ \midrule

$A_{\mathrm{o}}$ (\%) $\uparrow$
& \bf 66.47 & 47.29 & 32.37
& \bf 59.18 & 31.33 & 23.72 \\

$J_{\mathrm{macro}}$ $\uparrow$
& \bf 0.490 & 0.291 & 0.190
& \bf 0.405 & 0.193 & 0.180 \\

$J_{\mathrm{uw}}$ $\uparrow$
& \bf 0.498 & 0.310 & 0.193
& \bf 0.420 & 0.186 & 0.135 \\

$Q$ (\%) $\downarrow$
& \bf 6.42 & 18.69 & 32.62
& \bf 22.91 & 50.21 & 52.35 \\

$D_{\mathrm{a}}$ (\%) $\downarrow$
& \bf 27.11 & 34.03 & 35.01
& \bf 17.91 & 18.47 & 23.92 \\

\bottomrule
\end{tabular*}

\vspace{3pt}
{\footnotesize
\textit{Note:} $A_{\mathrm{o}}$ denotes overall susceptibility-to-risk correspondence; $J_{\mathrm{macro}}$ and $J_{\mathrm{uw}}$ denote the macro and union-weighted Jaccard indices, respectively.

$Q$ denotes quantity disagreement, and $D_{\mathrm{a}}$ denotes allocation disagreement.

\textbf{Boldface} indicates the best value for each metric for each study area for each training strategy, across single and multi-hazard analyses. 
\par}
\end{table}

\subsubsection{Susceptibility-to-risk Correspondence and Directional Transitions}
Susceptibility-to-risk correspondence is consistently higher for the flood maps than for the landslide and bivariate maps, as shown in Table~\ref{tab:susceptibility_risk_correspondence}. Flood correspondence ranges from 57.02\% to 66.47\% in Kerala and from 59.18\% to 60.57\% in Nepal. Landslide correspondence is lower, particularly under Nepal S2, where 31.33\% of pixels retain the same class. The bivariate maps show the lowest correspondence, ranging from 32.37\% to 34.19\% in Kerala and from 23.72\% to 26.63\% in Nepal.

The relative contributions of quantity and allocation disagreement vary between the regions and strategies. In Kerala, allocation disagreement exceeds quantity disagreement for the individual flood and landslide maps under both strategies. For the Kerala bivariate maps, quantity disagreement exceeds allocation disagreement under S1, whereas allocation disagreement exceeds quantity disagreement under S2. In Nepal, quantity disagreement exceeds allocation disagreement for every individual-hazard and bivariate comparison. Quantity disagreement is highest for the Nepal bivariate maps, reaching 56.59\% under S1 and 52.35\% under S2, and is also high for the S2 landslide map at 50.21\%.

Downward susceptibility-to-risk class transitions exceed upward transitions in every individual-hazard comparison, as shown in Table~\ref{tab:susceptibility_risk_directional_transition}. The difference is largest for Nepal landslide susceptibility under S2, where 60.98\% of pixels move to a lower classified risk level and 7.69\% move to a higher classified risk level. Nepal also exhibits more downward than upward flood-class transitions under both strategies. In Kerala, the difference between downward and upward transitions is larger for landslide than for flood susceptibility under both strategies, while the S2 flood map contains nearly equal proportions of the two transition directions.

\begin{table}[tbp]
\caption{Directional susceptibility-to-risk class transitions after exposure--vulnerability integration.}
\label{tab:susceptibility_risk_directional_transition}
\centering
\scriptsize
\setlength{\tabcolsep}{4.5pt}
\renewcommand{\arraystretch}{1.12}

\begin{tabular*}{\linewidth}{@{\extracolsep{\fill}}l cccc cccc@{}}
\toprule
& \multicolumn{4}{c}{\bf Kerala}
& \multicolumn{4}{c}{\bf Nepal}
\\
\cmidrule(lr){2-3}
\cmidrule(lr){3-4}
\cmidrule(lr){4-5}
\cmidrule(lr){6-7}
\cmidrule(lr){7-8}
\cmidrule(lr){8-9}

& \multicolumn{2}{c}{\bf Flood} & \multicolumn{2}{c}{\bf Landslide}
& \multicolumn{2}{c}{\bf Flood} & \multicolumn{2}{c}{\bf Landslide}
\\
\cmidrule(lr){2-3}
\cmidrule(lr){4-5}
\cmidrule(lr){6-7}
\cmidrule(lr){8-9}

\bf Metric
& \bf S1 & \bf S2 & \bf S1 & \bf S2
& \bf S1 & \bf S2 & \bf S1 & \bf S2 
\\ \midrule

$P_{\mathrm{c}}$ (\%) 
& 57.02 & 66.47 & 52.96 & 47.29
& 60.57 & 59.18 & 50.68 & 31.33 \\

$R_{\mathrm{up}}$ (\%) 
& 16.43 & 16.20 & 15.22 & 17.60
& 5.23 & 8.96 & 7.99 & 7.69 \\

$R_{\mathrm{down}}$ (\%) 
& 26.56 & 17.34 & 31.83 & 35.12
& 34.20 & 31.87 & 41.32 & 60.98 \\

\bottomrule
\end{tabular*}

\vspace{3pt}
{\footnotesize
\textit{Note:} $P_{\mathrm{c}}$ denotes class persistence, representing pixels that remain in the same class after exposure--vulnerability integration. $R_{\mathrm{up}}$ denotes an upward transition, representing pixels assigned to a higher class on the separately classified risk surface than on the susceptibility surface. $R_{\mathrm{down}}$ denotes a downward transition, representing pixels assigned to a lower class on the risk surface.

These directional measures describe transitions between separately classified susceptibility and risk maps and are reported only for the three-class flood and landslide maps, as the bivariate multi-hazard map does not form a single ordinal low--medium--high sequence.
\par}
\end{table}

Overall persistence of the same nine-class bivariate combination remains below 35\% in all susceptibility-to-risk comparisons, as shown in Table~\ref{tab:bivariate_mh_class_persistence}. Persistence nevertheless varies among individual classes. In Kerala, the highest class-specific persistence occurs for medium flood--medium landslide under S1 at 46.8\% and high flood--high landslide under S2 at 52.9\%. In Nepal, low flood--low landslide susceptibility retains the same bivariate class in the risk map for 92.7\% of pixels under S1 and 71.7\% under S2, whereas several mixed flood--landslide combinations show lower persistence.

\begin{table}[tbp]
\caption{Bivariate multi-hazard class persistence after exposure--vulnerability integration.}
\label{tab:bivariate_mh_class_persistence}
\centering
\scriptsize
\setlength{\tabcolsep}{4.5pt}
\renewcommand{\arraystretch}{1.12}

\begin{tabular*}{\linewidth}{@{\extracolsep{\fill}}lc cc cc@{}}
\toprule
&& \multicolumn{2}{c}{\bf Kerala}
& \multicolumn{2}{c}{\bf Nepal} \\
\cmidrule{3-4}
\cmidrule{5-6}

\textbf{Class} & \textbf{Flood / Landslide Class}
& \textbf{S1}
& \textbf{S2}
& \textbf{S1}
& \textbf{S2} \\
\midrule

Overall & Multi-hazard Persistence & 34.19 & 32.37 & 26.63 & 23.72 \\
\midrule
1 & Low / Low       & 43.2 & 34.3 & 92.7 & 71.7 \\
2 & Medium / Low    & 45.3 & 39.6 & 4.4  & 14.9 \\
3 & High / Low      & 35.3 & 28.2 & 33.2 & 25.1 \\
4 & Low / Medium    & 29.7 & 28.2 & 26.3 & 25.4 \\
5 & Medium / Medium & 46.8 & 48.2 & 18.5 & 30.9 \\
6 & High / Medium   & 37.4 & 35.7 & 40.3 & 44.9 \\
7 & Low / High      & 15.3 & 8.9  & 29.3 & 7.7  \\
8 & Medium / High   & 30.2 & 33.0 & 24.0 & 13.4 \\
9 & High / High     & 31.0 & 52.9 & 43.1 & 32.4 \\

\bottomrule
\end{tabular*}

\vspace{3pt}
{\footnotesize
\textit{Note:} Values are row-normalized diagonal percentages from the nine-class bivariate susceptibility-to-risk transition matrix. MH denotes multi-hazard. Class labels are expressed as flood class/landslide class.

Higher values indicate stronger persistence of the same bivariate class after exposure--vulnerability integration.
\par}
\end{table}

\section{Discussion} \label{sec:discussion}
This section interprets the effects of spatial training strategy on predictor retention, model performance, and mapped susceptibility patterns. It further examines regional and hazard-specific conditioning-factor relationships, the transition from susceptibility to relative risk following exposure--vulnerability integration, and the spatial consistency and planning implications of the resulting maps. The section concludes by outlining the principal limitations of the framework and directions for integrating broader regional learning with contextual-zone-specific modelling.

\subsection{Effect of Spatial Training Strategy on Model Performance} \label{sec:discussion_training_strategy}
The comparison between S1 and S2 shows that spatial training configuration affects both predictor retention and model performance. S1 achieves stronger mean classification and discrimination for both hazards and study regions, whereas S2 enables predictor screening and model development within individual contextual zones. Zone-constrained training therefore does not improve mean classification or discrimination in the present study, but it preserves spatial differences in predictor selection and hazard--factor relationships that are not represented by whole-region modelling. The two strategies consequently capture distinct aspects of regional susceptibility modelling: S1 supports information sharing across broader environmental settings, while S2 retains contextual-zone-specific variation.

In Kerala, the shift from whole-region screening of the entire state under S1 to ESZ-specific screening under S2 produces clear differences in the retention of hydrological, climatic, and anthropogenic conditioning factors. These differences are consistent with Kerala's pronounced west--east environmental gradient, within which coastal lowlands, midlands, river corridors, and the Western Ghats occur over relatively short distances. S1 draws training information from this broader range of environmental settings, whereas S2 estimates hazard--factor relationships separately within each ESZ category. The stronger discrimination achieved by S1 indicates that, for the available data and model configuration, the contextual specificity of S2 does not compensate for the narrower environmental and sample coverage within individual zones. Nevertheless, S2 reveals zone-specific differences in the retained conditioning factors that are obscured by whole-state screening.

Nepal retains a comparatively stable core of conditioning factors across the whole-region screening of the entire country and NNH-specific screening levels despite its pronounced lowland--hill--mountain gradient. S1 nevertheless produces stronger classification and discrimination for both hazards, indicating that the performance difference is not attributable to predictor retention alone. It may also reflect differences in the number, spatial distribution, and environmental coverage of occurrence and non-occurrence samples within the respective training domains. The contrast is strongest for flood susceptibility, for which S1 provides substantially clearer separation between occurrence and non-occurrence samples. Landslide performance is lower than flood performance under both strategies, which may reflect the greater complexity of landslide controls, limitations in the available conditioning factors, differences in inventory representation, or a combination of these influences rather than the spatial training strategy alone.

The lower Brier scores obtained by S2 in Nepal qualify the discrimination advantage of S1. Although S1 provides stronger classification and discrimination, S2 yields lower mean probabilistic prediction error for both hazards in Nepal. These values do not independently demonstrate better calibration because the Brier score reflects both calibration and discrimination and is influenced by event prevalence. Instead, the results show that the effect of spatial training strategy on probabilistic prediction error differs from its effect on classification and discrimination. Because the same pattern is not observed in Kerala, the Brier-score advantage of S2 appears specific to the Nepal models rather than a general benefit of zone-constrained training.

Taken together, the results show that S1 and S2 should not be interpreted solely as competing alternatives. S1 provides stronger regional discrimination and broader information sharing, whereas S2 preserves contextual differences in predictor selection and localized hazard--factor relationships. This complementarity motivates an integrated spatial heterogeneity-aware system that combines regional learning with contextual-zone-specific modelling rather than relying exclusively on either training structure.

\subsection{Regional and Hazard-specific Controls on Susceptibility} \label{sec:control_on_susceptibility}
The SHAP analysis shows that the importance and response patterns of the retained conditioning factors vary between hazards, study regions, and spatial configurations. Although several predictors recur across the models, their rankings and modelled contributions differ among locations and contextual zones. Because SHAP values describe associations within the fitted models, they should not be interpreted as independent causal effects, particularly where correlated predictors are retained.

In Kerala, elevation and LULC are the most consistent predictors of flood susceptibility. Under S1 and ESZ3, the lowest elevation values have positive SHAP contributions, whereas higher values generally have negative contributions. This pattern is consistent with earlier flood-susceptibility research in Kerala identifying elevation, drainage density, and distance to rivers as relevant conditioning factors~\citep{kendagannaswamy2025multi}. The susceptibility maps exhibit a comparable west--east contrast, with flood-dominated and mixed combinations covering more of the western and central parts of the state and landslide-dominated combinations occurring more widely in the eastern uplands. LULC remains among the leading flood predictors in all Kerala configurations, although the contributions of individual land-cover classes differ between the whole-state and contextual-zone models, indicating spatial variation in the modelled role of land cover.

The ESZ2 flood model differs from the other Kerala configurations, with population density having the highest mean absolute SHAP value. Its dependence pattern is nonlinear, with positive contributions concentrated within a limited range of lower observed values and near-zero or slightly negative contributions at higher values. This pattern should not be interpreted as evidence that increasing population density directly increases flood susceptibility. Instead, population density may act as a proxy for settlement-related conditions in low-relief areas affected by built-up expansion and modified drainage. Studies in central and southern Kerala have linked urban expansion, impermeable surfaces, wetland loss, and drainage alteration with increased flood depth or susceptibility~\citep{thaivalappil2024investigating,sarun2026spatiotemporal}. The ESZ2 response may therefore reflect the spatial coincidence of population concentration with these settlement-related conditions.

Kerala landslide susceptibility is more strongly associated with terrain ruggedness, distance to faults, drainage conditions, and LULC. TRI is the leading predictor under S1, ESZ3, and Non-ESZ. Its dependence plots show a transition from negative to positive contributions at low-to-intermediate values, followed by stable or gradually declining contributions at higher values. This pattern is consistent with studies in the Western Ghats linking landslide occurrence with rugged hills, steep slopes, relative relief, structural landforms, and hydrological conditions~\citep{achu2020spatial,sajinkumar2015geomorphic}. Distance to faults ranks first in ESZ1, whereas drainage density ranks first in ESZ2, demonstrating that the dominant landslide-related factors differ among contextual zones. The susceptibility maps reflect the same broad environmental contrast, with landslide-dominated combinations concentrated mainly in the eastern uplands and flood-dominated or mixed combinations occurring more widely across the western Non-ESZ belt.

In Nepal, flood susceptibility is associated mainly with elevation, slope, and LULC. Elevation has the largest contribution under S1 and NNH4, slope is dominant in NNH2, and LULC ranks first in Non-NNH and NNH3. Under S1 and NNH4, the lowest elevation values have positive contributions, while higher values generally have negative contributions. The NNH2 slope response follows a similar pattern, with low slopes contributing positively and steeper terrain contributing negatively. These responses are consistent with Nepal's south--north physiographic gradient. Flood-hazard modelling in the downstream Karnali basin similarly identifies the southern plains as highly exposed to monsoon flooding~\citep{aryal2020model}. The concentration of flood-dominated combinations in the southern NNH4 belt is consistent with this regional pattern. Differences in the LULC responses between Non-NNH and NNH3 further indicate that the modelled contributions of individual land-cover classes vary among contextual settings.

For Nepal landslide susceptibility, precipitation and distance to faults consistently rank among the top two predictors in the whole-country and contextual-zone models. Under S1, distance to faults exhibits a non-monotonic response: the shortest distances are associated mainly with negative contributions, intermediate distances with positive contributions, and greater distances with weaker or mixed contributions. The modelled response therefore does not indicate a simple increase in susceptibility toward mapped faults. Precipitation also exhibits nonlinear responses under S2, generally shifting from negative to positive contributions across intermediate values, although the position and form of this transition differ among Non-NNH, NNH2, NNH3, and NNH4. Previous studies in Nepal and the central Himalaya identify rainfall, topography, geological structure, and drainage conditions as important controls on landslide susceptibility~\citep{zhang2019size,gnyawali2023framework}. The extensive landslide-dominated combinations across NNH2 and NNH3 and the stronger flood dominance of the southern NNH4 belt are consistent with this physiographic differentiation.

Overall, the influential predictors and their response patterns differ between hazards, study regions, and spatial configurations. Flood susceptibility is associated primarily with elevation, slope, LULC, and lowland or riverine environmental conditions, whereas landslide susceptibility is more strongly associated with terrain ruggedness, precipitation, distance to faults, and drainage conditions. The zone-specific SHAP results demonstrate that predictor rankings and response patterns vary among contextual settings. S1 captures broader regional associations, while S2 reveals localized differences that are obscured when the study region is treated as a single modelling domain.

\subsection{From Susceptibility to Risk: Influence of Exposure and Vulnerability} \label{sec:discussion_susceptibility_to_risk}
The comparison between susceptibility and risk maps shows that they represent different dimensions of regional hazard conditions. Susceptibility identifies locations where the selected environmental conditions favor flood or landslide occurrence, whereas mapped relative risk also incorporates the spatial distribution of exposure and vulnerability. The comparatively low susceptibility-to-risk correspondence, particularly for the nine-class bivariate maps, demonstrates that the risk maps are not simple reproductions of the underlying susceptibility patterns.

In Kerala, exposure--vulnerability integration produces widespread downward class transitions alongside localized concentrations of high relative risk. Low flood--low landslide combinations occupy larger shares of the risk maps than of the corresponding susceptibility maps under both strategies, consistent with many susceptible locations having comparatively low combined exposure--vulnerability values and with the separate classification of the susceptibility and risk surfaces. At the same time, high flood--high landslide combinations also occupy larger shares of the risk maps, indicating a smaller set of locations where both hazard-specific risk surfaces are assigned to the high class after exposure and vulnerability are incorporated. This spatial concentration is consistent with Kerala's compact coast--midland--upland gradient, where densely settled lowlands, river corridors, and developed midlands occur close to landslide-prone Western Ghats terrain and upland margins.

The larger areal share of high flood--high landslide risk in Kerala does not conflict with the predominance of downward transitions in the individual-hazard maps. Flood and landslide transitions are evaluated separately, whereas each bivariate class is determined by the combination of the two independently classified hazard-specific risk surfaces. A limited set of locations can therefore be classified as high for both hazards even when downward transitions dominate the individual-hazard comparisons. Because the susceptibility and risk surfaces are classified separately, these transitions describe changes in class membership rather than direct numerical increases in continuous risk relative to susceptibility. The Kerala results therefore show that exposure and vulnerability alter both the spatial allocation of risk classes and the concentration of high multi-hazard risk.

Nepal exhibits more extensive downward susceptibility-to-risk transitions. Large areas of landslide-dominated susceptibility across the hills and mountains shift to lower risk combinations after exposure--vulnerability integration, whereas higher-risk combinations are concentrated mainly in the southern lowlands, major valleys, and settlement-linked corridors. The reduced areal share of low flood--high landslide combinations in the risk maps reflects the limited spatial overlap between highly susceptible mountain terrain and the main concentrations of population, built-up surfaces, infrastructure, and the vulnerability conditions included in the analysis. This contrast is particularly clear across the NNH zones: much of the NNH2 and NNH3 hill and mountain terrain exhibits landslide-dominated susceptibility but low corresponding risk, while higher flood-related risk remains concentrated mainly within the southern NNH4 belt and selected central valleys.

The correspondence and directional-transition measures reinforce these regional differences. Flood susceptibility shows greater correspondence with risk than the landslide and bivariate classifications, which may reflect the stronger spatial coincidence of flood-prone terrain and human activity in lowland and riverine environments. Downward transitions exceed upward transitions in every individual-hazard comparison, with the largest difference occurring for Nepal landslide susceptibility under S2. This pattern is consistent with the limited overlap between extensive landslide susceptibility in the hill and mountain zones and the exposure--vulnerability conditions represented in the risk assessment.

Quantity and allocation disagreement further distinguish the regional responses to exposure--vulnerability integration. In Kerala, allocation disagreement exceeds quantity disagreement for the individual-hazard maps, indicating that integration primarily changes the spatial locations assigned to each risk class while producing smaller changes in total class proportions. In Nepal, quantity disagreement is consistently larger, particularly for the landslide and bivariate maps, indicating stronger changes in overall class composition alongside widespread downward transitions.

The low persistence of the nine-class bivariate maps also reflects the structure of the class combination. A pixel retains its original bivariate class only when both its flood and landslide classes remain unchanged after exposure--vulnerability integration; a change in either component transfers it to another multi-hazard combination. The high persistence of low flood--low landslide conditions in Nepal indicates that many locations assigned to this susceptibility combination remain in the same bivariate risk class. By contrast, the lower persistence of several mixed combinations shows that their final classification is more sensitive to changes in either hazard component.

Susceptibility and mapped relative risk should therefore not be used interchangeably for spatial prioritization. Susceptibility maps identify the relative spatial propensity for hazard occurrence, including within sparsely occupied mountain and upland environments, whereas relative-risk maps show how these patterns are modified by the exposure and vulnerability conditions represented in the analysis. Low mapped relative risk should not be interpreted as low susceptibility, particularly in the landslide-prone hills and mountains of Nepal. The two products consequently support different planning functions: susceptibility maps inform hazard-sensitive land management and development control, while relative-risk maps support the prioritization of preparedness, exposure reduction, and other risk-management measures.

\subsection{Spatial Consistency and Planning Implications} \label{sec:discussion_implications_limitations}
Local Moran's~$I$ confirms that the susceptibility surfaces are spatially structured, with coherent high--high and low--low clusters in both study regions rather than isolated pixel-level predictions. The contrasting cluster arrangements in Kerala and Nepal reflect the different geographical organization of the two regions.

Kerala exhibits a comparatively interspersed arrangement of high--high and low--low clusters, consistent with its compressed coast--midland--upland gradient, where flood-prone lowlands, river corridors, modified midlands, and landslide-prone Western Ghats environments occur over short distances. Nepal exhibits a clearer belt-like pattern corresponding to the broader separation among the southern lowlands, central hills, and northern mountain terrain represented by the NNH framework. Flood-related high--high clusters are concentrated mainly toward the southern lowlands, whereas landslide-related clusters extend across the hill and mountain regions.

The spatial training strategy also affects the organization of the susceptibility clusters. S1 generally produces broader cluster belts, whereas S2 produces a more interspersed arrangement of high--high and low--low areas. The broader structures under S1 are consistent with information sharing across wider environmental domains, while the more localized patterns under S2 reflect the estimation of susceptibility relationships within individual contextual zones. Together with the performance results, these patterns show that training-region configuration influences both predictive discrimination and the spatial structure of the mapped susceptibility surfaces.

The significance maps identify belts and patches in which neighboring susceptibility values form statistically significant local associations. Significant high--high clusters indicate spatially coherent areas of elevated susceptibility, while high--low and low--high outliers identify localized departures from the surrounding pattern. Such outliers may be particularly informative near transitions between lowlands and uplands, river corridors and adjoining terrain, or neighboring contextual zones. Because the local tests are evaluated separately at $p<0.05$ without adjustment for multiple comparisons, the significance maps are interpreted as exploratory evidence of local spatial structure rather than definitive hazard boundaries.

Agreement between S1 and S2 is higher for the individual-hazard risk maps than for the nine-class bivariate maps because a change in either the flood or landslide component transfers a location to a different combined class. Agreement is strongest for the low-risk classes, particularly low flood--low landslide risk, whereas several combinations of medium-, high-, and mixed-risk classes are more sensitive to the training strategy. In Nepal, the difference between the macro and union-weighted Jaccard indices further shows that overall agreement is influenced strongly by the spatially extensive low flood--low landslide class.

Allocation disagreement exceeds quantity disagreement in every S1--S2 risk-map comparison, indicating that the strategies differ more in the locations assigned to the risk classes than in the total regional proportions of those classes. This distinction is important for spatial planning because similar class percentages can conceal substantial differences in the locations identified as high or mixed risk. Areas classified consistently under both strategies are less sensitive to training-domain configuration, whereas areas of disagreement require more cautious interpretation. In Kerala, this comparison can support coordinated land-use regulation, drainage management, slope monitoring, and preparedness across the densely occupied coast--midland--upland transition. In Nepal, it can help distinguish persistent lowland and valley risk concentrations from strategy-sensitive areas across the hills and mountains, particularly where mixed-hazard classes affect settlements, transport corridors, and future development.

\subsection{Limitations and Future Directions}
The spatial configurations are fixed to provide a consistent basis for comparing the two study regions and training strategies. The analysis uses 15~km $\times$ 15~km computational grids, eight-cell training regions under S1, two-unit training regions under S2, and two selected training regions within each comparison group. The relative behavior of S1 and S2 may therefore vary with alternative grid resolutions, neighborhood extents, or numbers of selected training regions.

The framework also uses one contextual-zonation system for each study region: the ESZ classification for Kerala and the NNH classification for Nepal. These zonations provide suitable broad environmental distinctions for evaluating zone-constrained learning, although alternative physiographic, ecological, hydro-geomorphological, or data-derived zonations may represent spatial heterogeneity differently.

S1 and S2 are evaluated as separate training strategies. The results show that S1 supports broader regional information sharing and stronger mean discrimination, whereas S2 preserves contextual differences in predictor selection, factor contributions, and mapped patterns. Future work can examine alternative spatial scales and zonation schemes, extend the framework to additional regions and hazard combinations, and develop an adaptive assignment or weighting mechanism that integrates regional and contextual-zone models. Such an extension would retain the broader regional learning of S1 while incorporating the localized environmental specificity represented by S2.

\section{Conclusion} \label{sec:conclusion}
This study examined how training-region construction and model assignment influence regional flood--landslide susceptibility and relative-risk mapping in Kerala and Nepal. The proposed generalized framework combined regular grid cells with region-specific contextual zones to accommodate regional-scale computations and spatial heterogeneity-aware ML models. This study used the framework to compare proximity-gated cross-zone training under S1 with zone-constrained training under S2.

The results show that S1 provides stronger mean classification and discrimination for both hazards and study regions, whereas S2 yields lower mean Brier scores for the two Nepal hazards. The zone-constrained models also preserve differences in predictor selection and modelled factor-response patterns among contextual zones, particularly in Kerala. These findings demonstrate that broader regional learning and contextual restriction influence different aspects of model behavior.

Both strategies retain the broad contrast between flood-prone lowlands and landslide-prone uplands but differ in the spatial distribution of susceptibility and risk classes. Differences between the classified S1 and S2 risk maps arise mainly from the locations assigned to the classes rather than from their total regional proportions. The low correspondence between the bivariate susceptibility and risk maps further demonstrates that incorporating exposure and vulnerability changes the spatial priorities identified by susceptibility alone. Susceptibility maps are therefore suited to identifying hazard-prone terrain, whereas relative-risk maps support the prioritization of exposed and vulnerable locations.

However, the takeaway from the study is not to regard S1 and S2 as competing alternatives. S1 supports information sharing across broader regional domains and provides stronger overall discrimination, whereas S2 preserves variation among environmental contexts. A spatial heterogeneity-aware multi-hazard system should therefore integrate cross-zone regional learning with contextual-zone-specific modelling. Although these components are evaluated separately in the present study, the findings provide a clear basis for developing and evaluating an integrated system.

\section*{Acknowledgments}
The authors would like to acknowledge the support from IIIT-B, GVCL, and the IEEE Geoscience and Remote Sensing Society (GRSS) for this work.

\bibliographystyle{unsrt}
\bibliography{papers_mhsm}

@misc{gsi2026bhukosh,
  author = {{Geological Survey of India}},
  title = {{Bhukosh: Geospatial Information System Portal}},
  year = {2026},
  institution = {Government of India},
  address = {Kolkata, India},
  url = {https://bhukosh.gsi.gov.in/Bhukosh/Public},
  note = {Accessed: 2026-08-05}
}

@article{diakoulaki1995determining,
  title={{Determining objective weights in multiple criteria problems: The CRITIC method}},
  author={Diakoulaki, Danae and Mavrotas, George and Papayannakis, Lefteris},
  journal={Computers \& Operations Research},
  volume={22},
  number={7},
  pages={763--770},
  year={1995},
  publisher={Elsevier},
  doi={10.1016/0377-2217(94)00262-X}
}

@article{weiss2020global,
  title={{Global maps of travel time to healthcare facilities}},
  author={Weiss, Daniel J and Nelson, Andrew and Gibson, Jessica S and Temperley, Harry and Peedell, Simon and Lieber, Allie and Etward, Marisa and others},
  journal={Nature Medicine},
  volume={26},
  number={12},
  pages={1835--1838},
  year={2020},
  publisher={Nature Publishing Group},
  doi={10.1038/s41591-020-1059-1}
}

@misc{worldpop_agesex2020,
  title = {{Global Estimated Age and Sex Structures of Residential Population 2000-2020 (Unconstrained)}},
  author = {{WorldPop}},
  publisher = {School of Geography and Environmental Science, University of Southampton},
  year = {2020},
  url = {https://www.worldpop.org/geodata/summary?id=24798},
  note = {Accessed: 2026-07-30}
}

@dataset{ciesin2022grdi,
  author = {{Center for International Earth Science Information Network - CIESIN - Columbia University}},
  title = {{Global Gridded Relative Deprivation Index (GRDI), Version 1}},
  year = {2022},
  publisher = {NASA Socioeconomic Data and Applications Center (SEDAC)},
  address = {Palisades, NY},
  doi = {10.7927/h4-g30p-q604},
  url = {https://sedac.ciesin.columbia.edu/data/set/povmap-grdi-v1}
}

@misc{geofabrik_osm,
  author = {{Geofabrik GmbH}},
  title = {{OpenStreetMap Data Extracts}},
  howpublished = {\url{https://download.geofabrik.de}},
  year = {2026},
  note = {Accessed: 2026-07-30}
}

@dataset{pesaresi2023ghsbuilt,
  author = {Pesaresi, Martino and Politis, Panagiotis},
  title = {{GHS-BUILT-S R2023A - GHS built-up surface grid, derived from Sentinel2 composite and Landsat, multitemporal (1975-2030)}},
  publisher = {European Commission, Joint Research Centre (JRC)},
  year = {2023},
  url = {https://human-settlement.emergency.copernicus.eu/ghs_buS2023.php}
}

@dataset{schiavina2026ghspop,
  author = {Schiavina, Marcello and Freire, Sergio and Carioli, Alessandra and MacManus, Kytt},
  title = {{GHS-POP R2023A - GHS population grid multitemporal (1975-2030)}},
  publisher = {European Commission, Joint Research Centre (JRC)},
  year = {2026},
  doi = {10.2905/2FF68A52-5B5B-4A22-8F40-C41DA8332CFE},
  url = {http://data.europa.eu/89h/2ff68a52-5b5b-4a22-8f40-c41da8332cfe}
}

@article{dinerstein2017ecoregion,
  title={{An ecoregion-based approach to protecting half the terrestrial realm}},
  author={Dinerstein, Eric and Olson, David and Joshi, Anup and Vynne, Carly and Burgess, Neil D and Wikramanayake, Eric and Hahn, Nathan and Palminteri, Suzanne and Hedao, Prashant and Noss, Reed and others},
  journal={BioScience},
  volume={67},
  number={6},
  pages={534--545},
  year={2017},
  publisher={Oxford University Press},
  doi={10.1093/biosci/bix014}
}

@techreport{wgeep2011report,
  author = {{Western Ghats Ecology Expert Panel} and Gadgil, Madhav},
  title = {{Report of the Western Ghats Ecology Expert Panel}},
  institution = {Ministry of Environment and Forests, Government of India},
  year = {2011},
  address = {New Delhi},
  url = {https://ruralindiaonline.org/en/library/resource/report-of-the-western-ghats-ecology-expert-panel/}
}

@article{gadgil2014western,
  title={{Western Ghats Ecology Expert Panel: A play in five acts}},
  author={Gadgil, Madhav},
  journal={Economic and Political Weekly},
  pages={38--50},
  year={2014},
  publisher={JSTOR}
}

@article{juang2019using,
  title={{Using citizen science to expand the global map of landslides: Introducing the Cooperative Open Online Landslide Repository (COOLR)}},
  author={Juang, Caroline S and Stanley, Thomas A and Kirschbaum, Dalia B},
  journal={PloS One},
  volume={14},
  number={7},
  pages={e0218657},
  year={2019},
  publisher={Public Library of Science San Francisco, CA USA},
  doi={10.1371/journal.pone.0218657},
}

@misc{unspider2024gee,
  author = {{UN-SPIDER}},
  title = {{Recommended Practice: Flood Mapping and Damage Assessment Using Sentinel-1 SAR Data in Google Earth Engine}},
  howpublished = {United Nations Platform for Space-Based Information for Disaster Management and Emergency Response (UN-SPIDER)},
  url = {https://www.un-spider.org/advisory-support/recommended-practices/recommended-practice-google-earth-engine-flood-mapping},
  note = {Accessed: 2026-07-30}
}

@article{gorelick2017google,
    title={{Google Earth Engine: Planetary-scale geospatial analysis for everyone}},
    author={Gorelick, Noel and Hancher, Matt and Dixon, Mike and Ilyushchenko, Simon and Thau, David and Moore, Rebecca},
    journal={Remote Sensing of Environment},
    year={2017},
    publisher={Elsevier},
    doi={10.1016/j.rse.2017.06.031},
    url={https://doi.org/10.1016/j.rse.2017.06.031}
  }

@article{chen2021one,
  title={{A one-class-classifier-based negative data generation method for rapid earthquake-induced landslide susceptibility mapping}},
  author={Chen, Shuai and Miao, Zelang and Wu, Lixin and Zhang, Anshu and Li, Qirong and He, Yueguang},
  journal={Frontiers in Earth Science},
  volume={9},
  pages={609896},
  year={2021},
  publisher={Frontiers Media SA},
  doi={10.3389/feart.2021.609896},
}

@article{ye2023generating,
  title={{Generating accurate negative samples for landslide susceptibility mapping: A combined self-organizing-map and one-class SVM method}},
  author={Ye, Chengming and Tang, Rong and Wei, Ruilong and Guo, Zixuan and Zhang, Huajun},
  journal={Frontiers in Earth Science},
  volume={10},
  pages={1054027},
  year={2023},
  publisher={Frontiers Media SA},
  doi={10.3389/feart.2022.1054027},
}

@article{li2025evaluation,
  title={{Evaluation of geological hazard susceptibility based on the multi-kernel density information method}},
  author={Li, Yang and Lei, Yutian and Chen, Bo and Chen, Jiale},
  journal={Scientific Reports},
  volume={15},
  number={1},
  pages={7892},
  year={2025},
  publisher={Nature Publishing Group UK London},
  doi={10.1038/s41598-025-91713-6},
}

@article{fan2023comparison,
  title={{Comparison of earthquake-induced shallow landslide susceptibility assessment based on two-category LR and KDE-MLR}},
  author={Fan, Xinyue and Liu, Bin and Luo, Jie and Pan, Ke and Han, Suyue and Zhou, Zhongli},
  journal={Scientific Reports},
  volume={13},
  number={1},
  pages={833},
  year={2023},
  publisher={Nature Publishing Group UK London},
  doi={10.1038/s41598-023-28096-z},
}

@misc{cred2021human,
author={{CRED}},
note={Centre for Research on the Epidemiology of Disasters (CRED) United Nations Office for Disaster Risk Reduction (UNDRR), last accessed on 2026-07-29},
title={{The human cost of disasters: An overview of the last 20 years (2000-2019)}},
howpublished={\url{https://www.undrr.org/publication/human-cost-disasters-overview-last-20-years-2000-2019}},
year={2021},
}

@article{wang2025region,
  title={{Region similarity assessment for empowering physics-informed transfer learning-based landslide susceptibility mapping}},
  author={Wang, Yunhao and Wang, Luqi and Liu, Songlin and Han, Liang and Zhang, Wengang and Hong, Li and Zhu, Zhengwei and Zhu, Xing},
  journal={Journal of Rock Mechanics and Geotechnical Engineering},
  year={2025},
  publisher={Elsevier},
  doi={10.1016/j.jrmge.2025.06.030}
}

@inproceedings{mundayatt2024scaling,
title={{Scaling up Study Area Size in Flood Susceptibility Mapping}},
author={Mundayatt, Aswathi and Sreevalsan-Nair, Jaya},
booktitle={Proceedings of 2024 IEEE International Geoscience and Remote Sensing Symposium (IGARSS)},
year={2024},
pages={3211--3214},
doi={10.1109/IGARSS53475.2024.10640798},
month={July}
}

@article{uddin2015development,
  title={Development of 2010 national land cover database for the Nepal},
  author={Uddin, Kabir and Shrestha, Him Lal and Murthy, MSR and Bajracharya, Birendra and Shrestha, Basanta and Gilani, Hammad and Pradhan, Sudip and Dangol, Bikash},
  journal={Journal of Environmental Management},
  volume={148},
  pages={82--90},
  year={2015},
  publisher={Elsevier}
}

@article{karki2017rising,
  title={Rising precipitation extremes across Nepal},
  author={Karki, Ramchandra and Hasson, Shabeh ul and Schickhoff, Udo and Scholten, Thomas and B{\"o}hner, J{\"u}rgen},
  journal={Climate},
  volume={5},
  number={1},
  pages={4},
  year={2017},
  publisher={MDPI}
}

@article{gaire2015disaster,
  title={Disaster risk profile and existing legal framework of Nepal: floods and landslides},
  author={Gaire, Surya and Castro Delgado, Rafael and Arcos Gonz{\'a}lez, Pedro},
  journal={Risk management and healthcare policy},
  pages={139--149},
  year={2015},
  publisher={Taylor \& Francis}
}

@article{velmurugan2026climate,
  title={Climate Oscillations, Aerosol Variability, and Land Use Change: Assessment of Drivers of Flood Risk in Monsoon-Dependent Kerala},
  author={Velmurugan, Sowmiya and Jayanarayanan, Brema and Sathian, Srinithisathian and Kantamaneni, Komali},
  journal={Earth},
  volume={7},
  number={1},
  pages={15},
  year={2026},
  publisher={MDPI}
}

@article{hao2020constructing,
  title={Constructing a complete landslide inventory dataset for the 2018 monsoon disaster in Kerala, India, for land use change analysis},
  author={Hao, Lina and Rajaneesh, A and Van Westen, Cees and Sajinkumar, KS and Martha, Tapas Ranjan and Jaiswal, Pankaj and McAdoo, Brian G},
  journal={Earth system science data},
  volume={12},
  number={4},
  pages={2899--2918},
  year={2020},
  publisher={Copernicus GmbH}
}

@article{kuriakose2009history,
  title={History of landslide susceptibility and a chorology of landslide-prone areas in the Western Ghats of Kerala, India},
  author={Kuriakose, Sekhar L and Sankar, G and Muraleedharan, C},
  journal={Environmental geology},
  volume={57},
  number={7},
  pages={1553--1568},
  year={2009},
  publisher={Springer}
}

@article{ward2022invited,
  title={Invited perspectives: A research agenda towards disaster risk management pathways in multi-(hazard-) risk assessment},
  author={Ward, Philip J and Daniell, James and Duncan, Melanie and Dunne, Anna and Hananel, C{\'e}dric and Hochrainer-Stigler, Stefan and Tijssen, Annegien and Torresan, Silvia and Ciurean, Roxana and Gill, Joel C and others},
  journal={Natural Hazards and Earth System Sciences},
  volume={22},
  number={4},
  pages={1487--1497},
  year={2022},
  doi= "10.5194/nhess-22-1487-2022"
}

@article{shrestha2026compound,
  title={Compound multi-hazard assessment under CMIP6 climate scenarios: tracking seasonal flood-landslide and drought-fire interactions across Nepal},
  author={Shrestha, Kripa and Duddukunta, Karthik Reddy and Lal, Pankaj and Hasan, Md Mehedi and Rahman, Md Mahfuzur and Pradhan, Neera Shrestha},
  journal={Natural Hazards},
  volume={122},
  number={14},
  pages={542},
  year={2026},
  doi= "10.1007/s11069-026-08310-7"
}

@article{pourghasemi2020assessing,
  title={Assessing and mapping multi-hazard risk susceptibility using a machine learning technique},
  author={Pourghasemi, Hamid Reza and Kariminejad, Narges and Amiri, Mahdis and Edalat, Mohsen and Zarafshar, Mehrdad and Blaschke, Thomas and Cerda, Artemio},
  journal={Scientific reports},
  volume={10},
  number={1},
  pages={3203},
  year={2020},
  doi= "10.1038/s41598-020-60191-3"
}

@article{stalhandske2024global,
  title={Global multi-hazard risk assessment in a changing climate},
  author={Stalhandske, Z{\'e}lie and Steinmann, Carmen B and Meiler, Simona and Sauer, Inga J and Vogt, Thomas and Bresch, David N and Kropf, Chahan M},
  journal={Scientific Reports},
  volume={14},
  number={1},
  pages={5875},
  year={2024},
  doi= "10.1038/s41598-024-55775-2"
}

@article{rusk2022multi,
  title={Multi-hazard susceptibility and exposure assessment of the Hindu Kush Himalaya},
  author={Rusk, Jack and Maharjan, Amina and Tiwari, Prakash and Chen, Tzu-Hsin Karen and Shneiderman, Sara and Turin, Mark and Seto, Karen C},
  journal={Science of the total environment},
  volume={804},
  pages={150039},
  year={2022},
  doi ="10.1016/j.scitotenv.2021.150039"
}

@article{fell2008guidelines,
  title={Guidelines for landslide susceptibility, hazard and risk zoning for land use planning},
  author={Fell, Robin and Corominas, Jordi and Bonnard, Christophe and Cascini, Leonardo and Leroi, Eric and Savage, William Z and JTC-1 Joint Technical Committee on Landslides and Engineered Slopes and others},
  journal={Engineering geology},
  volume={102},
  number={3-4},
  pages={85--98},
  year={2008},
  doi="10.1016/j.enggeo.2008.03.022"
}

@article{rahman2022assessment,
  title={Assessment of landslide susceptibility, exposure, vulnerability, and risk in shahpur valley, eastern hindu kush},
  author={Rahman, Ghani and Bacha, Alam Sher and Ul Moazzam, Muhammad Farhan and Rahman, Atta Ur and Mahmood, Shakeel and Almohamad, Hussein and Al Dughairi, Ahmed Abdullah and Al-Mutiry, Motrih and Alrasheedi, Mona and Abdo, Hazem Ghassan},
  journal={Frontiers in Earth Science},
  volume={10},
  pages={953627},
  year={2022},
  doi="10.3389/feart.2022.953627"
}

@article{ullah2022multi,
  title={Multi-hazard susceptibility mapping based on Convolutional Neural Networks},
  author={Ullah, Kashif and Wang, Yi and Fang, Zhice and Wang, Lizhe and Rahman, Mahfuzur},
  journal={Geoscience Frontiers},
  volume={13},
  number={5},
  pages={101425},
  year={2022},
  doi="10.1016/j.gsf.2022.101425"
}

@article{kumar2023assessment,
  title={Assessment and mapping of riverine flood susceptibility (RFS) in India through coupled multicriteria decision making models and geospatial techniques},
  author={Kumar, Ravi and Kumar, Manish and Tiwari, Akash and Majid, Syed Irtiza and Bhadwal, Sourav and Sahu, Netrananda and Avtar, Ram},
  journal={Water},
  volume={15},
  number={22},
  pages={3918},
  year={2023},
  doi="10.3390/w15223918"
}

@article{kumar2025random,
  title={{Random cross-validation produces biased assessment of machine learning performance in regional landslide susceptibility prediction}},
  author={Kumar, Chandan and Walton, Gabriel and Santi, Paul and Luza, Carlos},
  journal={Remote Sensing},
  volume={17},
  number={2},
  pages={213},
  year={2025},
  publisher={MDPI},
  doi={10.3390/rs17020213}
}

@article{usta2024comparison,
  title={Comparison of tree-based ensemble learning algorithms for landslide susceptibility mapping in Murgul (Artvin), Turkey},
  author={Usta, Ziya and Ak{\i}nc{\i}, Halil and Ak{\i}n, Alper Tunga},
  journal={Earth Science Informatics},
  volume={17},
  number={2},
  pages={1459--1481},
  year={2024},
  doi="10.1007/s12145-024-01259-w"
}

@article{saravanan2023flood,
  title={Flood susceptibility mapping using machine learning boosting algorithms techniques in Idukki district of Kerala India},
  author={Saravanan, Subbarayan and Abijith, Devanantham and Reddy, Nagireddy Masthan and Kss, Parthasarathy and Janardhanam, Niraimathi and Sathiyamurthi, Subbarayan and Sivakumar, Vivek},
  journal={Urban Climate},
  volume={49},
  pages={101503},
  year={2023},
  doi="10.1016/j.uclim.2023.101503"
}

@article{huang2024modelling,
  title={Modelling landslide susceptibility prediction: A review and construction of semi-supervised imbalanced theory},
  author={Huang, Faming and Xiong, Haowen and Jiang, Shui-Hua and Yao, Chi and Fan, Xuanmei and Catani, Filippo and Chang, Zhilu and Zhou, Xiaoting and Huang, Jinsong and Liu, Keji},
  journal={Earth-Science Reviews},
  volume={250},
  pages={104700},
  year={2024},
  doi="10.1016/j.earscirev.2024.104700"
}

@article{pourzangbar2025analysis,
  title={Analysis of the utilization of machine learning to map flood susceptibility},
  author={Pourzangbar, Ali and Oberle, Peter and Kron, Andreas and Franca, M{\'a}rio J},
  journal={Journal of Flood Risk Management},
  volume={18},
  number={2},
  pages={e70042},
  year={2025},
  doi="10.1111/jfr3.70042Digital Object Identifier (DOI)"
}

@article{chalkias2020exploring,
  title={Exploring spatial non-stationarity in the relationships between landslide susceptibility and conditioning factors: a local modeling approach using geographically weighted regression},
  author={Chalkias, Christos and Polykretis, Christos and Karymbalis, Efthimios and Soldati, Mauro and Ghinoi, Alessandro and Ferentinou, Maria},
  journal={Bulletin of Engineering Geology and the Environment},
  volume={79},
  number={6},
  pages={2799--2814},
  year={2020},
  doi="10.1007/s10064-020-01733-x"
}

@article{lin2021spatial,
  title={Spatial prediction of flood-prone areas using geographically weighted regression},
  author={Lin, Jia Min and Billa, Lawal},
  journal={Environmental Advances},
  volume={6},
  pages={100118},
  year={2021},
  doi="10.1016/j.envadv.2021.100118"
}

@article{li2022spatial,
  title={Spatial non-stationarity-based landslide susceptibility assessment using PCAMGWR model},
  author={Li, Yange and Huang, Shuangfei and Li, Jiaying and Huang, Jianling and Wang, Weidong},
  journal={Water},
  volume={14},
  number={6},
  pages={881},
  year={2022},
  doi="10.3390/w14060881"
}

@article{dai2023examining,
  title={Examining the spatially varying relationships between landslide susceptibility and conditioning factors using a geographical random forest approach: A case study in Liangshan, China},
  author={Dai, Xiaoliang and Zhu, Yunqiang and Sun, Kai and Zou, Qiang and Zhao, Shen and Li, Weirong and Hu, Lei and Wang, Shu},
  journal={Remote Sensing},
  volume={15},
  number={6},
  pages={1513},
  year={2023},
  doi="10.3390/rs15061513"
}

@article{zhang2026interpretable,
  author		= "Zhang, Xuan and Guo, Dongdong",
  title			= "Interpretable machine learning framework for urban flood susceptibility assessment: a multi-model comparison with spatial heterogeneity analysis in Yancheng",
  journal		= "Scientific Reports",
  volume		= "16",
  pages			= "21315",
  year			= "2026",
  doi			= "10.1038/s41598-026-47925-5"
}

@article{xu2025ecoregion,
  title={Ecoregion-Based Landslide Susceptibility Mapping: A Spatially Partitioned Modeling Strategy for Oregon, USA},
  author={Xu, Zhixiang and Zuo, Peng and Zhao, Wen and Zhou, Zeyu and Shao, Xiangyu and Yu, Junpo and Yu, Haize and Wang, Weijie and Gan, Junwei and Duan, Jinshun and others},
  journal={Applied Sciences},
  volume={15},
  number={20},
  pages={11242},
  year={2025},
  doi="10.3390/app152011242"
}

@article{teke2024spatially,
  title={Spatially aware landslide susceptibility prediction using a geographical random forest approach},
  author={Teke, A and Kavzoglu, T},
  journal={The International Archives of the Photogrammetry, Remote Sensing and Spatial Information Sciences},
  volume={48},
  pages={363--370},
  year={2024},
  doi="10.5194/isprs-archives-XLVIII-4-W9-2024-363-2024"
}

@article{wang2022transfer,
  title={Transfer learning for landslide susceptibility modeling using domain adaptation and case-based reasoning},
  author={Wang, Zhihao and Goetz, Jason and Brenning, Alexander},
  journal={Geoscientific Model Development},
  volume={15},
  number={23},
  pages={8765--8784},
  year={2022},
  doi="10.5194/gmd-15-8765-2022"
}

@article{zhao2024heterogeneous,
  title={Heterogeneous transfer learning considering feature representation and environmental consistency for landslide spatial prediction},
  author={Zhao, Zheng and Wang, Weihong and Chen, Jianhua and Yao, Jiaming and Liao, Yangyang and Liu, Jie},
  journal={GIScience \& Remote Sensing},
  volume={61},
  number={1},
  pages={2349343},
  year={2024},
  doi="10.1080/15481603.2024.2349343"
}

@article{wu2026multi,
  title={Multi-Scale High-Resolution Urban Flood Susceptibility Mapping Using MaxEnt and Multi-Source Geospatial Data},
  author={Wu, Xianyu and Lin, Hui and Xiao, Xin},
  journal={Remote Sensing},
  volume={18},
  number={11},
  pages={1864},
  year={2026},
  doi="10.3390/rs18111864"
}

@article{seleem2022transferability,
  title={Transferability of data-driven models to predict urban pluvial flood water depth in Berlin, Germany},
  author={Seleem, Omar and Ayzel, Georgy and Bronstert, Axel and Heistermann, Maik},
  journal={Natural Hazards and Earth System Sciences Discussions},
  volume={2022},
  pages={1--23},
  year={2022},
  doi="10.5194/nhess-23-809-2023"
}

@article{singh2024ensembled,
  title={Ensembled transfer learning approach for error reduction in landslide susceptibility mapping of the data scare region},
  author={Singh, Ankit and Dhiman, Nitesh and Niraj, KC and Shukla, Dericks Praise},
  journal={Scientific Reports},
  volume={14},
  number={1},
  pages={29060},
  year={2024},
  doi="10.1038/s41598-024-76541-4"
}

@article{koldasbayeva2025foundation,
  title={Foundation for unbiased cross-validation of spatio-temporal models for species distribution modeling},
  author={Koldasbayeva, Diana and Zaytsev, Alexey},
  journal={Ecological Informatics},
  pages={103521},
  year={2025},
  doi="10.1016/j.ecoinf.2025.103521"
}

@article{lu2025geographically,
  title={Geographically weighted random forest based on spatial factor optimization for the assessment of landslide susceptibility},
  author={Lu, Feifan and Zhang, Guifang and Wang, Tonghao and Ye, Yumeng and Zhao, Qinghao},
  journal={Remote Sensing},
  volume={17},
  number={9},
  pages={1608},
  year={2025},
  doi="10.3390/rs17091608"
}

@article{rehman2022multi,
  title={Multi-Hazard Susceptibility Assessment Using the Analytical Hierarchy Process and Frequency Ratio Techniques in the Northwest Himalayas, Pakistan},
  author={Rehman, Adnanul and Song, Jinxi and Haq, Fazlul and Mahmood, Shakeel and Ahamad, Muhammad Irfan and Basharat, Muhammad and Sajid, Muhammad and Mehmood, Muhammad Sajid},
  journal={Remote Sensing},
  volume={14},
  number={3},
  pages={554},
  year={2022},
  doi="10.3390/rs14030554"
}

@article{sajid2025integrated,
  title={Integrated Risk Assessment of Floods and Landslides in Kohistan, Pakistan},
  author={Sajid, Taliah and Maimoon, Sakina Khuzema and Waseem, Muhammad and Ahmed, Shiraz and Khan, Muhammad Arsalan and Tr{\"a}nckner, Jens and Pasha, Ghufran Ahmed and Hamidifar, Hossein and Skoulikaris, Charalampos},
  journal={Sustainability},
  volume={17},
  number={8},
  pages={3331},
  year={2025},
  doi="10.3390/su17083331"
}

@article{chauhan2025machine,
  title={Machine learning and GIS-based multi-hazard risk modeling for Uttarakhand: integrating seismic, landslide, and flood susceptibility with socioeconomic vulnerability},
  author={Chauhan, Vipin and Gupta, Laxmi and Dixit, Jagabandhu},
  journal={Environmental and Sustainability Indicators},
  volume={26},
  pages={100664},
  year={2025},
  doi="10.1016/j.indic.2025.100664"
}

@article{drakes2022social,
  title={Social vulnerability in a multi-hazard context: a systematic review},
  author={Drakes, Oronde and Tate, Eric},
  journal={Environmental research letters},
  volume={17},
  number={3},
  pages={033001},
  year={2022},
  doi="10.1088/1748-9326/ac5140"
}

@article{chen2024urban,
  title={Urban flood risk assessment based on a combination of subjective and objective multi-weight methods},
  author={Chen, Jinyi and Gao, Cheng and Zhou, Hong and Wang, Qian and She, Liangliang and Qing, Dandan and Cao, Chunyan},
  journal={Applied Sciences},
  volume={14},
  number={9},
  pages={3694},
  year={2024},
  doi="10.3390/app14093694"
}

@article{kendagannaswamy2025multi,
  title={Multi-criteria decision analysis for regional-scale flood susceptibility mapping in Kerala state, India},
  author={Kendagannaswamy, MS and Roopa, CK and Harish, BS and Mukesh, MS},
  journal={Discover Applied Sciences},
  volume={7},
  number={6},
  pages={598},
  year={2025},
  doi="10.1007/s42452-025-07182-z"
}

@article{thaivalappil2024investigating,
  title={Investigating the impact of recent and future urbanization on flooding in an Indian River catchment},
  author={Thaivalappil Sukumaran, Sonu and Birkinshaw, Stephen J},
  journal={Sustainability},
  volume={16},
  number={13},
  pages={5652},
  year={2024},
  doi="10.3390/su16135652"
}

@article{sarun2026spatiotemporal,
  title={Spatiotemporal analysis of the linkage between environmental degradation and flood susceptibility in an urban tropical wetland ecosystem},
  author={Sarun, S and Vineetha, P},
  journal={Discover Cities},
  volume={3},
  number={1},
  pages={122},
  year={2026},
  doi="10.1007/s44327-026-00285-1"
}

@article{achu2020spatial,
  title={Spatial modelling of shallow landslide susceptibility: a study from the southern Western Ghats region of Kerala, India.},
  author={Achu, Ashokan Laila and Aju, Chandrika Dhanapalan and Reghunath, Rajesh},
  journal={Annals of GIS},
  volume={26},
  number={2},
  pages={113--131},
  year={2020},
  doi=" 10.1080/19475683.2020.1758207"
}

@article{sajinkumar2015geomorphic,
  title={Geomorphic appraisal of landslides on the windward slope of Western Ghats, southern India},
  author={Sajinkumar, KS and Anbazhagan, S},
  journal={Natural Hazards},
  volume={75},
  number={1},
  pages={953--973},
  year={2015},
  doi="10.1007/s11069-014-1358-2"
}

@article{aryal2020model,
  title={A model-based flood hazard mapping on the southern slope of Himalaya},
  author={Aryal, Dibit and Wang, Lei and Adhikari, Tirtha Raj and Zhou, Jing and Li, Xiuping and Shrestha, Maheswor and Wang, Yuanwei and Chen, Deliang},
  journal={Water},
  volume={12},
  number={2},
  pages={540},
  year={2020},
  doi="10.3390/w12020540"
}

@article{zhang2019size,
  title={How size and trigger matter: analyzing rainfall-and earthquake-triggered landslide inventories and their causal relation in the Koshi River basin, central Himalaya},
  author={Zhang, Jianqiang and van Westen, Cees J and Tanyas, Hakan and Mavrouli, Olga and Ge, Yonggang and Bajrachary, Samjwal and Gurung, Deo Raj and Dhital, Megh Raj and Khanal, Narendral Raj},
  journal={Natural hazards and earth system sciences},
  volume={19},
  number={8},
  pages={1789--1805},
  year={2019},
  doi="10.5194/nhess-19-1789-2019"
}

@article{gnyawali2023framework,
  title={Framework for rainfall-triggered landslide-prone critical infrastructure zonation},
  author={Gnyawali, Kaushal and Dahal, Kshitij and Talchabhadel, Rocky and Nirandjan, Sadhana},
  journal={Science of the Total Environment},
  volume={872},
  pages={162242},
  year={2023},
  doi="10.1016/j.scitotenv.2023.162242"
}

@misc{FriedlSullaMenashe2022,
  author    = {Friedl, Mark and Sulla-Menashe, Damien},
  title     = {{MODIS/Terra+Aqua Land Cover Type Yearly L3 Global 500m SIN Grid V061}},
  year      = {2022},
  publisher = {NASA EOSDIS Land Processes Distributed Active Archive Center},
  doi       = {10.5067/MODIS/MCD12Q1.061},
  note      = {Accessed: 2026-08-01}
}

@misc{PoggioDeSousa2020,
  author    = {Poggio, Laura and de Sousa, Luis},
  title     = {{SoilGrids250m 2.0: WRB Classes and Probabilities}},
  year      = {2020},
  publisher = {ISRIC -- World Soil Information},
  doi       = {10.17027/isric-soilgrids.c4dc161c-d62d-11ea-a1a3-292680b15169},
  note      = {Digital soil map}
}

@book{IUSSWorkingGroupWRB2006,
  author    = {{IUSS Working Group WRB}},
  title     = {{World Reference Base for Soil Resources 2006: A Framework for International Classification, Correlation and Communication}},
  year      = {2006},
  series    = {World Soil Resources Reports},
  number    = {103},
  publisher = {Food and Agriculture Organization of the United Nations},
  address   = {Rome},
  isbn      = {9789251055113}
}

@article{StyronPagani2020,
  author  = {Styron, Richard and Pagani, Marco},
  title   = {The {GEM} Global Active Faults Database},
  journal = {Earthquake Spectra},
  year    = {2020},
  volume  = {36},
  number  = {1\_suppl},
  pages   = {160--180},
  doi     = {10.1177/8755293020944182}
}
\end{document}